%% file: arxiv_version.tex
\documentclass{article}

\PassOptionsToPackage{numbers}{natbib}

    \usepackage[preprint]{neurips_2024}

\usepackage[utf8]{inputenc} 
\usepackage[T1]{fontenc}    
\usepackage{hyperref}       
\usepackage{url}            
\usepackage{booktabs}       
\usepackage{amsfonts}       
\usepackage{nicefrac}       
\usepackage{microtype}      
\usepackage{xcolor}         
\usepackage{mathrsfs}
\usepackage{graphicx}
\usepackage{subfig}
\usepackage{listings}
\usepackage{hyperref}
\usepackage{enumitem}
\usepackage[noend]{algpseudocode}
\usepackage{verbatim}
\usepackage{todonotes}
\usepackage{subfloat} 
\usepackage{complexity}
\usepackage{algorithm}
\usepackage{amsmath}
\usepackage{amsthm}
\usepackage{amssymb}
\usepackage{mathtools}
\usepackage{nicefrac} 
\usepackage{bbm}
\usepackage{wrapfig}
\usepackage{array}
\usepackage{caption}  

\theoremstyle{plain}
\newtheorem{theorem}{Theorem}[section]
\newtheorem{proposition}[theorem]{Proposition}

\newtheorem{lemma}[theorem]{Lemma}

\theoremstyle{definition}
\newtheorem{definition}[theorem]{Definition}

\theoremstyle{remark}
\newtheorem{remark}[theorem]{Remark}

\title{The Fairness-Quality Trade-off in Clustering}

\author{%
  \hspace{11mm}Rashida Hakim \\
  \hspace{11mm}Columbia University \\
  \And
  \hspace{5mm}Ana-Andreea Stoica \\
  \hspace{5mm}Max Planck Institute for\\
\hspace{5mm}Intelligent Systems, T\"{u}bingen \\
  \AND
  Christos H. Papadimitriou \\
  Columbia University \\
    \And
  Mihalis Yannakakis \\
  Columbia University 
}

\begin{document}

\maketitle

\begin{abstract}
Fairness in clustering has been considered extensively in the past; however, the trade-off between the two objectives --- e.g., can we sacrifice just a little in the quality of the clustering to significantly increase fairness, or vice-versa? --- has rarely been addressed.  We introduce novel algorithms for tracing the complete trade-off curve, or Pareto front, between quality and fairness in clustering problems; that is, computing all clusterings that are not dominated in both objectives by other clusterings. Unlike previous work that deals with specific objectives for quality and fairness, we deal with all objectives for fairness and quality in two general classes encompassing most of the special cases addressed in previous work. Our algorithm must take exponential time in the worst case as the Pareto front itself can be exponential. Even when the Pareto front is polynomial, our algorithm may take exponential time, and we prove that this is inevitable unless P = NP. However, we also present a new polynomial-time algorithm for computing the entire Pareto front when the cluster centers are fixed, and for perhaps the most natural fairness objective: minimizing the sum, over all clusters, of the imbalance between the two groups in each cluster.~\looseness=-1  
\end{abstract}
 

\section{Introduction}
\label{sec:intro}

Clustering is a fundamental problem in unsupervised learning with many applications, in which data points must be grouped into clusters so that the {\em quality} or {\em cost} of the clustering is optimized --- where the cost can be $k$-median or $k$-center, among a long array of different quality objectives treated in the literature.  In some of these applications the data points being clustered are people --- for example, when households are clustered to determine the ideal location of a bus stop or a hospital --- and in those cases considerations such as fairness and equity come into play.  In recent work, various frameworks and methods have been proposed for improving the representation of different sensitive groups in clustering, in cases where optimizing for clustering cost alone may be quite unfair. In such works, the fairness criterion is often set as a constraint, for which we must compute the optimal solution. This is generally an intractable computational problem even for the simplest fairness objectives, since clustering on its own is NP-hard. The best one could hope for is an approximate solution based on a vanilla approximation algorithm for clustering that improves fairness; such approaches have indeed been proposed recently, e.g.,~\citet{esmaeili2021fair}, see~\citet{chhabra2021overview} for a survey.~\looseness=-1

We take a more general approach to fair clustering: We aim to recover the entire {\em Pareto front} of  clustering cost and fairness, rather than a single point of that front (such as the best-quality clustering under a fairness constraint). Our work aims to enable a practitioner to choose any trade-off point, rather than computing one of them.  We give general sufficient conditions for which we can recover the Pareto front, up to an approximation inherited from the clustering problem itself, and encompassing a variety of quality and fairness objectives used in the literature.~\looseness=-1

\subsection{Our contributions}

We present a novel family of algorithms for tracing the Pareto front between a quality/cost objective and a fairness objective, where each objective can belong in a general class that encompasses most of the ones previously proposed in the literature. 
Our main algorithm requires that the fairness objective satisfy
an intuitive property: that it is \textit{pattern-based}.
Informally, a fairness objective is pattern-based if it is solely a function of the number of nodes of different attributes in each cluster (see Definition~\ref{defn:patternbasedfairness}).
Many group fairness objectives used in the literature are pattern-based, from the well-known balance~\cite{chierichetti2017fair} to objectives that measure the proportional violation of pre-specified upper and lower bounds for a sensitive attribute within a cluster~\cite{bercea2018cost,huang2019coresets,ahmadian2019clustering,bera2019fair,esmaeili2021fair,gupta2023efficient,dickerson2024doubly}.  For the quality objective we consider metric-based cost functions.
We study the computation of the quality/fairness trade-off in two settings: in the {\em assignment problem}, the centers of the clusters are given and the cost is the sum of the distances of the data points from their cluster center; whereas in the {\em clustering problem}, determining the centers is a part of the problem.  We show that our algorithm finds the exact Pareto front of the assignment problem, and an approximation of the Pareto front of the clustering problem (Theorems~\ref{thm:dynprogr-FC}, \ref{thm:nonmerge-FC}), with an approximation ratio that depends on the baseline approximation ratio (denoted $\alpha$) for the minimum cost clustering problem without a fairness objective. The running time is $O(kn^{l(k-1)})$,  where $k$ is the number of clusters, $l$ is the number of sensitive attribute values, and $n$ is the size of the dataset.~\looseness=-1

Our Pareto front algorithms take time exponential in $\ell$ and $k$, and this is necessary to obtain a good approximation. One reason is that the Pareto front may be exponential in size. For many fairness objectives the Pareto front is provably polynomial in size, but even in those cases exponential time is still necessary in general because the baseline problem is NP-hard. For comparison, we explore alternative methods: we extend a recent algorithm~\cite{esmaeili2021fair} for optimizing particular fairness objectives under a bounded clustering cost to recover the entire Pareto front by performing a sweep on cost bound. Our extended heuristic, denoted \texttt{repeated-FCBC}, recovers a faster yet looser approximation of the Pareto front than our dynamic programming approach (Section~\ref{sec:experiments}). In doing so, we explore the runtime-approximation guarantees trade-off in computing the Pareto front. 

Importantly, we also present the first nontrivial algorithmic result for the quality/fairness Pareto front problem in clustering: A {\em polynomial-time algorithm} for computing the Pareto front for an appealing fairness objective, namely the one minimizing the sum (or the maximum), over all clusters, of the deviations from parity of the two protected attributes in each cluster (see Theorem~\ref{thm:matching}). 

We believe that our work is the first to address and carry out the computation of the entire Pareto front, and also the first to simultaneously cover a large range of quality and fairness criteria. In comparison, extant literature on fair clustering typically optimizes one objective subject to a constraint on the other objective, with an individualized approach for each combination. Our work is particularly applicable in cases where a decision-maker may be willing to be suboptimal in one objective in order to achieve a better value in a second objective, but does not know \emph{a priori} what bounds to set on either objective. Our algorithms allow them to explore the entire trade-off, from the highest quality clustering all the way down to the fairest clustering. Finally, our polynomial-time algorithm for computing the Pareto front when the fairness objective is the sum or maximum of imbalances (see Theorem~\ref{thm:matching}) is another novel contribution to the subject of fair clustering.


\subsection{Related work}
\label{sec:relwork}
Unsupervised clustering is known to be NP-hard even in simple settings~\cite{aloise2009np}. A variety of algorithms have been proposed to approximate the best possible clustering, with approximation ratios differing for different cost objectives (e.g. $k$-means~\cite{ahmadian2019better}, $k$-median~\cite{byrka2017improved}, $k$-center~\cite{hochbaum1986unified}).~\looseness=-1

Our work is inspired by a plethora of recent studies on fair clustering that propose various metrics for improving the representation of different sensitive attributes within clusters: from maximizing the worst ratio between two groups across clusters~\cite{chierichetti2017fair}, to minimizing the discrepancy between cluster representation and proportional representation of a group in a population~\cite{bercea2018cost,huang2019coresets,ahmadian2019clustering,bera2019fair,esmaeili2021fair,gupta2023efficient,dickerson2024doubly}, and to equalizing the clustering cost for various groups~\cite{mahabadi2020individual,negahbani2021better,ghadiri2021socially,chakrabarti2022new,vakilian2022improved,bateni2024scalable}. Many of these works focus on building specialized approaches for each of the fairness objectives defined, and their objective is to find the best quality clustering that satisfies a specified fairness constraint.~\looseness=-1
Optimizing even for simplest fairness objectives can be NP-hard~\cite{esmaeili2021fair}, with approximation guarantees involving a multiplicative and additive factor, which may depend on the particular objective form and data topology. Our work differs in two significant ways: first, we compute the entire trade-off curve rather than a single optimization point on the curve; second, we provide algorithms that are agnostic to the specific objectives used, giving sufficient conditions on the objectives (which we show most fairness objectives in the literature satisfy).~\looseness=-1

Multi-objective optimization has seen a variety of methods for computing points close to the Pareto front in various domains, with methods ranging from evolutionary algorithms~\cite{tamaki1996multi,deb2002fast,parsopoulos2002particle,zhang2007moea} to gradient-based methods~\cite{boyd2004convex,liu2021stochastic,lin2019pareto,mahapatra2020multi}, and more recently to hypernetworks~\cite{lin2020controllable,ruchte2021scalable,chen2022multi,hoang2023improving} (often requiring inputting a preference vector of the objective values that will output a point on the Pareto front). These methods have also been applied to a problem closely related to clustering, that of facility location with multiple objectives~\cite{redondo2015approximating,hajipour2016multi,rahmati2013soft}, considering specific objectives and without theoretical approximation guarantees.~\looseness=-1 

Computing the entire Pareto front remains a difficult problem. When an underlying single-objective optimization problem is NP-hard, such as in many combinatorial optimization problems, computing an approximation for the Pareto front is the best one could hope for~\cite{papadimitriou2000approximability}, for example for the multiobjective shortest path problem~\cite{kostreva1993time,warburton1987approximation,henig1986shortest,bokler2020approximating}, zero one knapsack problem~\cite{villarreal1981multicriteria,klamroth2000dynamic,erlebach2002approximating}, the multiobjective spanning tree problem~\cite{papadimitriou2000approximability,bazgan2013approximation,diakonikolas2010small}, among others. To our knowledge, our work is novel in proposing algorithms for solving a biobjective optimization problem based on clustering and fairness objectives which require only general properties for the fairness objectives to recover an approximate Pareto front with theoretical guarantees.

\section{Preliminaries}
\label{sec:prelims}
We denote by $\mathcal{X}$ the set of data points in $\mathbb{R}^d$. Assume that $\mathcal{X} =\{x_1,\ldots,x_n\}$, and for $j\leq n$ define $\mathcal{X}_j =\{x_1,\ldots,x_j\}$.
A {\em clustering map} is defined as $\phi : \mathcal{X} \rightarrow S$, where $S$ is a set of $k$ cluster centers in $ \mathbb{R}^d$ and $k$ is the number of clusters, considered fixed. A cluster $C_i$ is defined as all the data points for which $\phi(x) = s_i$, where $s_i \in S$ is the center for the $i$-th cluster. We call a \emph{clustering} $\mathcal{C}$ the set of all clusters $C_i$ (so $|\mathcal{C}| = k$),
and by $\cal K$ the set of all possible clusterings.
Finally, we denote by $\sigma :\mathcal{X} \rightarrow [l]$ the map between data points and a set of sensitive attributes, indexed from $1$ to $l$ (which may represent demographic groups, interest groups, etc). We denote by $C^a$ the set of data points in a cluster $C$ with attribute $a$, and by $\mathcal{X}^a$ the set of data points with attribute $a$. In this work, we assume that the sensitive attributes are non-overlapping. One can extend these results for overlapping attributes by creating a new attribute for every attribute overlap set, however, increasing the time complexity in doing so.~\looseness=-1

\paragraph{Clustering and Assignment Problems.} Unsupervised clustering optimizes a clustering cost objective $c$, often defined as the sum of distances between points and a set of candidate cluster centers. In a general form, the cost of a clustering is $\left(\sum_{x\in \mathcal{X}}d^p(x,\phi(x))\right)^{1/p}$, for metric $d$ and some value of $p$. We call such objectives \emph{metric-based}. By varying $p=1, 2, \infty$ we can obtain the $k$-median, $k$-means, and $k$-center objectives, respectively. The {\em clustering problem} is thus finding the clustering that has the minimum cost, over all possible sets of centers and maps from $\cal X$ to the set of centers. Since the clustering problem is hard, one often considers, as a stepping stone, the {\em assignment problem}, where the centers have been fixed. We shall consider both problems in this paper.
In the fair clustering problem we have two objectives, the clustering cost objective $c$, and a fairness objective $f$, that aims to balance the representation of sensitive attributes in clusters, further detailed below. 

\paragraph{Pareto front.} Consider the set of all clusterings $\cal K$ of $\cal X$, and the two objectives $c$ (the cost objective) and $f$ (the fairness objective\footnote{We will treat $f$ as a minimization objective, thus minimizing the \emph{un}fairness of a clustering.}), each mapping $\cal K$ to $\mathbb{R}$. We say that clustering $\cal C$ dominates clustering $\cal C'$ if $c({\cal C})\leq c({\cal C'})$, $f({\cal C})\leq f({\cal C'})$, and one of the inequalities is strict. Intuitively, if a clustering $\mathcal{C}'$ is dominated, it is unworthy of further consideration, because it lags behind in both objectives of interest. If however a clustering $\mathcal{C}$ is {\em un}dominated, that is, there is no clustering in $\cal K$ that is simultaneously better on both fronts, then it is part of the solution. The {\em Pareto front} is the set of all undominated clusterings.~\looseness=-1

\paragraph{Fairness objectives.}  Our main contribution is an algorithm for computing the Pareto front of the clustering and assignment problems for any metric-based quality function, and any fairness objective -- always a function to be minimized -- that satisfies the following general condition.

\begin{definition}[Pattern-based objectives]
For a clustering $\cal C \in \cal K$, $i\in [k]$, and $a\in [l]$, let its \textit{pattern} $P^{\cal C}[i,a]$ be the number of data points in cluster $C_i$ with attribute value $a$. A fairness objective $f$ that maps $\cal K$ to $\mathbb{R}$ is called \textit{pattern-based} if $f({\cal C})$ only depends on the values in $P^{\cal C}[i,a]$.~\looseness=-1
    \label{defn:patternbasedfairness}
\end{definition}

\begin{definition}[Mergeable objectives]
    Consider two  clusterings ${\cal C}$ and ${\cal C'}$.  We say that ${\cal C'}$ is the result of a merging of ${\cal C}$ (or ${\cal C}$ is a refinement of ${\cal C'}$) if 
    every non-empty cluster of ${\cal C'}$ is the union of clusters of ${\cal C}$.
    A fairness objective $f$ that maps $\cal K$ to $\mathbb{R}$ is called \textit{mergeable} if, whenever ${\cal C'}$ is the result of a merging of ${\cal C}$, $f({\cal C'}) \leq f({\cal C})$. In other words, a mergeable fairness objective weakly improves in fairness when existing clusters are combined. As we show in Appendix \ref{sec:appendix-analyzingfairnessobj}, mergeability holds for a wide array of fairness objectives that avoid over or under representation of sensitive attributes.   ~\looseness=-1
    \label{defn:mergeability}
\end{definition}

We discuss the pattern-based and mergeability properties in Appendix~\ref{sec:appendix-analyzingfairnessobj}, where we give examples of fairness objectives that are not pattern-based or mergeable. Below, we introduce several fairness objectives commonly found in the literature, which we use in our experiments. All of these objectives are pattern-based and mergeable, as proved formally in Appendix~\ref{sec:appendix-fairnessproofs}.

\emph{Balance objective (Definition $1$ in~\citet{chierichetti2017fair}):} For sensitive attributes that can take two values, indexed $1$ and $2$, the balance of a cluster $C$ is defined as $\textsc{Balance}(C) = \min \left( |C^1|/|C^2|, |C^2|/|C^1|\right)$. The balance objective for a clustering $\mathcal{C}$ is then defined as 
\begin{equation}
    \textsc{Balance}(\mathcal{C}) =\max \min\limits_{C \in \mathcal{C}} \textsc{Balance}(C)
    \label{eq:balance}
\end{equation}
The aim is to maximize balance, or equivalently, to minimize the negative balance $f(\mathcal{C}) = - \textsc{Balance}(\mathcal{C})$. \textsc{Balance} has been used in fair clustering as a measure of equalizing proportions of different groups across clusters~\cite{chierichetti2017fair}. In practice, optimizing \textsc{Balance} is both difficult and does not measure how far the proportions of sensitive groups in a clustering are from their \textit{true} proportions in the population. For this reason, objectives based on \textit{proportional violation} have been proposed, allowing a central decision-maker to input upper and lower bounds for desired proportions of groups in each cluster, and measured the deviation from these bounds~\cite{bercea2018cost,huang2019coresets,ahmadian2019clustering,bera2019fair,esmaeili2021fair,gupta2023efficient,dickerson2024doubly}.~\looseness=-1

\emph{Proportional violation objectives:} As defined in~\citet{esmaeili2021fair}, for every sensitive attribute $a \in [l]$, define the upper and lower bounds as $\alpha_a$ and $\beta_a$, aimed at satisfying: $\beta_a |C| \leq |C^a| \leq \alpha_a |C|$, for all clusters $C \in \mathcal{C}$. Since this is not always feasible, we define the worst proportional violation of attribute $a$ in cluster $C$ as the minimum non-negative value $\Delta_a^C \in [0,1]$ such that 
\begin{equation}
    \left(\beta_a - \Delta_a^C\right) |C| \leq |C^a| \leq \left(\alpha_a + \Delta_a^C\right) |C|
    \label{eq:delta}
\end{equation}
Then, the proportional violation-based objectives are defined as: 
\begin{align}
        \textsc{Group Utilitarian} &= \min \sum \limits_{a \in [l]} \max\limits_{C \in \mathcal{C}} \Delta_a^C, \quad & \textsc{Group Utilitarian-Sum} &= \min \sum \limits_{a \in [l],  C \in \mathcal{C}} \Delta_a^C, \\ 
        \textsc{Group Egalitarian} &= \min \max\limits_{a \in [l], C \in \mathcal{C}} \Delta_a^C, \quad & \textsc{Group Egalitarian-Sum} &= \min \max \limits_{a \in [l]}\sum\limits_{ C \in \mathcal{C}} \Delta_a^C
        \label{eq:fairness_objectives}
\end{align}

These objective operationalize utilitarian and egalitarian concepts from social choice~\cite{brandt2016handbook}, minimizing either the sum of proportional violations or the worst violation across attributes.

\emph{Sum of imbalances objective:} Finally, for two interest groups in the population ($l=2$) the following objective is quite natural:   
$$\textsc{Sum of Imbalances} = \sum_{i \in [k]}|C_i^1-C_i^2|,$$
that is to say, the sum of the deviations from equality between the two attribute values in the clusters. This objective is most appropriate when datasets contain relatively equal proportions of the two groups. Rather remarkably, the Pareto front for this objective can be computed in polynomial time, highlighting the loss in complexity by not including the population proportions.~\looseness=-1

\section{Algorithms for Computing the Pareto Front} 
\label{sec:theory}

\subsection{A Dynamic Programming Algorithm for Recovering the Assignment Pareto Front}
\label{sec:theory:1}

We shall now present the main Algorithm. Define an $\cal X$-pattern $P$ to be a $k$ by $l$ matrix of non-negative integers such that $\sum_{i\in[k]} P[i, a] = |\mathcal{X}^a|, \forall a \in [l]$. $P[i,a]$ specifies the number of data points in cluster $i$ of attribute $a$. That is, an $\cal X$-pattern is a clustering except only the attribute values of the points have been specified.\footnote{Each clustering $\mathcal{C}$ maps to a pattern $P^{\mathcal{C}}$, with many different clusterings mapping to the same pattern.} Complete proofs to all subsequent results can be found in Appendix~\ref{sec:appendix-proofs}.~\looseness=-1

\begin{algorithm}  
\caption{Dynamic Programming Algorithm for Computing the Assignment Pareto Front}
\begin{algorithmic}[1]
\State \textbf{Input:} Number of clusters $k$, a set of $n$ data points $\mathcal{X}$ with $l$ attribute values, $k$ centers $S = \{s_1,\ldots, s_k\}$, and metric-based cost objective $c$ parameterized by $d, p$.
\State {\bf Output:} A table $T_n$ containing the solutions of the assignment problem for $\mathcal{X}$ for all $\mathcal{X}$-patterns.
\State {\bf Method:} Dynamic programming.  We shall compute $T_0,T_1,\ldots T_n$.
\State Initialize $T_0$ to contain the null pattern with cost 0 and the empty clustering.
\For{$j = 1$ to $n$}
    \State Generate all $\mathcal{X}_j$-patterns, where $\mathcal{X}_j =\{x_1,\ldots,x_j\}$. 
    \State For each $\mathcal{X}_j$-pattern $P$, and for each cluster $i$ such that $P[i,a]>0$, where $a$ is the attribute 
    \State value of $x_j$, look up in $T_{j-1}$ the cost of the pattern $P_i$, which is $P$ with $x_j$ omitted from 
    \State cluster $i$, and compute $T_{j - 1}(P_i) + d^p(x_j,s_i)$. 
    \State Let $i^*$ be the cluster index that minimizes $T_{j - 1}(P_i) + d^p(x_j,s_i)$.
    \State Set $T_j(P) \gets T_{j - 1}(P_{i^*}) + d^p(x_j,s_{i^*})$. 
    \State Store at $T_j(P)$ the clustering from $T[P_{i^*}]$, with $x_j$ added at cluster $i^*$.
\EndFor
\Return $T_n$
\end{algorithmic}
\label{alg:dynprogr}
\end{algorithm}

At the conclusion of the algorithm, the table $T_n$ contains the lowest cost clustering $\mathcal{C}$ for each $\cal X$-pattern $P$, such that $P^{\mathcal{C}} = P$, together with its cost $c(\mathcal{C})$. Then, we can find the Pareto front by first sorting these clusterings for all $\cal X$-patterns $P$ in increasing $c$, and then traversing them in order, computing the unfairness of each pattern, remembering the smallest unfairness we have seen so far, and omitting any pattern that has unfairness larger than the smallest seen so far.~\looseness=-1
\begin{remark} The above calculation of $T_n$ can be achieved alternatively by a simpler to state but slightly slower algorithm:  first generate all $\cal X$-patterns, and then compute the optimum assignment of each by min-cost flow.~\looseness=-1
\end{remark}
\begin{theorem}
    Algorithm~\ref{alg:dynprogr} finds the Pareto front of the assignment problem in time $O(kn^{l(k-1)})$, for any metric-based clustering objective and any pattern-based fairness objective.
    \label{thm:dynprogr-FA}
\end{theorem}
\textbf{Proof Sketch:} We prove this theorem by finding an invariant of Algorithm \ref{alg:dynprogr}: $T_j[P]$ stores the lowest cost assignment of $\mathcal{X}_j$ that maps to pattern $P$. We maintain this invariant as we build up our table by searching over possible smaller patterns that we can add our next datapoint $x_j$ to and creating the lowest cost assignment that maps to $P$. Therefore, the assignments stored at $T_n[P]$ are the candidate points for the Pareto front. A simple filtering heuristic, as described above, removes the dominated points from this set of candidates. In terms of running time, the number of possible patterns of total size up to $n$ is upper bounded by $n^{l(k-1)}$, since each of the $lk$ entries of $P$ takes values between $0$ and $n$, and the row corresponding to the last cluster $k$ is fully determined by the other clusters. For each considered pattern, we need to look up at most $k$ previous entries of the table $T$.~\looseness=-1 

\subsection{Approximating the Pareto Front for the Clustering Problem}
\label{sec:theory:2}

As we saw, Algorithm~\ref{alg:dynprogr} computes the Pareto front of the Assignment problem exactly for any input centers $S$. We next show that it also provides an approximation for the Pareto front of the clustering problem. For this purpose, we first use a vanilla clustering algorithm $\mathcal{A}$ for the single-objective problem of minimizing the cost $c$, to obtain the set $S$ of cluster centers, and then apply Algorithm~\ref{alg:dynprogr} with this set of centers $S$.
Let $\alpha$ be the approximation ratio of algorithm $\mathcal{A}$.~\looseness=-1

\begin{definition}[$\mathcal{W}$-approximation of the Pareto Set for clustering]
For parameters $\mathcal{W} = (w_c, w_f)$, we define the $\mathcal{W}$-approximation of the Pareto set $X_P$ as a set of feasible points $X_P'$ such that 
$\forall x\in X_P,  \exists x'\in X_P'$ such that 
$c(x') \leq w_c\cdot c(x)$ and $f(x') \leq w_f\cdot f(x)$.
\label{defn:defn_eps_approx}    
\end{definition}
This definition is a direct generalization of $\epsilon$-approximate Pareto set defined by~\citet{papadimitriou2000approximability}. One may recover the $\epsilon$-approximate definition by setting $\epsilon = \max(w_c, w_f) - 1$. 

\begin{theorem}
    Algorithm~\ref{alg:dynprogr} finds a $(2 + \alpha, 1)$-approximation of the Pareto set of the clustering problem with a metric-based cost objective $c$ and a pattern-based and mergeable fairness objective $f$. 
    \label{thm:dynprogr-FC}
\end{theorem}
\textbf{Proof Sketch: } We argue that for any clustering map $\phi^*$ with centers $S^*$ in the Pareto set for clustering, there exists an assignment to the centers $S$ found by an approximate vanilla clustering algorithm that achieves the same or better fairness and at most $(2 + \alpha)$ times the clustering cost. Then, since Algorithm~\ref{alg:dynprogr} finds the Pareto set of the assignment problem, we are guaranteed the stated approximation. We construct this assignment using a ``routing'' argument, first introduced in~\citet{bera2019fair}: we create an assignment $\phi'$ by routing all points in $\phi^*$ with center $s^* \in S^*$ to the center in $S$ nearest to $s^*$. Given that the clustering cost is metric-based, we use the triangle inequality on the cost objective to argue that the cost of $\phi'$ w.r.t. $S$ is not more than $(2 + \alpha)$ times the cost of $\phi^*$ w.r.t. $S^*$. Then, we use the mergeability property of the fairness objective to argue that $\phi'$ has a weakly better fairness than $\phi^*$.~\looseness=-1
We can, in fact, modify Algorithm~\ref{alg:dynprogr} to include non-mergeable fairness objectives, guaranteeing the same approximation ratio and time complexity: 

\begin{theorem}
    We can compute a $(2 + \alpha, 1)$-approximation of the Pareto set for the clustering problem with a metric-based cost objective $c$ and a pattern-based fairness objective $f$ in time $O(k n^{l(k-1)})$. 
    \label{thm:nonmerge-FC}
\end{theorem}

\textbf{Proof Sketch:} The trick here is to transform non-mergeable fairness objectives into mergeable ones, and then apply Algorithm \ref{alg:dynprogr}. In doing so, we control this transformation through re-assigning the centers of potentially empty clusters (which become an issue in non-mergeable fairness functions). Specifically, for patterns that have empty clusters, we search for the best possible fairness over all ways to reassign the empty clusters' centers to other centers' locations and divide up the points in the non-empty centers. A detailed description and proof for this modified algorithm can be found in Appendix~\ref{sec:appendix-proofs}.~\looseness=-1

\subsection{Intractability}
The running time of Algorithm~\ref{alg:dynprogr} for computing the Pareto front has the parameters $k,l$ in the exponent of $n$. What evidence do we have that this computation is necessary?

One reason would be that the sheer size of the Pareto front may be exponential. For some objectives, including the \textsc{Balance} and the \textsc{Group Egalitarian} objectives defined in Section~\ref{sec:prelims}, the Pareto front is provably of polynomial size. The reason is that all possible values of these objectives are rational numbers involving integers that are all smaller than $n$, and there are $O(n^2)$ possible different such rational numbers. For other objectives, there may be instances of exponential Pareto fronts, as the objectives that sum over the clusters. 

However, a different argument provides a justification of the algorithm's exponential performance:  In several papers in recent literature (see e.g.~\cite{esmaeili2021fair,esmaeili2022fair}), it is shown that, for a variety of cost functions $c$ and fairness objectives $f$ (including all mentioned proportional violation based objectives), it is NP-hard to find the assignment ${\cal C}$ that has the smallest $c({\cal C})$ under the constraint that $f({\cal C}) \leq F$. We conclude that, unless P = NP, there is no polynomial-time 
algorithm for outputting the Pareto front, for at least some combinations of $c$ and $f$. As a footnote to this discussion, the above complexity argument rules out polynomial algorithms, but not exponential algorithms of the form, for example, $O(2^kn^3)$, which is much more benign than $O(n^{(k-1)l})$. These would be ruled out, subject to complexity conjectures, if the problem of finding the least costly clustering subject to fairness constraints were shown to be W-complete, a more severe form of complexity that rules out this more benign exponential performance~\cite{downey2012parameterized}. We leave this as an interesting open question in the theory of the fairness-cost trade-off of clustering. 

\subsection{A Polynomial Algorithm for the Pareto Front of the Sum of Imbalances}
\label{sec:matching}
For \emph{specific} cost and fairness objectives, there is still hope that polynomial-time algorithms exist. As we show below, for a simple objective derived from the \textsc{Balance} objective, such an algorithm is possible.~\looseness=-1
For the \textsc{Sum of Imbalances} fairness objective and a clustering objective defined in Section~\ref{sec:prelims}, we want to compute the Pareto front when $l = 2$. Surprisingly, it turns out that this problem can be solved in polynomial time by a reduction to the weighted matching problem. 

\begin{theorem} \label{thm:matching}
If $l=2$ and the fairness objective $f$ is the sum of imbalances $f({\cal C})=\sum_{i\in [k]} |C_i^1-C_i^2|$,  then the Pareto front of the assignment problem can be computed in polynomial time.  
\end{theorem}
\textbf{Proof Sketch:} The image of $f$ is contained in the integer set $\{1,\cdots,n\}$. For each potential value $j$, we construct a graph $G_j$ that contains $\cal X$ as nodes and another $j$ `dummy' nodes. We put an edge $(u,v)$ between every $u, v \in \mathcal{X}$ with different sensitive attribute value with weight equal to the cost $\min_i (d^p(u,s_i)+d^p(v,s_i))$. Between every data point $u \in \mathcal{X}$ and dummy point $v$ put an edge $(u,v)$ with cost $\min_{i} d^p(x,s_i)$.
Finding the minimum cost perfect matching in this graph gives the minimum cost of an assignment with fairness value $j$. The same result can be shown similarly for the \textsc{Max Imbalance} objective, $\max_{i \in [k]} | C_i^1 - C_i^2|$ (Theorem~\ref{thm:maximbalance}).~\looseness=-1

\begin{remark}
    These objectives are quite natural extensions of the \textsc{Balance} objective, as they minimize the sum or max, over the $k$ clusters, of the deviation from equality between the two groups. It is worth noting that slight modification of these objectives place the problem back in the NP-complete space: A construction of~\citet{bercea2018cost} (see also~\citet{esmaeili2021fair}, Theorem 5.1) implies that for minimizing the deviations not from equality ($1:1$) but from the ratio $1:3$, it is NP-complete to find even the point of the Pareto front with the best fairness value. Thus, relating the fairness objective to the population proportions adds computational complexity, compared to fairness objectives that simply sum the differences in the number of sensitive attributes present in each cluster.~\looseness=-1   
\end{remark}

\section{Experiments}
\label{sec:experiments}

We implement our proposed algorithm (Algorithm~\ref{alg:dynprogr}) for finding the Pareto front on three real-world datasets and five fairness objectives, as detailed below. 
\paragraph{Datasets:} We use the following real-world datasets for our experiments: the Adult dataset and the Census dataset retrieved from the UCI repository (as the Census1990 version)~\cite{lichman2013uci,kohavi1996scaling}, and the BlueBike trip history dataset.\footnote{The BlueBike trip history dataset is retrieved from \url{https://bluebikes.com/system-data}.} The Adult and Census datasets contain numeric attributes such as income, age, education level, often used in applications of fair clustering~\cite{chierichetti2017fair,backurs2019scalable,esmaeili2021fair}. The BlueBike dataset contains the starting location, ending location, as well as various user attributes for users of the BlueBike bike sharing system in the Boston area. We use as features the starting and ending longitude and latitude values for all rides during a period of a week in May, 2016. These routes are a proxy for common traffic patterns, for which clustering can inform of high-density areas, with the purpose of deciding on new locations for bike stations or public transportation. Clustering and related framings have long been used in facility location problems, for which fairness is a central question~\cite{marsh1994equity,jung2019center,abbasi2021fair,vakilian2022improved}. As the datasets are prohibitively large, we sample from each $1,000$ data points. Further details about the datasets can be found in Appendix~\ref{sec:appendix-datadetails}.~\looseness=-1

\paragraph{Objectives:} We implement the $k$-means clustering objective and five different fairness objectives, as defined in Section~\ref{sec:prelims} (\textsc{Balance}, \textsc{Group Utilitarian}, \textsc{Group Utilitarian-Sum}, \textsc{Group Egalitarian}, and \textsc{Group Egalitarian-Sum}). 

\paragraph{Experimental details:} 
We first note that our approach for finding the Pareto front is agnostic to the specific vanilla clustering algorithm used. For our datasets, we use the $k$-means++ clustering algorithm as the vanilla clustering that has an approximation ratio of $O(log(k))$~\cite{arthur2007k}. All datasets have numeric attributes, allowing a direct embedding into Euclidean space and using k-means++ directly on the features. We use the self-reported gender (male or female) as the sensitive attribute for all datasets. Each sensitive attribute $a$ has a proportion $p_a$ in the general population. For the proportional violation objectives, we set upper and lower bounds as a $\delta$-deviation from the true proportions $(p_a)_a$: $\alpha_a = (1 + \delta)p_a, \beta_a = (1 - \delta)p_a$. We set $\delta \in [0.005, 0.05]$ for all experiments and $k = 2$ clusters. Additional experiments for $k = 3$ are reported in Appendix~\ref{sec:appendix-moreexp}, noting qualitatively similar results. All experiments are run on local computers, using Python 3.9, $k$-means++ \cite{arthur2007k}, NetworkX~\cite{hagberg2013networkx}, and CPLEX for the implementation of the FCBC algorithm~\cite{manual2016version,esmaeili2021fair} with $\epsilon = 2^{-10}$ and $N = 50$ runs. An empirical analysis of the running time for Algorithms~\ref{alg:dynprogr} and~\ref{alg:repeated_fcbc} can be found in Section~\ref{sec:appendix-exp-details}, Figure~\ref{fig:dynprogr_fcbc_runningtime}.\footnote{All code and data used in the paper can be found at \url{https://anonymous.4open.science/r/the_fairness_quality_tradeoff_in_clustering-1109/}.}~\looseness=-1

\subsection{Pareto Front on Real-World Data}

\begin{figure*}[!ht]
\centering
\subfloat[Balance]
{\includegraphics[width=0.2\textwidth] {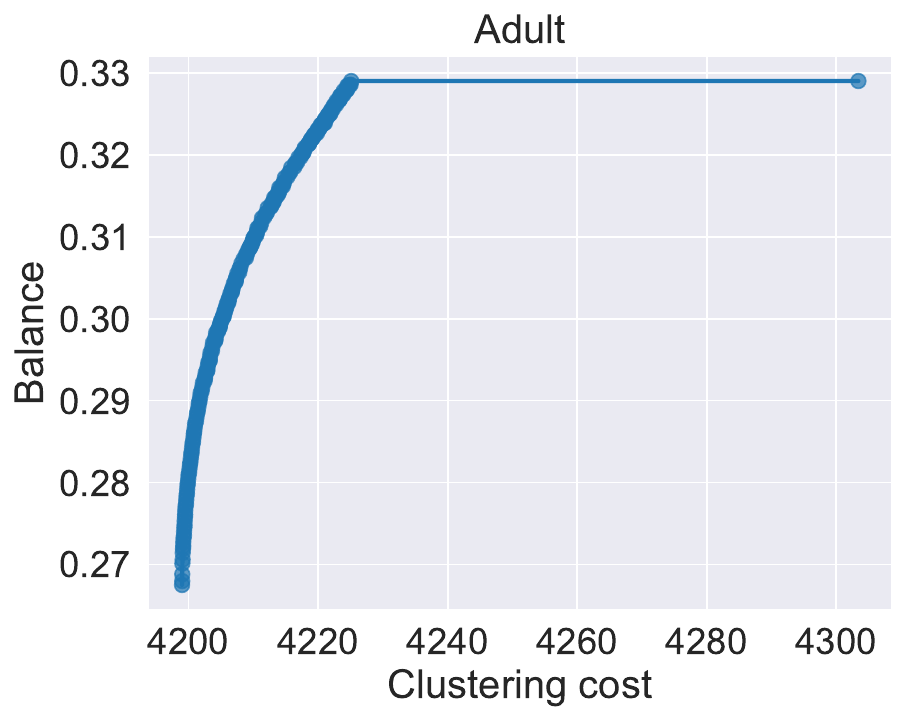}}
\subfloat[Group Util]
{\includegraphics[width=0.2\textwidth] {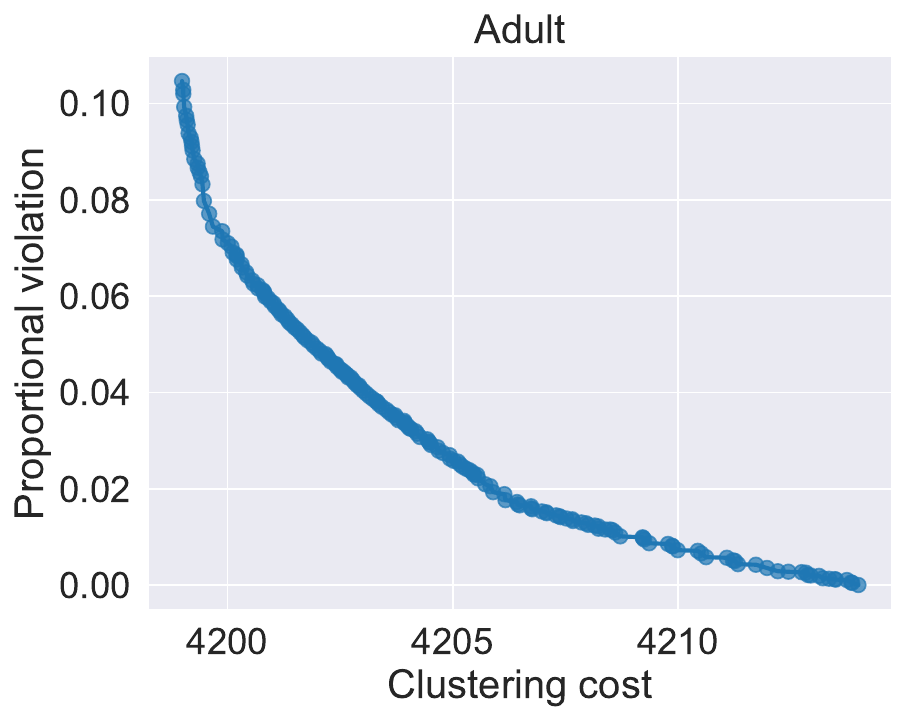}}
\subfloat[Group Util-Sum]
{\includegraphics[width=0.2\textwidth] {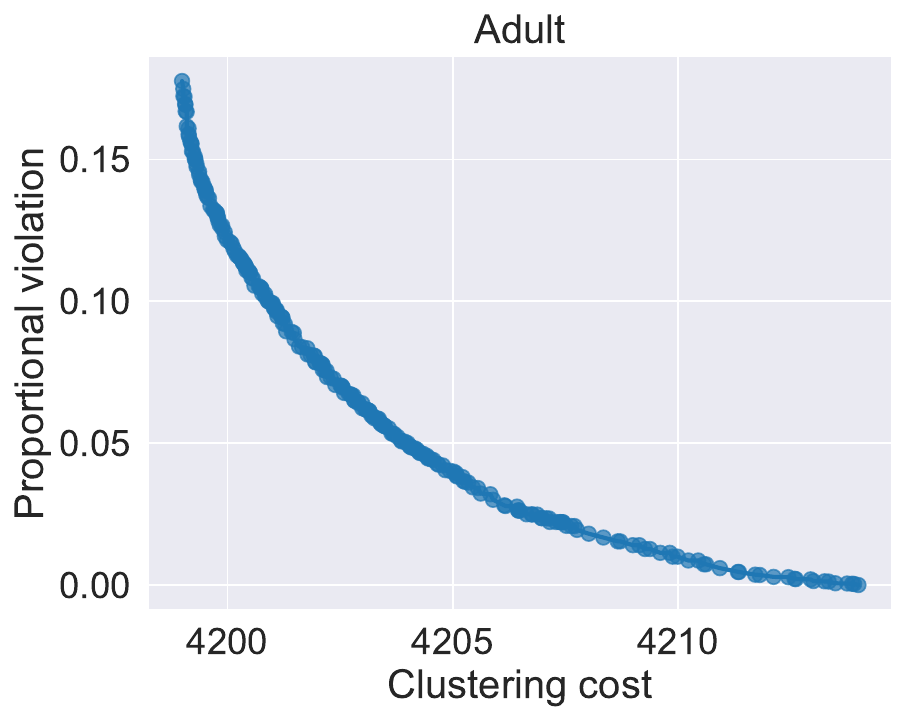}} 
\subfloat[Group Egalit]
{\includegraphics[width=0.2\textwidth] {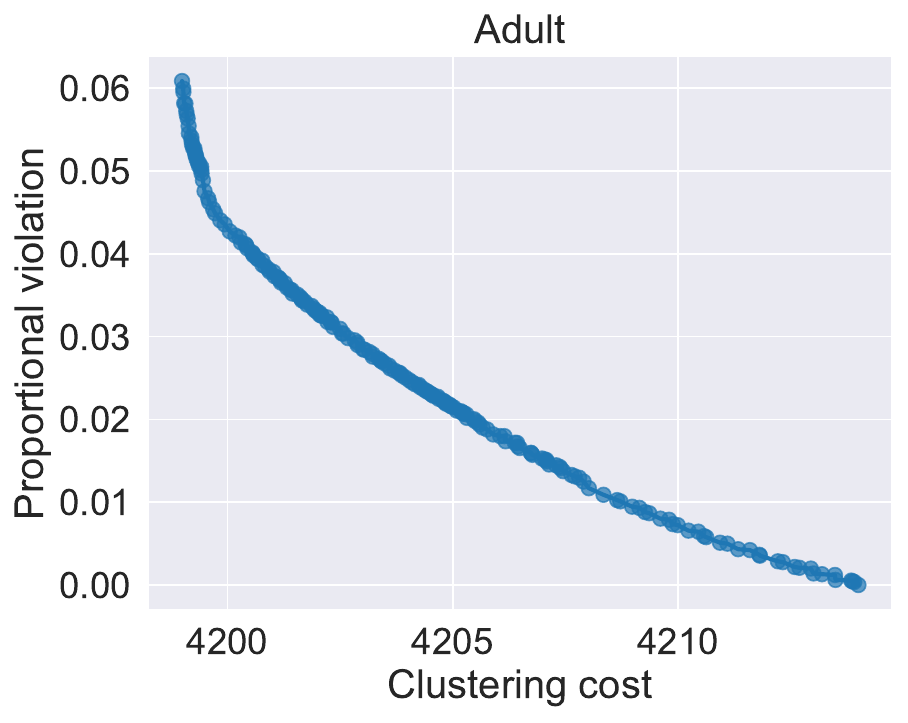}} 
\subfloat[Group Egalit-Sum]
{\includegraphics[width=0.2\textwidth] {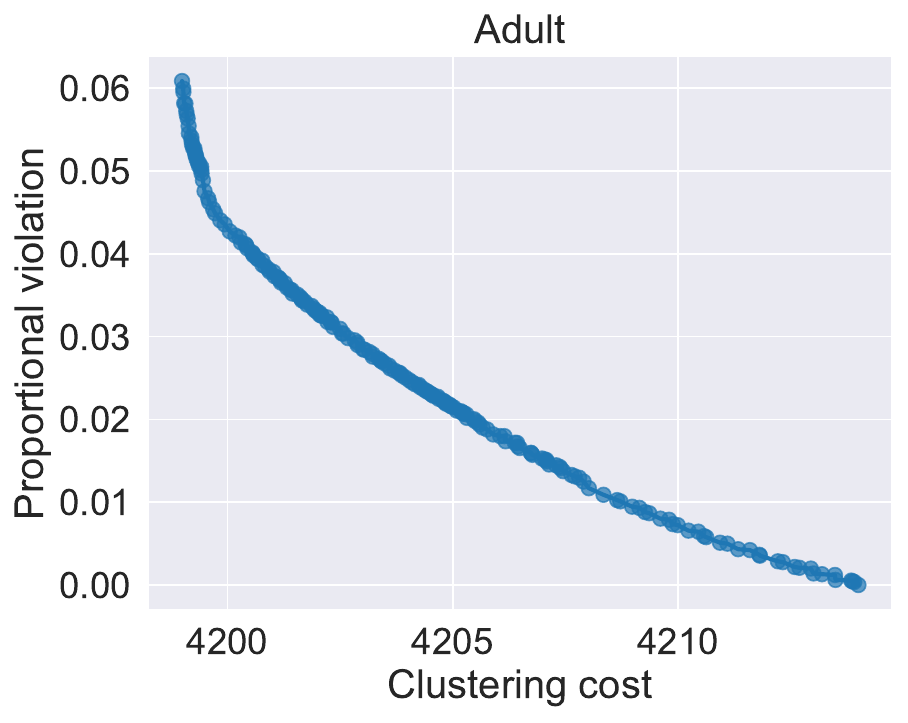}} \\ 
\subfloat[Balance]
{\includegraphics[width=0.2\textwidth] {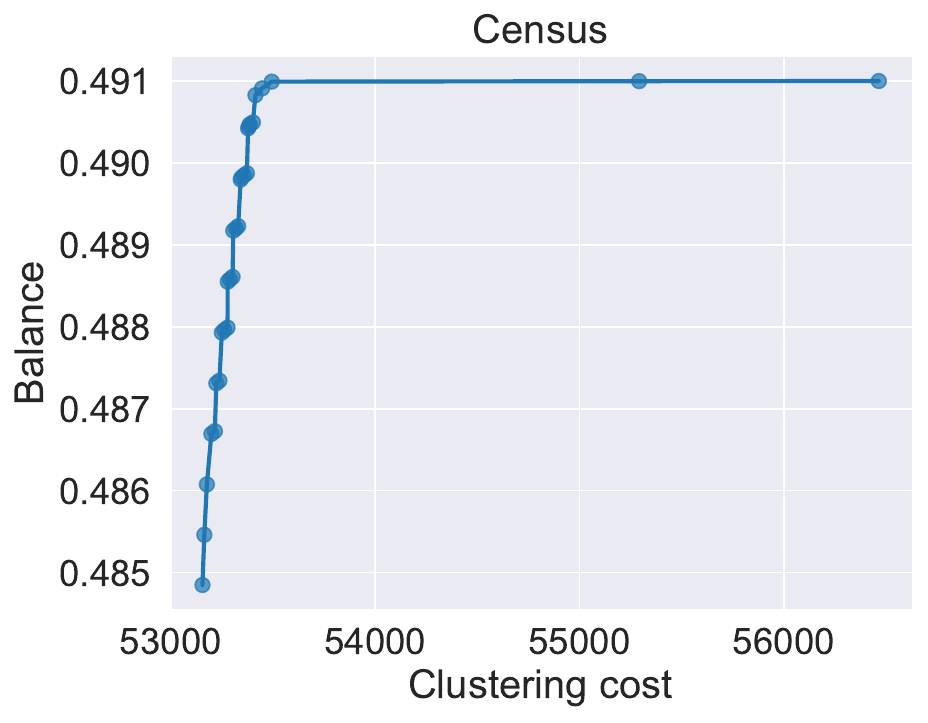}}
\subfloat[Group Util]
{\includegraphics[width=0.2\textwidth] {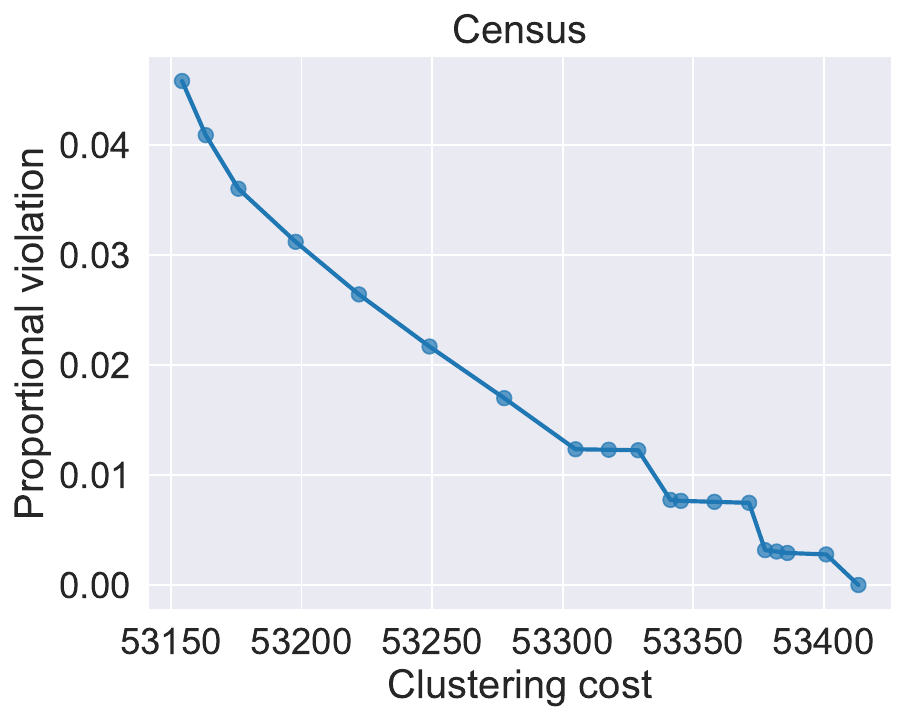}}
\subfloat[Group Util-Sum]
{\includegraphics[width=0.2\textwidth] {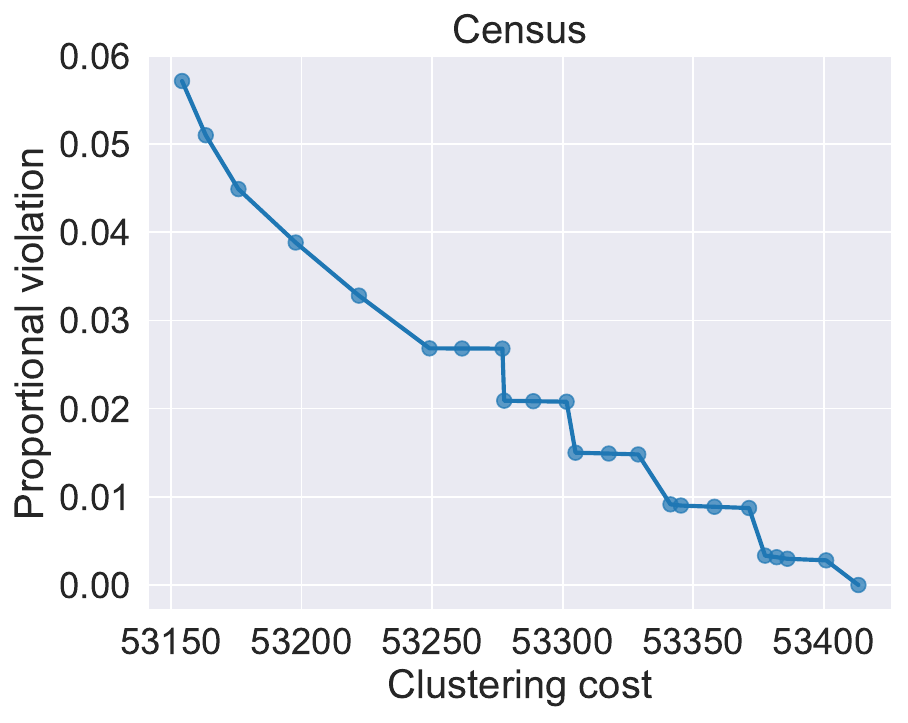}} 
\subfloat[Group Egalit]
{\includegraphics[width=0.2\textwidth] {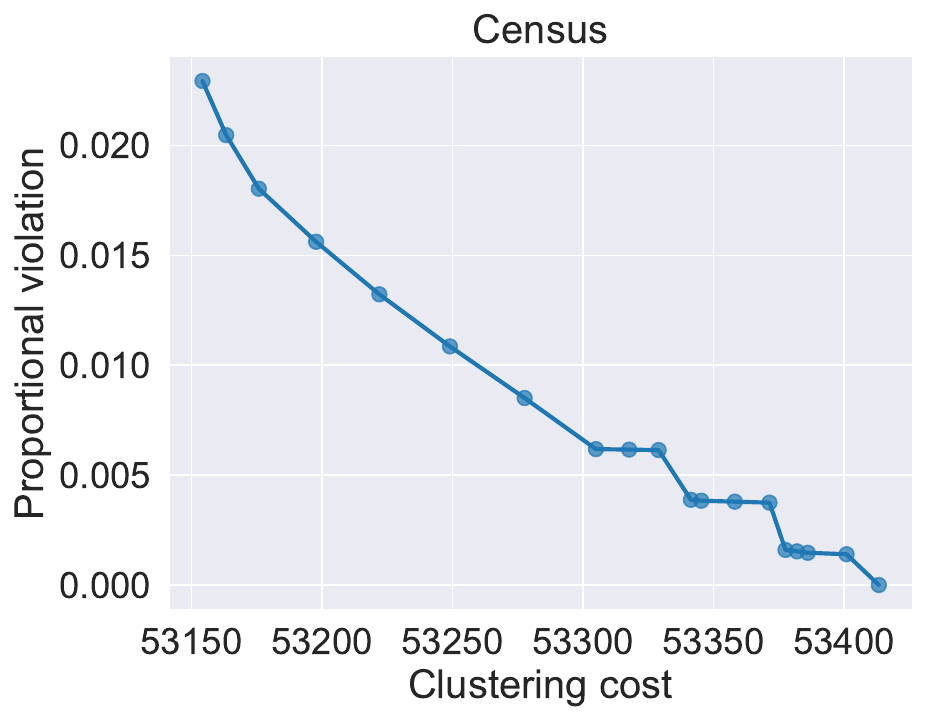}} 
\subfloat[Group Egalit-Sum]
{\includegraphics[width=0.2\textwidth] {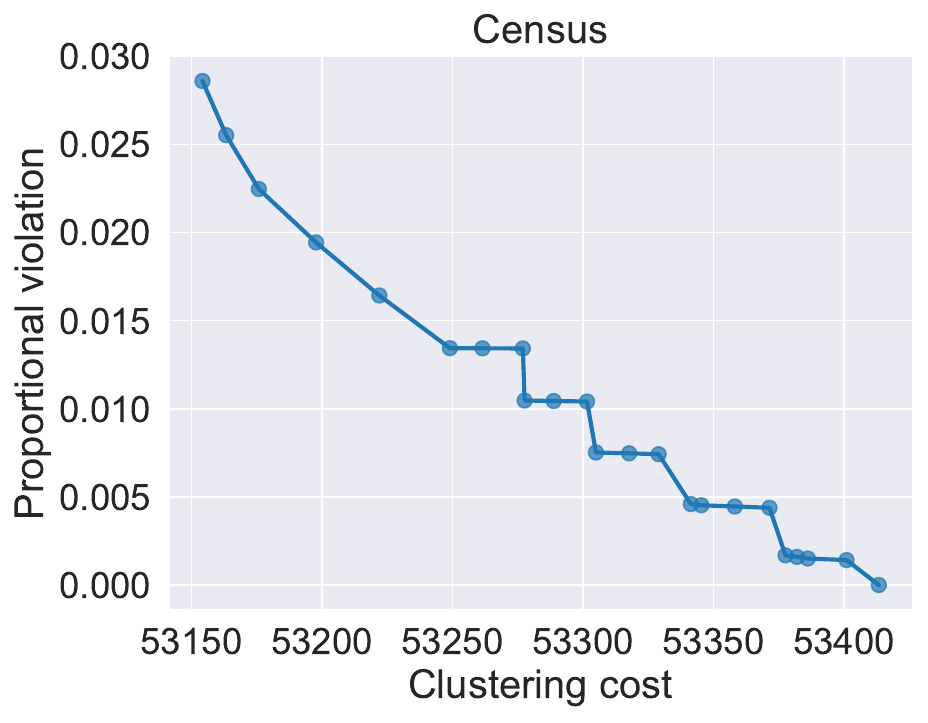}} \\
\subfloat[Balance]
{\includegraphics[width=0.2\textwidth] {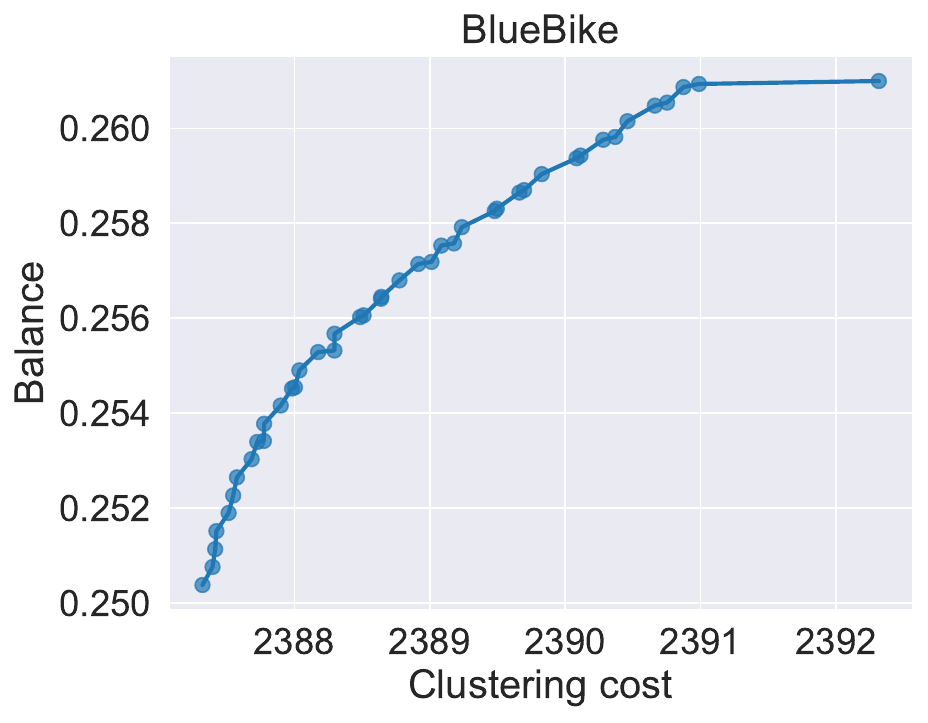}}
\subfloat[Group Util]
{\includegraphics[width=0.2\textwidth] {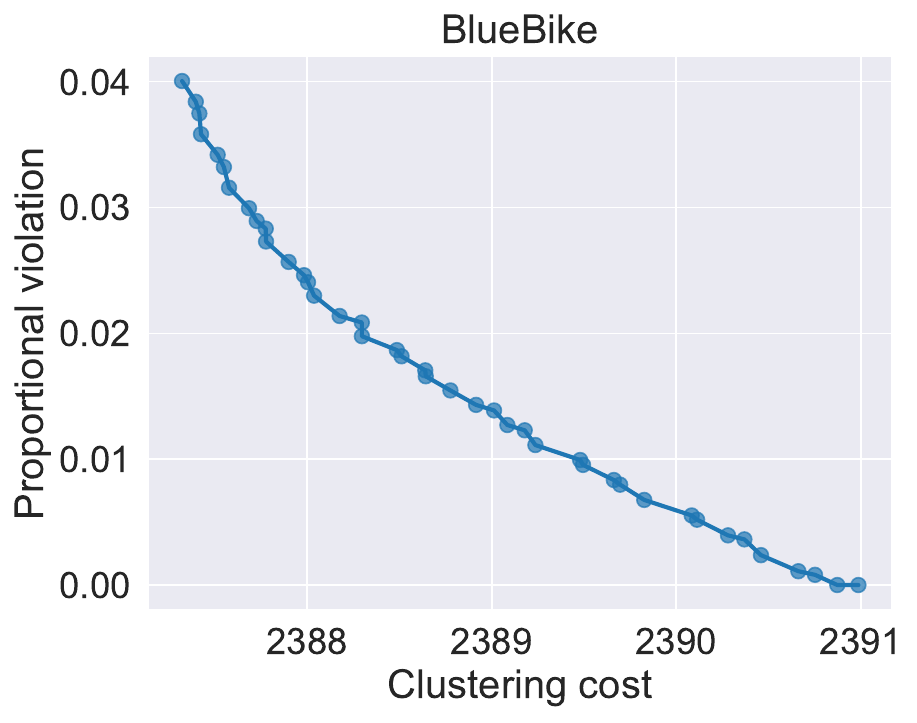}}
\subfloat[Group Util-Sum]
{\includegraphics[width=0.2\textwidth] {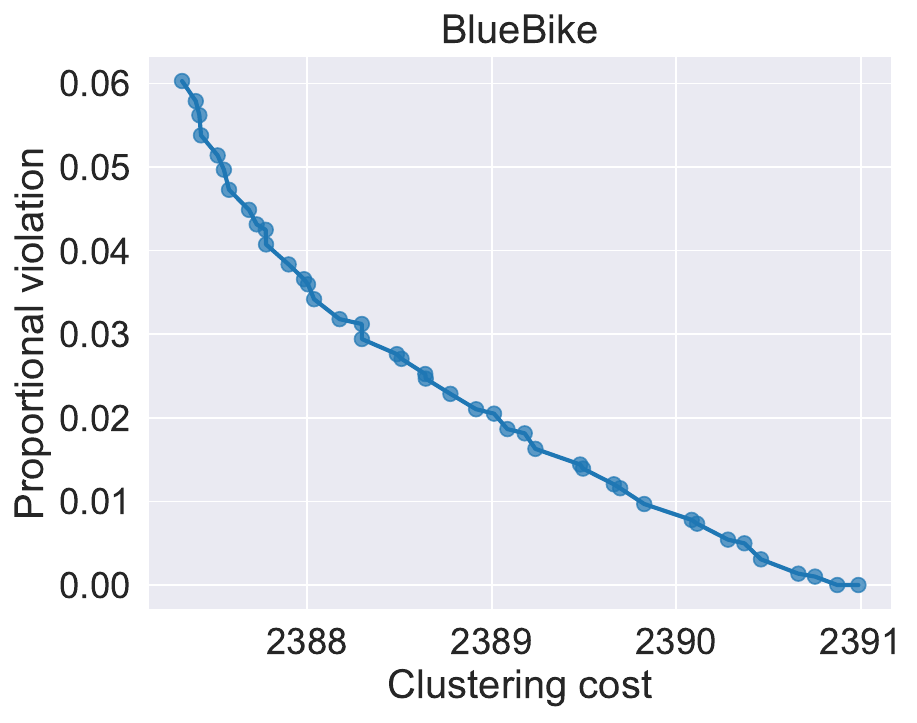}} 
\subfloat[Group Egalit]
{\includegraphics[width=0.2\textwidth] {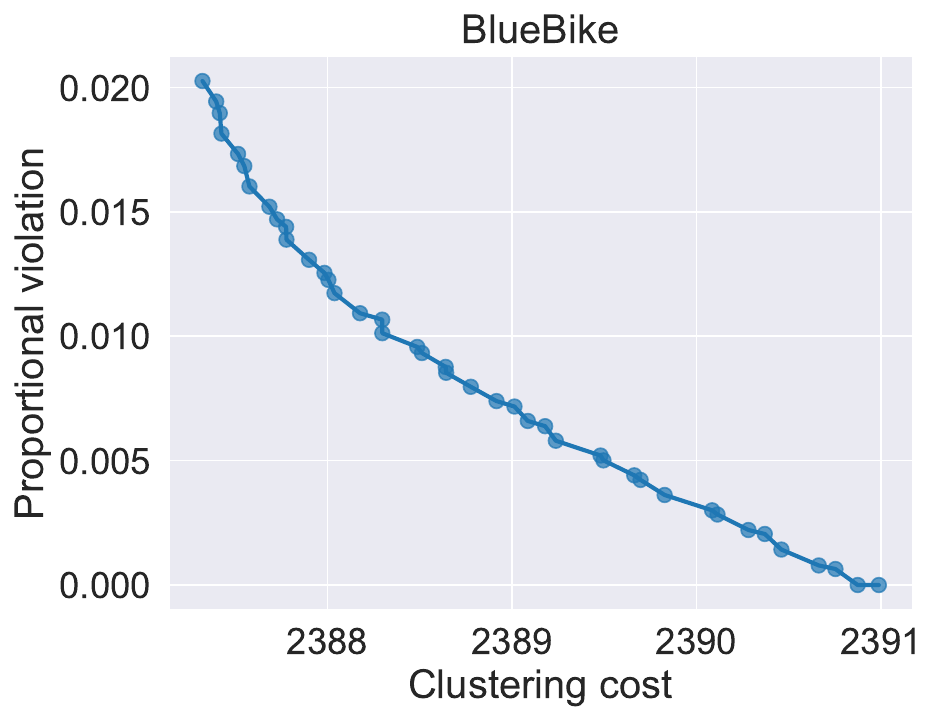}} 
\subfloat[Group Egalit-Sum]
{\includegraphics[width=0.2\textwidth] {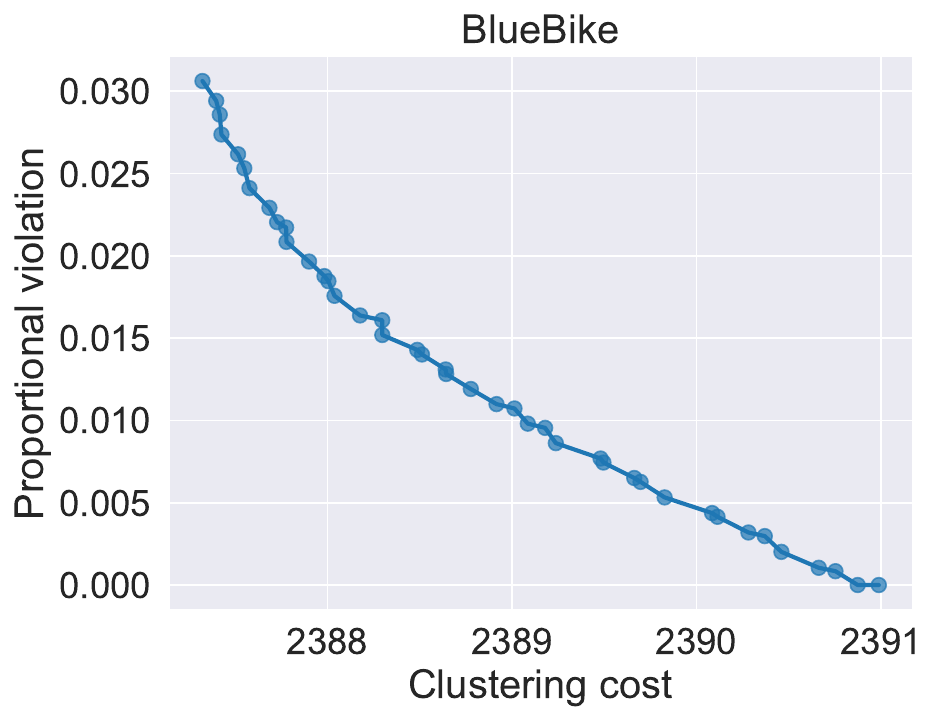}} 
\caption{Pareto front recovered by Algorithm~\ref{alg:dynprogr} for the Adult, Census, and BlueBike datasets (by row), for various fairness objectives (by column), for $k = 2$ clusters.}
\label{fig:alldata_1000_dynprogr}
\end{figure*}

Figure~\ref{fig:alldata_1000_dynprogr} illustrates the Pareto front recovered by our dynamic programming approach (Algorithm~\ref{alg:dynprogr}) on the real-world datasets: the curves obtained are an exact recovery of the Pareto front for the \textit{assignment} problem (as we are not re-computing the clusters centers during the implementation), and thus an approximation for the true Pareto front of the \textit{clustering} problem. We note that the \textsc{Balance} objective and the proportional violation objectives differ: higher balance is considered fairer, whereas lower proportional violation is considered fairer, hence the different shapes of the Pareto fronts. From an evaluation point of view, for each assignment found on the Pareto front, we compute its clustering cost with respect to its \textit{actual} centers, rather than the initial centers found by $k$-means++.~\looseness=-1 

We note that the Pareto front is often, but not always strictly convex or concave, as it simply contains all the undominated points. We note that the proportional violation values will always be worse for the summed objectives than for their min-max equivalents, since the worst proportional violation $\Delta_a^C$ is always non-negative, $\forall a \in [l], C \in \mathcal{C}$.~\looseness=-1

A particular advantage of finding the entire Pareto front is visible for the \textsc{Balance} objective in all datasets: as the clustering cost increases, the gain in the \textsc{Balance} objective becomes negligible; thus, a practitioner wishing to achieve some level of fairness may gain a lot in quality by allowing \textsc{Balance} to decrease by a minimal amount. ~\looseness=-1

\subsection{Exploring faster Pareto Front approximations}
\label{sec:fcbc}

We explore faster algorithms for recovering the Pareto front for the clustering problem. In particular, we leverage a recently proposed linear programming approach that imposes an upper bound on the clustering objective and optimizes a fairness objective, the FCBC algorithm proposed by~\citet{esmaeili2021fair}. For the \textsc{Group Utilitarian} and \textsc{Group Egalitarian} objectives with a clustering cost upper bound input $U$, FCBC presents a polynomial-time approach for finding an approximation of a point $x$ on the Pareto front that has a clustering cost upper bounded by a quantity $(2 + \alpha)U$ with a fairness additive approximation of $\eta$. Here, $\alpha$ is the approximation ratio of a vanilla clustering algorithm for the clustering objective, and $\eta$ is an additive approximation for the fairness objective. Thus, we can extend the $\mathcal{W}$-approximation of a Pareto set definition (Definition~\ref{defn:defn_eps_approx}) to include an additive approximation term: for parameters $\mathcal{W} = ((w_c, v_c), (w_f, v_f))$, we can define the $(\mathcal{W},\mathcal{V})$-approximation of the Pareto set $X_P$ as a set of feasible points $X_P'$ such that 
$\forall x\in X_P,  \exists x'\in X_P'$ such that 
$c(x') \leq w_c\cdot c(x) + v_c$ and $f(x') \leq w_f\cdot f(x) + v_f$. We note that Algorithm~\ref{alg:dynprogr} gives only a multiplicative approximation, so $v_c = v_f = 0$. In contrast, for a point $x \in X_P$ on the true Pareto front, FCBC recovers a $((2 + \alpha, 0), (1, \eta))$-approximation.\footnote{In particular, $\eta$ is dependent on the fairness objective, on U, and on the specific instance of the fair clustering problem (see Theorem 6.1 in~\citet{esmaeili2021fair}). A limitation of their approximation is that $\eta$ is not efficiently computable in closed-form for a particular instance.} 

We extend the FCBC algorithm by allowing a sweep over the clustering cost upper bound $U$, thus, in theory, obtaining an approximation of the Pareto front in polynomial time (for a detailed description, see Algorithm~\ref{alg:repeated_fcbc} in Appendix~\ref{sec:appendix-exp-details}). 

Figure~\ref{fig:adult_300_500_dynprogr_fcbc} shows that doing repeated FCBC (Algorithm~\ref{alg:repeated_fcbc}) recovers few, if any, points on the Pareto front recovered by Algorithm~\ref{alg:dynprogr}: sometimes it recovered the vanilla clustering cost and fairness (the upper most left point in panels b,c,e,f), whereas in other cases it recovers dominated clustering assignments (in panels a-d). We attribute this inaccuracy in recovery to the additive approximation in the fairness objective. Furthermore, whereas both our dynamic programming approach and repeated FCBC have a $(2 + \alpha)$-approximation ratio in the clustering objective, dynamic programming often gets a strictly better cost for similar fairness values. In other words, where repeated FCBC gains in running time,\footnote{The FCBC algorithm uses linear programming through the simplex method, with a worst-case running time of $O(2^n)$; in practice, it is faster than that, as noted in the Appendix.} it loses in recovery accuracy and clustering cost. This is particularly problematic when the only point recovered by FCBC is the vanilla clustering assignment: this means that even when a practitioner may be willing to trade-off significant clustering cost in order to improve fairness, that trade-off is not realizable in practice solely through FCBC. 

Furthermore, the FCBC algorithm does not work for objectives summing over the clusters, such as \textsc{Group Utilitarian-Sum} and \textsc{Group Egalitarian-Sum}. For such objectives, there is no polynomial time algorithm that can recover the Pareto front to within an additive approximation of $O(n^{\delta})$, for $\delta \in [0,1)$ (see Theorem 7.1 in~\citet{esmaeili2021fair}). Our approach, however, can also include such objectives, as we show both in theory and in practice.

\begin{figure*}[!ht]
\centering
\subfloat[Group Util]
{\includegraphics[width=0.33\textwidth] {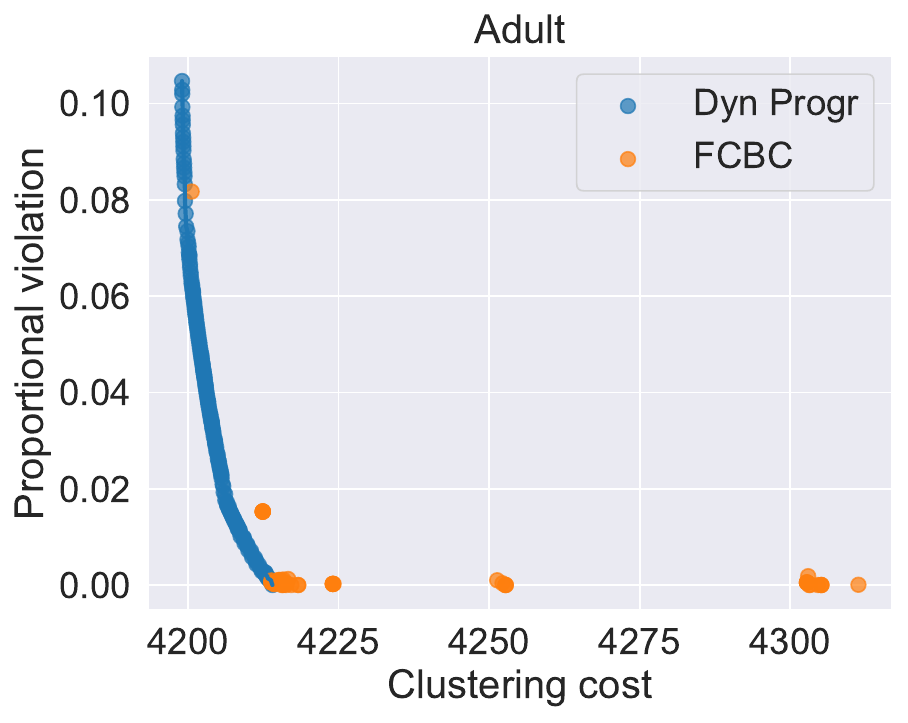}} 
\subfloat[Group Util]
{\includegraphics[width=0.33\textwidth] {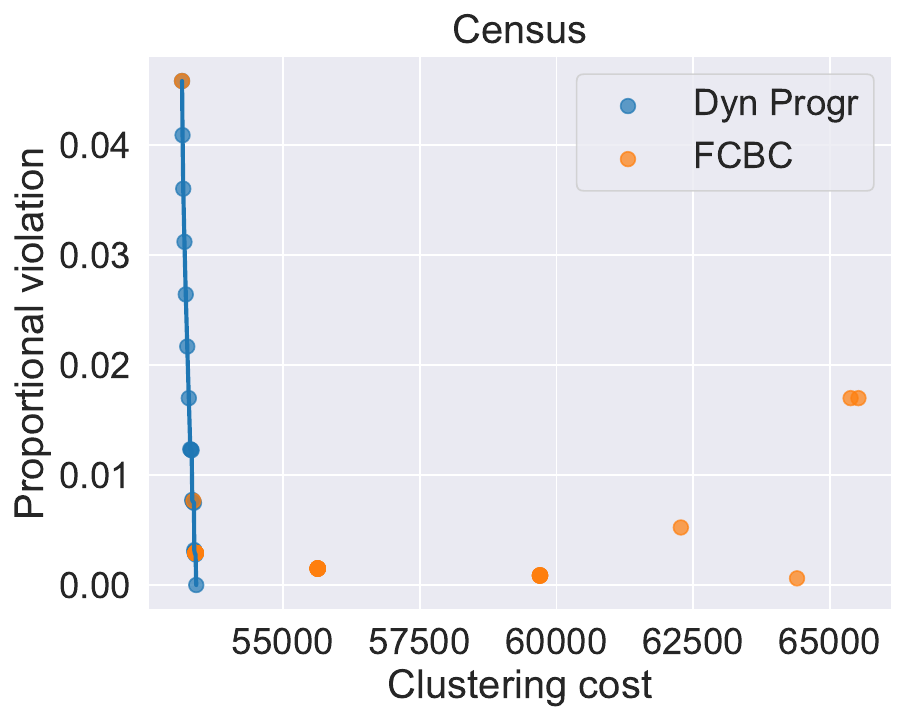}} 
\subfloat[Group Util]
{\includegraphics[width=0.33\textwidth] {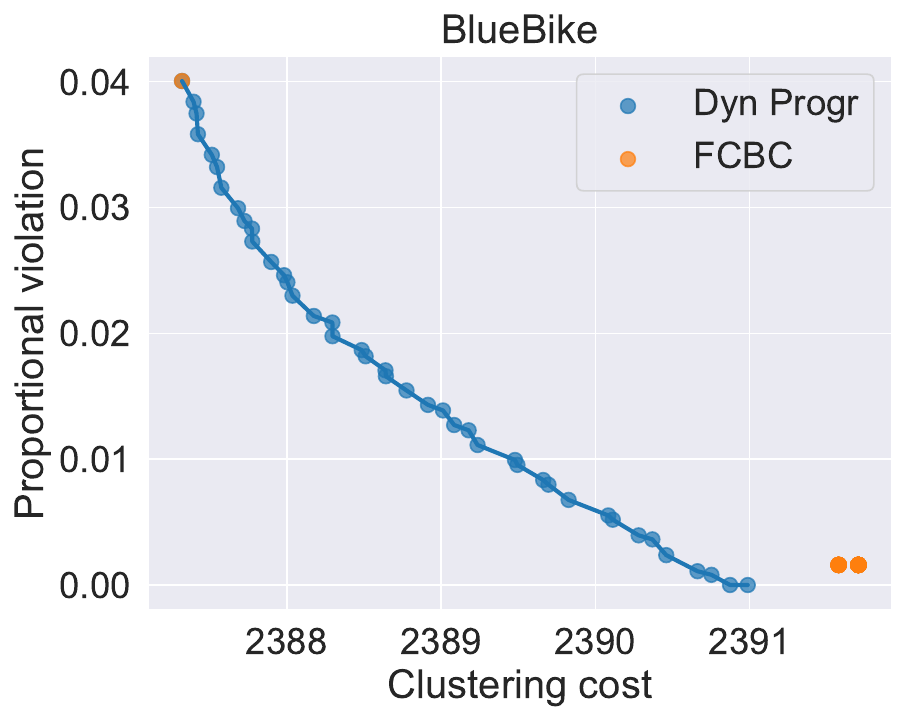}} \\
\subfloat[Group Egalit]
{\includegraphics[width=0.34\textwidth] {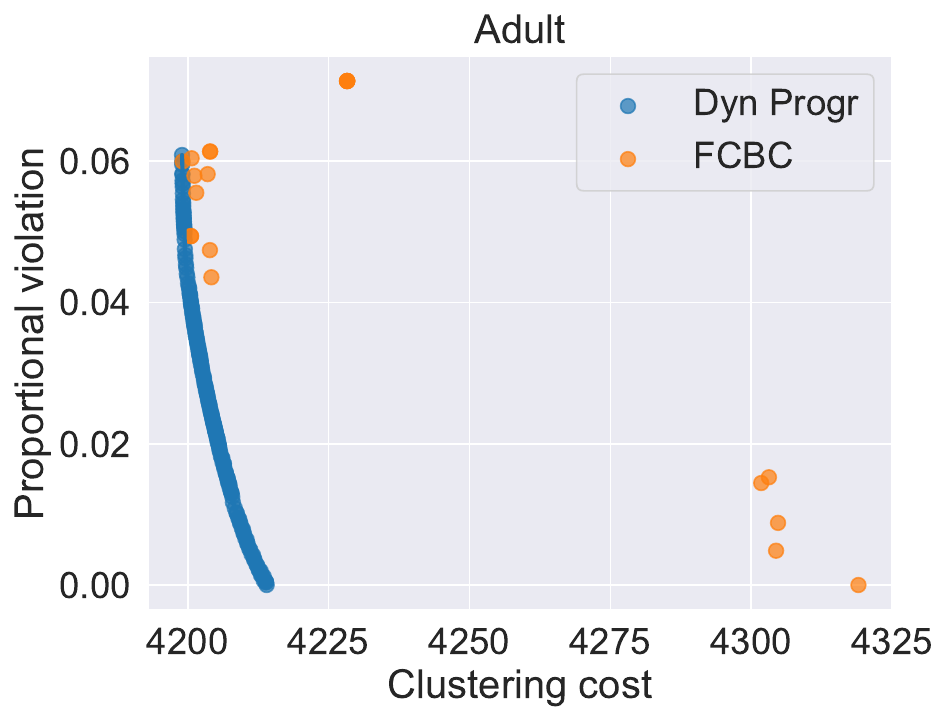}} 
\subfloat[Group Egalit]
{\includegraphics[width=0.33\textwidth] {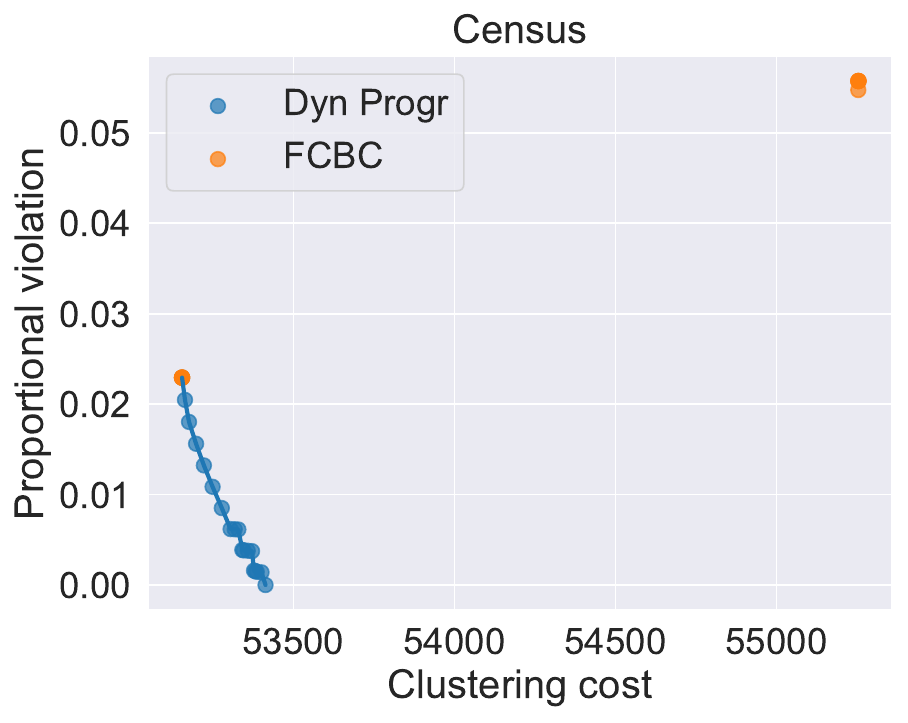}} 
\subfloat[Group Egalit]
{\includegraphics[width=0.34\textwidth] {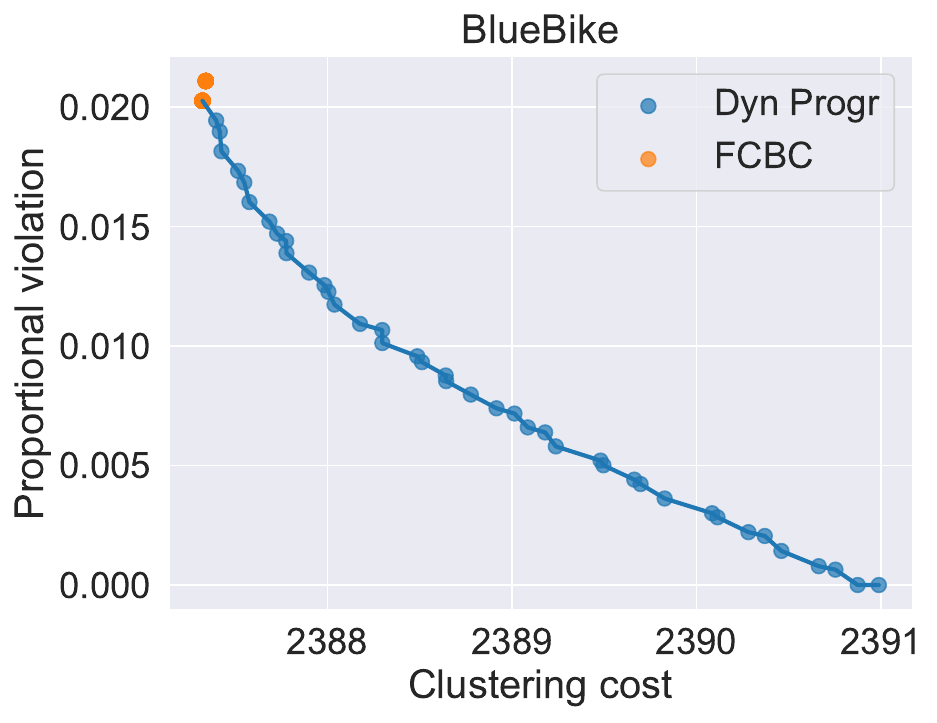}} 
\caption{Pareto front recovered by Algorithm~\ref{alg:dynprogr} (labeled `Dyn Progr', blue) and by Algorithm~\ref{alg:repeated_fcbc} (labeled `FCBC', orange), for $k = 2$ clusters.}
\label{fig:adult_300_500_dynprogr_fcbc}
\end{figure*}

\section{Discussion}
\label{sec:discussion}

Overall, our work shows the versatility of our proposed algorithms for a variety of fairness and clustering objectives. We provide simple properties sufficient in theory for the recovery of the Pareto front for the (clustering, fairness)-biobjective optimization problem, with extensive experiments on real-world datasets. We discuss limitations and future directions of our work in this section.~\looseness=-1

First, while our approach has the advantage of being agnostic to specific objectives, it loses in running time as compared to approaches that optimize for specific fairness objectives. A future direction could study faster algorithms that can recover a similar approximation of the Pareto front. Interesting theoretical open questions emerge: while our analysis ruled out polynomial algorithms for general objectives, are there exponential algorithms (i.e. of the form $O(2^kn^3)$), that still provide an improvement from $O(n^{(k - 1)l})$-type of algorithms? And, are there other objectives that admit polynomial time algorithms for computing an approximate Pareto, aside from \textsc{Sum/Max of Imbalances}? Furthermore, future work could investigate algorithms for recovering the Pareto front for fairness objectives that are not pattern-based, which is a prerequisite for our algorithms. We provide some examples of such objectives in the Appendix. 

Our approach is best suited for group fairness definitions. Extending this work for other definitions, such as individual fairness definitions, would be an excellent avenue for future work. We note that for some individual fairness definitions, the notion of a Pareto front does not apply: for example, the socially fair $k$-means clustering notion proposed by ~\citet{ghadiri2021socially} defines a single optimization objective that already incorporates fairness (by minimizing the max distance between points in each group and their cluster centers). With this definition, the fairness objective and the clustering objective are not separated in a multi-objective optimization problem, but rather combined in a single objective. Therefore, there is no Pareto front, since this notion applies to a multi-objective optimization problem.~\looseness=-1

Finally, we assumed that the sensitive attributes are disjoint. For overlapping attributes, one could assign a new sensitive attribute to every overlap set and apply our approach, however, with a significant increase in running time. Future work could investigate alternative approaches that optimize running time for overlapping attributes.~\looseness=-1

\begin{ack}
The authors would like to thank Vivian Y. Nastl for valuable feedback on earlier drafts of the paper. A.-A. S. acknowledges support from the T\"{u}bingen AI Center. M. Y. acknowledges support from NSF grants CCF-2332922 and CCF-2212233.
\end{ack}

\newpage
\bibliographystyle{plainnat}

\bibliography{main}


\appendix

\include{appendix}

\end{document}

%% file: appendix.tex
\section{Complete Proofs}
\label{sec:appendix-proofs}
\begin{proof}[Proof of Theorem \ref{thm:dynprogr-FA}:] Define $V_P$ as the set of clusterings $\mathcal{C}$ that have the pettern $P^\mathcal{C} = P$ and $\mathcal{C}$ is a clustering of $\mathcal{X}_j$. Let $c$ be the metric-based cost function for the assignment to centers $S$, which is parameterized by a distance function $d$ and non-negative exponent $p$. We will argue that, for all $j$, the clustering $\mathcal{C}$ stored at $T_j[P]$ has the property that $\mathcal{C} = \arg\min_{\mathcal{C}' \in V_{P}} c(\mathcal{C}')$. We will show that this invariant always holds by induction on the number of data points in the current level of the table $j$ (equivalently, the number of points clustered by $\mathcal{C}$ in the main loop of Algorithm~\ref{alg:dynprogr}). The base case is $j=0$, where we store the empty clustering, the unique clustering of $0$ points.

Suppose we have some pattern $P$ for the set of the first $j$ points in our ordering. Let $\mathcal{C}^* = \arg\min_{\mathcal{C} \in V_{P}} c(\mathcal{C})$, so $\mathcal{C}^*$ is optimal cost for $P$. Point $x_j$ must be in some cluster $C_i \in \mathcal{C}^{*}$. The clustering $\mathcal{C}_i'$ of the remaining points (which are exactly the first $j-1$ points from the data) induces some pattern $P_i' = P^{\mathcal{C}_i'}$ which is an $\mathcal{X}_{j-1}$-pattern. Since $\mathcal{C}^{*}$ was assumed to be of optimal quality, the clustering $\mathcal{C}_i'$ must have the property that $\mathcal{C}_i' = \arg\min_{\mathcal{C} \in V_{p_i'}} c(\mathcal{C})$. Otherwise, $\mathcal{C}_i'$ could be replaced in $\mathcal{C}^{*}$ with the lower cost clustering that has the same pattern and reduce the cost of $\mathcal{C}^{*}$ (a contradiction).  

We have shown that $\mathcal{C}^* = \mathcal{C}_i$, where $\mathcal{C}_i = \mathcal{C}'_i + x_j$ is generated by assigning the first $j-1$ points according to $\mathcal{C}'_i$, and then adding point $x_j$ to cluster $i$. By the inductive assumption $T_{j-1}[P_i']$ correctly stores the optimal $\mathcal{C}_i'$. 

Since Algorithm~\ref{alg:dynprogr} minimizes $c(\mathcal{C}_i)$ over cluster indices $i$, we correctly compute the optimal clustering of $j$ points that maps to $P$. All candidate points for the Pareto set of the assignment problem are stored in $T_n$, meaning that every clustering $\mathcal{C}'$ of $\mathcal{X}$ is weakly dominated by some clustering in $T_n$. This is because every clustering $\mathcal{C}\in\mathcal{K}$ must map to some $\mathcal{X}$-pattern $P$. By our algorithm invariant, $\mathcal{C}^* \in T_n[P]$ is the clustering that minimizes assignment cost across clusterings that have pattern $P$, and thus $c(\mathcal{C}^*) \leq c(\mathcal{C})$. In addition, since $P^{\mathcal{C}^*} = P^{\mathcal{C}'}$, and $f$ is a pattern-based fairness objective, $f(\mathcal{C}^*) = f(\mathcal{C})$. So, $\mathcal{C}$ is weakly dominated by $\mathcal{C}^* \in T_n$. Therefore, filtering dominated points out of the clusterings in $T_n$ gives us the complete Pareto set for the assignment problem with pattern-based fairness objectives and metric-based cost objectives on $n$ data points. Note that the output of Algorithm~\ref{alg:dynprogr} recovers the Pareto set $\{\mathcal{C}\}$ as well as the Pareto front $\{(c(\mathcal{C}), f(\mathcal{C})\}$, since the Pareto front is just the image of the Pareto set under the two objectives. 
\end{proof}

\begin{proof}[Proof of Theorem \ref{thm:dynprogr-FC}:] Theorem \ref{thm:dynprogr-FA} states that Algorithm \ref{alg:dynprogr} finds (exactly) the Pareto front of the assignment problem. Let $\phi^*, S^*$ be the optimal cost clustering map and set of centers, respectively, that satisfy a particular fairness bound $F$. Let $\phi, S$ be the clustering map and set of centers found by a vanilla clustering algorithm $\mathcal{A}$, which achieves an $\alpha$-approximation to the best clustering cost. We define a new assignment $\phi', S$, by applying a ``routing'' argument, first introduced in~\citet{bera2019fair} and reused in~\citet{esmaeili2022fair}.

Define a function $\text{nrst}(s^*) = \arg\min_{s\in S} d^p(s, s^*)$ which returns the nearest center in $S$ to an input center $s^*$. Now define an assignment map $\phi'$ where vertices are routed from their center $s^*_i\in S^*$ to $\text{nrst}(s^*_i) \in S$. In other words, for every point $x$, $\phi'(x) = \text{nrst}(\phi^*(x))$. Now we can write:
\begin{align*}
d^p(x, \phi'(x)) = d^p(x, \text{nrst}(\phi^*(x))) &\leq d^p(x, \phi^*(x)) + d^p(\phi^*(x), \text{nrst}(\phi^*(x)))\\
&\leq d^p(x, \phi^*(x)) + d^p(\phi^*(x), \phi(x))\\
&\leq 2d^p(x, \phi^*(x)) + d^p(x, \phi(x))
\end{align*}
The first and third inequalities follow from the triangle inequality on $d^p$ and the second inequality is due to the definition of the $\text{nrst}$ function. In addition, $\alpha \left(\sum_x d^p(x, \phi^*(x))\right)^{1/p} \geq \left(\sum_x d^p(x, \phi(x))\right)^{1/p}$, since $\phi$ is the clustering found by the vanilla clustering algorithm $\mathcal{A}$. Applying our previous inequality together with the triangle inequality on the $p$-norm:
\begin{align*}
\left(\sum_x d^p(x, \phi'(x)\right)^{1/p} &\leq \left(\sum_x 2d^p(x, \phi^*(x))\right)^{1/p} + \left(\sum_x d^p(x, \phi(x))\right)^{1/p}\\
&\leq (2 + \alpha) \left(\sum_x d^p(x, \phi^*(x))\right)^{1/p}
\end{align*}
This implies that $\phi'S$ is an assignment map with respect to centers $S$ that has at most $(2+\alpha)$ times the cost of $\phi^*$ with respect to centers $S^*$. 

In addition, the clustering $\mathcal{C'}$ associated with $\phi'$ can be generated by, for each $s_j \in S$, merging all clusters $C^*_i \in \mathcal{C^*}$ such that $\text{nrst}(s^*_i) = s_j$. This procedure can be done by sequentially merging pairs of clusters. Since the fairness objective is mergeable (see Def. \ref{defn:mergeability}), this implies that $f(\mathcal{C}^*) \geq f(\mathcal{C'})$. So, the assignment map $\phi'$ with respect to centers $S$ also satisfies the fairness bound $F$. There exists an assignment to centers $S$ that is a $(2 + \alpha, 1)$-approximation to $\phi^*, S^*$, so the Pareto front of the assignment problem is a $(2 + \alpha, 1)$-approximation to the Pareto front of the clustering problem as desired.~\looseness=-1
\end{proof}

\begin{proof}[Proof of Theorem~\ref{thm:nonmerge-FC}]
    We describe in detail the algorithm modification needed to compute the Pareto front for non-mergeable fairness objectives.  First, we give some necessary preliminaries: 
 
\begin{definition}[Refinement of a pattern and refinement DAG $D$] 
Consider the directed graph $D$ with the $\mathcal{X}$-patterns as nodes and edges $(P_1,P_2)$ if merging two nonempty rows of the pattern $P_2$ yields the pattern $P_1$. A pattern $P'$ is a refinement of another pattern $P$ iff there is a path from $P$ to $P'$ in $D$, i.e. if $P$ can be obtained from $P'$ by merging different parts of $P'$. Note that $D$ is a directed acyclic graph (DAG). Let $R_P$ be the set of $P'$ that are refinements of $P$.
\end{definition}

Similarly, one can define the refinement of a clustering $\cal C$. Note that if a clustering $\mathcal{C'}$ is a refinement of another clustering $\mathcal{C}$, then its pattern $P^{\mathcal{C'}}$ is a refinement of the pattern of cluster $\cal C$, $P^{\mathcal{C}}$. 

Next, we define a modified fairness function created from the non-mergeable function $f$, with the purpose of reducing it to a mergeable function and applying Algorithm~\ref{alg:dynprogr}. 

\begin{definition}[Modified fairness function] 
If $f$ is a pattern-based fairness function,
define its associated modified function $\hat{f}$ to
be $\hat{f}(P) = \min_{P'\in R_P} (f(P'))$ for every pattern $P$.
\end{definition}

Note that $f$ is still required to be pattern-based, and thus modified function is well-defined. By definition, $\hat{f}$ is a mergeable function. Furthermore, if $f$ is mergeable then $\hat{f}=f$.

\begin{lemma}\label{lem:modfunction}
We can compute in $O(n^{l(k-1)})$ time the modified function $\hat{f}$ for all $\mathcal{X}$-patterns $P$, and compute for each $P$ a refinement $P'$ such that $\hat{f}(P) = f(P')$. 
\end{lemma}
\begin{proof}
We compute $\hat{f}$ bottom up in the DAG $D$. 
We also compute for each node a pointer to its refinement pattern that has the minimum fairness cost.
At the sinks $P$ (patterns that have no outward edge pointing from them to other patterns), $\hat{f}(P) =f(P)$ and $P$ points to itself only. 
For every non-sink node $P$, we set $\hat{f}(P)= \min_{v \in N(P)} \hat{f}(v)$ where $N(P)$ is the set of neighbors of $P$,
and we set the pointer of $P$ to the descendant pointed by the
neighbor $v$ of $P$ with the minimum $\hat{f}(v)$.

The total time complexity is linear in the number of nodes and edges of $D$.  The number of edges is at most $k^2$ times the number of nodes, since every node $P'$ has at most $k^2$ incoming edges (there are at most $k^2$ choices for the two parts of $P'$ that are merged to form a parent pattern).
The number of nodes of $D$, i.e.  $\mathcal{X}$-patterns, is at most $4k(\frac{n}{2})^{l(k-1)}$:
All the components of all the rows of a pattern $P$, except possibly at most two components, are $< n/2$; furthermore, specifying $k-1$ rows of $P$ determines also the last row
because the sum for each attribute value must match the given set of points. From the bounds on the number of nodes and edges of $D$,
it follows that the time complexity is $O(n^{l(k-1)})$.
\end{proof}

Recall that in a clustering we associate a center with each cluster, and the cost of a clustering is computed from the
distances of the points from the center of their cluster.
We allow different clusters to have the same point as their center.
We show the following result, needed in the proof of Theorem~\ref{thm:nonmerge-FC}: 

\begin{lemma}\label{lem:modcluster}
If we are given a clustering $\mathcal{C}$ with centers $S$ and $P'$ is a refinement of $P^{\mathcal{C}}$ then we can compute efficiently a clustering $\mathcal{C}'$ with centers $S'$, which we call a center reassignment of $\mathcal{C}$, such that $P^{\mathcal{C}'} = P'$ and $c(\mathcal{C}', S') = c(\mathcal{C}, S)$. The centers $S'$ will be a subset of the points in $S$ but with multiplicity
(i.e. different clusters may have the same center).
\label{lemma:refinement}
\end{lemma}
\begin{proof}
Suppose that row $i$ of pattern $P$ is formed by merging a set $J_i$ of two or more nonempty parts of $P'$.
Then we split the cluster $C_i$ of $\mathcal{C}$ into a set 
of $|J_i|$  subclusters, one for each part in $J_i$,
and we place in each subcluster a number of points from $C_i$ for each attribute value that matches the corresponding entry in the row of $P'$; we assign to all the subclusters the same center as the center of $C_i$.

After performing the above splitting for all parts $i$ of $P$
that are refined in the pattern $P'$, we obtain a clustering
 $\mathcal{C}'$ such that $P^{\mathcal{C}'} = P'$.
Since the subclusters created from splitting a cluster $C_i$ of $\mathcal{C}$ are assigned the same center as the center of $C_i$, it follows from the definition of a metric-based cost function, that $\mathcal{C}$ and $\mathcal{C'}$ have the same cost.

\end{proof}

We now describe the algorithm modification for non-mergeable fairness objectives. For a non-mergeable fairness objective $f$, let $\hat{f} = \min_{P'\in R_P} (f(P'))$ be its associated modified function. 
Apply the algorithm of Lemma ~\ref{lem:modfunction} to compute
$\hat{f}(P)$ for every $\mathcal{X}$-pattern $P$, and the corresponding pointer to its optimal refinement $P'$.
As before, use a vanilla clustering approximation algorithm $\mathcal{A}$ to compute a clustering that approximates the minimum cost. Use the centers of this clustering in Algorithm \ref{alg:dynprogr} to construct the dynamic programming table $T_n$. 
Process the patterns $P$ as before in order of increasing cost, but now use the modified function $\hat{f}$ as the fairness objective to filter out
dominated patterns, and for each undominated pattern $P$, replace the clustering $\mathcal{C}$ in $T_n[P]$ by the center reassignment clustering $\mathcal{C'}$ constructed as in Lemma \ref{lem:modcluster}.
The algorithm returns the set of these center reassignment clusterings $\mathcal{C'}$ for all
undominated patterns. By the definition of the modified fairness function and Lemma~\ref{lem:modcluster}, these are undominated clusterings for the original (non-mergeable) fairness function $f$ and the clustering cost $c$. 

From our timing analysis, the time complexity of the algorithm
is $O(kn^{l(k-1)})$.

The proof of the approximation follows the proof of Theorem~\ref{thm:dynprogr-FC}. We define $\phi^*, S^*$, $\phi, S$, and $\phi', S$ similarly as before. Observe that in the construction of $\phi'$, we simply merged clusters from $\phi^*$ and gave them new centers. Since the pattern of a clustering is independent of the identities of the centers, the pattern of $\phi^*$ is a refinement of the pattern of $\phi'$. So by Lemma \ref{lem:modcluster}, there exists a clustering $\phi''$ that is a center reassignment of $\phi'$ and has the same pattern as $\phi^*$. 

Therefore, $\phi''$, which is one of the reassignments we search over in the algorithm, has $f(\mathcal{C}'') = f(\phi^*)$ and $c(\mathcal{C}'') = c(\phi')$ (here, we used interchangeably the clustering cost function $c$ applied to the clustering or the clustering assignment map, which are equivalent). Since $\phi'$ has at most $2 + \alpha$ worse cost than $\phi^*$ (see proof of Theorem~\ref{thm:dynprogr-FC}), so does $\mathcal{C}''$. Therefore, the Pareto front computed by the algorithm is a $(2 + \alpha, 1)$-approximation to the Pareto front of the clustering problem, as desired.
\end{proof}

\begin{proof}[Proof of Theorem~\ref{thm:matching}:]
First we notice that in this case $f$ can take at most $n+1$ values: If $n = |\cal X|$ is even, then it can take only values the even integers between 0 and $n$, and if it is odd all odd integers between $1$ and $n-1$.  We shall treat the case of even $n$, the case of even $n$ being very similar. 

Given dataset $\cal X$, metric $d$, and exponent $p$, for each even number $F$ between $0$ and $n$ we must compute the best assignment that has fairness $F$.  Once this is done, we only have to sort with respect to fairness and remove the dominated assignments (in a similar vein to the filtering heuristic of Algorithm~\ref{alg:dynprogr}). Clearly, this step is polynomial time since we can sort the points in $O(n\log(n))$ and then make a single pass over them to remove the dominated points. All that remains to show is that we can compute the best assignment with fairness $F$ also in polynomial time. We do so as follows.

Given an even number $F$, we construct a weighted graph $G_F$ with $n$ nodes corresponding to the data points, plus $F$ {\em dummy nodes.}  We join every data point $x$ with every dummy node $z$ by an edge of weight $\min_{i \in [k]} d^p(x,s_i)$.  We join any two data points $x,y$ with different attribute value (recall that there are only two attribute values) by an edge of length $\min_{i=1}^k (d^p(x,s_i)+d^p(y,s_i))$.  We call an edge of $G_F$ to be {\em of type $i\in [k]$} if the $i$-th cluster achieves the minimum that defines the weight of the edge. (In other words, $i$ is the $\mathrm{argmin}$ of the minimization expression.)

A perfect matching in a graph is a set of disjoint edges that includes all the nodes.
The weight of a matching $M$ is defined as the sum of the weights of edges that are contained in the matching: $w(M) = \sum_{e \in M} w(e)$.
Compute the minimum weight perfect matching $\hat{M}$ in the graph $G_F$. This can be done in polynomial time using Edmonds' algorithm \cite{edmonds1965matching},
specifically in time $O(N(E+N \log N))$ for a graph with $N$ nodes and $E$ edges \cite{Gabow18data}; in our case the graph $G_F$ has at most $2n$ nodes and $2n^2$ edges.~\looseness=-1

We will show that the minimum weight of a perfect matching is equal to the minimum cost of a clustering with fairness $F$,
and furthermore we can derive a minimum cost clustering from the minimum weight matching $\hat{M}$.

Consider any clustering $\cal C$ with $f({\cal C})=F$. Starting from $\cal C$, we construct a perfect matching $M$ of $G_F$ as follows: For each cluster $C \in \mathcal{C}$, choose the attribute value $a$ that has fewer nodes in $C$ (breaking ties arbitrarily). Now, match these nodes of attribute $a$ in $C$ arbitrarily with nodes of the larger attribute group in $C$. Finally, match any remaining nodes of the other group (the larger attribute group) to dummy nodes. 

We claim that such a matching $M$ is possible, because the total number of nodes that cannot be matched is precisely $F$, the number of dummy nodes.  We claim now that the weight of this matching $w(M)$ is at most the cost of the clustering, $c({\cal C})$; that is, $w(M) \leq c({\cal C}).$  

The weight of the matching is at most the clustering cost for the following reason: each matched pair $x,y$ contributes to $c(\cal C)$ at least the weight of the corresponding matched edge; and any data point matched to a dummy node again contributes to $c(\cal C)$ at least the weight of the matched edge. 

Now consider the minimum weight perfect matching $\hat M$ of the weighted graph $G_F$. Clearly, $w(\hat M) \leq c(\cal C)$. 
Construct from $\hat M$ a corresponding clustering $\hat{\mathcal{C}}$ as follows: any matched data point whose edge is of type $i$ is placed in cluster $i$, while any data point matched in $\hat M$ with a dummy node with an edge of type $i$ is added to cluster $i$.
This clustering satisfies $c(\hat{\mathcal{C}}) = w(\hat M) \leq c(\mathcal{C})$.  Since $\cal C$ was assumed to be an arbitrary clustering with $f(\mathcal{C})=F$, $\hat{\mathcal{C}}$ is the optimum such clustering.\end{proof}

The same result can be shown for the \textsc{Max Imbalance} objective.

\begin{theorem} \label{thm:maximbalance}
If $l=2$ and the fairness objective $f$ is the max imbalance $f({\cal C})=\max_{i\in [k]} | |C_i^1| -|C_i^2| |$,  then the Pareto front of the assignment problem can be computed in polynomial time.  
\end{theorem}
\begin{proof}
The objective can take again at most $n+1$ values (more precisely, at most
$1+\max( | \mathcal{X}^1|, |\mathcal{X}^2|)$ values). For each possible value $F$,  construct a weighted graph $G_F$, whose set $N_F$ of nodes consists of $n$ nodes corresponding to the given set $\mathcal{X}$ of data points, a set $D_i$ of $F$ dummy nodes for each $i\in [k]$, and if $n+kF$ is odd, then $N_F$ has one more dummy node so that the total number of nodes is even.
The edge set $E_F$ of $G_F$ consists of the following edges: for each pair $x,y$ of data points with different attribute values, we include an edge $(x,y)$ with weight $\min_{i=1}^k (d^p(x,s_i)+d^p(y,s_i))$ and associate with the edge as its type the index $i \in [k]$ that achieves the minimum in the weight;
for each data point $x$ and dummy node $z$ in $D_i$, we include an edge $(x,z)$ with weight $d^p(x,s_i)$ and associate type $i$ to the edge; for every pair $z,w$ of dummy nodes we include an edge $(z,w)$ with weight 0 (we do not associate a type with these edges). 

We then compute the minimum weight perfect matching $\hat{M}$ of the graph $G_F$.
Just like in the proof of Theorem~\ref{thm:matching}, the minimum weight perfect matching can be found in polynomial time.
The matching $\hat{M}$ induces an assignment $\hat{\mathcal{C}}$ of the data points: every point $x$ is assigned to the cluster $i$ corresponding to the type of the edge of $\hat{M}$ incident to $x$.
By construction, the cost of the assignment is equal to the weight $w(\hat{M})$ of the matching. Furthermore, for every cluster $i \in [k]$, the number of data points in the cluster that are matched with dummy nodes in $D_i$ is at most $F$,
hence $| |\hat{C}_i^1| - |\hat{C}_i^2| | \leq F$.
Therefore the fairness cost of the assignment is at most $F$.

Conversely, any assignment $\mathcal{C}$ with fairness $F$ induces a perfect matching $M$ of $G_F$, as follows:
For each cluster $i$, match data points of cluster $C_i$ in $\mathcal{C}$ with opposite attribute values arbitrarily in pairs, and match the remaining $| |C_i^1| - |C_i^2| |$ data points to dummy nodes in $D_i$. The rest of the dummy nodes that are not matched to a data point are matched arbitrarily in pairs.
The weight of this matching $M$ is then at most the cost of the clustering $\mathcal{C}$, $w(M) \leq c(\mathcal{C})$. 
Therefore, $c(\mathcal{C}) \geq w(\hat{M}) = c(\hat{\mathcal{C}})$, that is, any assignment $\cal C$ with fairness $F$ is dominated by the assignment $\hat{\mathcal{C}}$ that we derived from the minimum weight perfect matching.
\end{proof}

\section{Analyzing Fairness Objectives}
\label{sec:appendix-analyzingfairnessobj}

\subsection{Balance and proportional violation-based objectives are pattern-based and mergeable}
\label{sec:appendix-fairnessproofs}
\begin{proposition}
    The fairness objectives \textsc{Balance}, \textsc{Sum of Imbalances}, \textsc{Group Utilitarian}, \textsc{Group Utilitarian-Sum}, \textsc{Group Egalitarian}, and \textsc{Group Egalitarian-Sum} are pattern-based and mergeable. 
    \label{prop:5objpatternbasedandmergeable}
\end{proposition}
\begin{proof}
    
\emph{Balance-based Objectives:} As defined in equation~\ref{eq:balance}, the \textsc{Balance} objective is always pattern-based by definition. As above, say that we have two sensitive attributes  (as this objective is originally defined for $ l =2$), called $R$ and $B$. Then, for a cluster $C \in \mathcal{C}$, $|C^R|$ and $|C^B|$ are the number of $R$ and $B$ data points in cluster $C$, respectively. For two clusterings $\mathcal{C}$ and $\mathcal{C}'$ that induce the same pattern $p$, it means that we can pair each $C\in \mathcal{C}$ with a different cluster $C' \in \mathcal{C}'$ such that $|C^R| = |C'^R|$ and $|C^B| = |C'^B|$. Thus $\textsc{Balance}(C) = \textsc{Balance}(C')$ for all such pairs, and thus $\textsc{Balance}(\mathcal{C}) = \textsc{Balance}(\mathcal{C})'$. 

We show that \textsc{Balance} is also mergeable through a simple induction over the number of clusters to be merged. For the base case, take a clustering $\mathcal{C}$ and construct a clustering $\mathcal{C}'$ in which two clusters from $\mathcal{C}$ have been merged (assume without loss of generality that $\mathcal{C} = \{C_1, C_2, \cdots, C_k \}$ and $\mathcal{C}' = \{\emptyset, C_1 \cup C_2, \cdots, C_k \}$). For ease of notation, denote by $a = |C_1^R|, b = |C_1^B|, c = |C_2^R|, d = |C_2^B|$. Then, $a + c = |(C_1 \cup C_2)^R|, b + d = |(C_1 \cup C_2)^B|$. Note that all variables are non-negative. We assume that neither $C_1$ and $C_2$ are empty. We need to show that the fairness objective is at least as good for the merged clustering than for the original clustering. It is sufficient to show that 

\begin{equation}
    \min \left( \min \left(\frac{a}{b}, \frac{b}{a} \right), \min \left(\frac{c}{d}, \frac{d}{c} \right) \right) \leq \min \left( \frac{a + c}{b  + d}, \frac{b + d}{a + c}\right)
    \label{eq:prop-balance}
\end{equation}

First, this property is true as we can easily prove by considering possible cases: 

\begin{enumerate}
    \item If $a \leq b$ and $c \leq d \Rightarrow \frac{a}{b} = \min \left(\frac{a}{b}, \frac{b}{a} \right), \frac{c}{d} = \min \left(\frac{c}{d}, \frac{d}{c} \right), \frac{a + c}{b + d} = \min \left( \frac{a + c}{b  + d}, \frac{b + d}{a + c}\right).$ Without loss of generality $\frac{a}{b} = \min \left( \frac{a}{b}, \frac{c}{d}\right)$. Then, it is easy to see that $\frac{a}{b} \leq \frac{a + c}{b + d} \Leftrightarrow ab + ad \leq ab + bc \Leftrightarrow ad \leq bc \Leftrightarrow \frac{a}{b} \leq \frac{c}{d}$. 
    \item If $a \geq b$ and $c \geq d \Rightarrow \frac{b}{a} = \min \left(\frac{a}{b}, \frac{b}{a} \right), \frac{d}{c} = \min \left(\frac{c}{d}, \frac{d}{c} \right), \frac{b + d}{a + c} = \min \left( \frac{a + c}{b  + d}, \frac{b + d}{a + c}\right).$ Then, the argument from the first case follows identically. 
    \item If $a \leq b$ and $c \geq d$, then we need to show that $\min \left(\frac{a}{b},\frac{d}{c} \right) \leq \min \left( \frac{a + c}{b + d}, \frac{b + d}{a + c}\right)$. Without loss of generality, $\frac{a + c}{b + d} = \min \left( \frac{a + c}{b + d}, \frac{b + d}{a + c}\right)$. Now we have two cases: 
    \begin{itemize}
        \item If $\frac{a}{b} \leq \frac{d}{c}$, then $\frac{a}{b} \leq \frac{a + c}{b + d} \Leftrightarrow ab + ad \leq ab + ac \Leftrightarrow ad \leq bc \Leftrightarrow \frac{a}{b} \leq \frac{c}{d}$, which is true since $\frac{a}{b} \leq \frac{d}{c} \leq \frac{c}{d}$. 
        \item If $\frac{a}{b} \geq \frac{d}{c}$, then $\frac{d}{c} \leq \frac{a + c}{b + d} \Leftrightarrow bd + d^2 \leq ac + c^2$, which is true since $bd \leq ac \Leftrightarrow \frac{a}{b} \geq \frac{d}{c} $ and $d^2 \leq c^2 \Leftrightarrow d \leq c$. 
    \end{itemize}
    \item If $a \geq b$ and $c \leq d$, then the proof follows identically to case $3$. 
\end{enumerate}

Then, if equation~\ref{eq:prop-balance} holds, then $\textsc{Balance}(\mathcal{C}) \leq \textsc{Balance}(\mathcal{C}')$. To see this, we there is a cluster index $i > 2$ for which $\textsc{Balance}(\mathcal{C}) = \textsc{Balance}(C_i)$. Then, $\textsc{Balance}(C_i) \leq \min \left( \min \left(\frac{a}{b}, \frac{b}{a} \right), \min \left(\frac{c}{d}, \frac{d}{c} \right) \right)$, so $\textsc{Balance}(\mathcal{C}') = \textsc{Balance}(C_i) \Rightarrow \textsc{Balance}(\mathcal{C}) = \textsc{Balance}(\mathcal{C}')$. If there is no such cluster $i > 2$, then $\textsc{Balance}(\mathcal{C}) = \min \left( \min \left(\frac{a}{b}, \frac{b}{a} \right), \min \left(\frac{c}{d}, \frac{d}{c} \right) \right) \leq \textsc{Balance}(C_i), \forall i > 2$. Then, it follows that $\textsc{Balance}(\mathcal{C}) \leq  \textsc{Balance}(\mathcal{C}')$. 

The \textsc{Sum of Imbalances} and \textsc{Max Imbalance} objectives are derived from the \textsc{Balance} objective. As they are only a function of the number of data points of different attributes in each cluster, they are also clearly pattern-based. To see that \textsc{Sum of Imbalances} is also mergeable, we consider again the base case of the induction proof, for $2$ clusters. For the \textsc{Sum of Imbalances} objective, note that $|C_i^1 - C_i^2| + |C_i^1 - C_i^2| \geq |C_i^1 - C_i^2 + C_i^1 - C_i^2|$ by the triangle inquality for any clusters $C_i, C_j$. In merging two clusters $C_i$ and $C_j$ from a clustering $\cal C$, obtaining a clustering $\cal C'$, the contribution to the objective of $C_i \cup C_j$ and the empty cluster is exactly $|C_i^1 - C_i^2 + C_i^1 - C_i^2|$ (an empty cluster has an imbalance of $0$). Since all other clusters remained unchanged, we conclude that $\textsc{Sum of Imbalances}(\mathcal{C}) \geq \textsc{Sum of Imbalances}(\mathcal{C}')$, and thus, merging two clusters can only improve the objective. 

\begin{remark}
    As a note, the \textsc{Max Imbalance} objective is not mergeable. As a simple objective, take the follwing clustering $\cal C$: each of the clusters $C_1$ and $C_2$ has one data point of attribute $1$ and two data points of attribute $2$. Then, $\textsc{Max Imbalance}(\mathcal{C}) = 1$. The clustering $\cal C'$ obtained by merging clusters $C_1$ and $C_2$ has $\textsc{Max Imbalance}(\mathcal{C}') = 2$.~\looseness=-1
\end{remark}

Finally, for the induction step, if mergeability holds for merging a set of $w$ clusters, then it holds for merging a set of $w + 1$ clusters as well, by reducing to the base case: without loss of generality, denote the $w + 1$ clusters to be merged by $C_1, \cdots, C_{w + 1}$. By the induction hypothesis, mergeability holds for merging any $w$ clusters, so for $\cup_{j \in [w]} C_j$. Then, the base case applies for the clusters $\cup_{j \in [w]} C_j$ and $C_{w + 1}$.~\looseness=-1

\emph{Proportional violation objectives:} First, we easily note that for any two clusterings $\mathcal{C}$ and $\mathcal{C}'$ that induce the same pattern  $\Delta_a^C$, we pair each cluster $C \in \mathcal{C}$ with a different cluster $C' \in \mathcal{C}'$ such that $|C| = |C'|$ and $|C^a| = |C'^a|$ for all attributes $a \in [l]$. Thus, from equation~\ref{eq:delta} it follows that $\Delta_a^C = \Delta_a^{C'}, \forall a \in [l]$. Since all proportional violation-based objectives defined are only a function of $(\Delta_a^C)_{a, C}$, it follows that all clusterings that have the same pattern will also have the same objective value for all four of the proportional violation-based objectives. 

Finally, we will show that they are also mergeable. We show this again by induction over the number of merged clusters for each of the objectives. We start with the base case of two clusters, denoted without loss of generality by $C_1$ and $C_2$. We assume that neither $C_1$ and $C_2$ are empty. We say that $C_1$ and $C_2$ are part of a clustering $\mathcal{C}$, and we aim to show that the clustering $\mathcal{C}'$ in which $C_1$ and $C_2$ got merged will have $\textsc{Objective}(\mathcal{C}') \leq \textsc{Objective}(\mathcal{C})$, for \textsc{Objective} is one of the \textsc{Group Utilitarian}, \textsc{Group Utilitarian-Sum}, \textsc{Group Egalitarian}, \textsc{Group Egalitarian-Sum} objectives. For a sensitive attribute $a\in [l]$, assume without loss of generality that $\frac{|C_1^a|}{|C_1|} \leq \frac{|C_2^a|}{|C_2|}$. Then, the following property holds, also known as the \textit{mediant inequality}: 

\begin{equation}
    \frac{|C_1^a|}{|C_1|} \leq \frac{|C_1^a| + |C_2^a|}{|C_1| + |C_2|} \leq \frac{|C_2^a|}{|C_2|}
\end{equation}

To see this, note that the left hand side is equivalent to: 

\begin{equation}
\begin{gathered}
    \frac{|C_1^a|}{|C_1|} \leq \frac{|C_1^a| + |C_2^a|}{|C_1| + |C_2|} \Leftrightarrow |C_1^a| \cdot |C_1| + |C_1^a| \cdot |C_2| \leq |C_1^a| \cdot |C_1| + |C_1| \cdot |C_2^a| \Leftrightarrow \\
     |C_1^a| \cdot |C_2| \leq |C_1| \cdot |C_2^a| \Leftrightarrow \frac{|C_1^a|}{|C_1|} \leq \frac{|C_2^a|}{|C_2|}
\end{gathered}
\end{equation}

The right hand side is equivalent to: 

\begin{equation}
    \begin{gathered}
        \frac{|C_1^a| + |C_2^a|}{|C_1| + |C_2|} \leq \frac{|C_2^a|}{|C_2|} \Leftrightarrow |C_1^a| \cdot |C_2| + |C_2^a| \cdot |C_2| \leq |C_1| \cdot |C_2^a| + |C_2| \cdot |C_2^a| \Leftrightarrow \\
        |C_1^a| \cdot |C_2| \leq |C_1| \cdot |C_2^a| \Leftrightarrow \frac{|C_1^a|}{|C_1|} \leq \frac{|C_2^a|}{|C_2|}
    \end{gathered}
\end{equation}

By the definition from equation~\ref{eq:delta}, we have 

\begin{equation}
\begin{gathered}    
    \beta_a - \Delta_a^{C_1} \leq \frac{|C_1^a|}{|C_1|} \leq \alpha_a + \Delta_a^{C_1}, \\ 
    \beta_a - \Delta_a^{C_2} \leq \frac{|C_2^a|}{|C_2|} \leq \alpha_a + \Delta_a^{C_2},
\end{gathered}
\label{eq:deltac1c2}
\end{equation}
for clusters $C_1, C_2 \in \mathcal{C}$. As $C_1$ and $C_2$ got merged in $\mathcal{C'}$, they got replaced by the  empty cluster $C_{\emptyset}$ and the merged cluster $C_1 \cup C_2$ in $\mathcal{C}'$. We note that $\Delta_a^{C_{\emptyset}} = \beta_a$, while $\Delta_a^{C_1 \cup C_2}$ is the minimum value that satisfies

\begin{equation}
    \beta_a - \Delta_a^{C_1 \cup C_2} \leq \frac{|C_1^a| + |C_2^a|}{|C_1| + |C_2|} \leq \alpha_a + \Delta_a^{C_1 \cup C_2}
    \label{eq:deltac1cupc2}
\end{equation}

Using inequalities~\ref{eq:deltac1c2} and~\ref{eq:deltac1cupc2} with the mediant inequality, we get that 

\begin{equation}
    \begin{gathered}
        \beta_a - \Delta_a^{C_1} \leq \frac{|C_1^a| + |C_2^a|}{|C_1| + |C_2|} \leq \alpha_a + \Delta_a^{C_2} \Rightarrow \\ 
        \beta_a - \max(\Delta_a^{C_1}, \Delta_a^{C_2}) \leq \frac{|C_1^a| + |C_2^a|}{|C_1| + |C_2|} \leq \alpha_a + \max(\Delta_a^{C_1}, \Delta_a^{C_2})
    \end{gathered}
\end{equation}

Since by definition, $\Delta_a^{C_1 \cup C_2}$ is the minimum value that satisfies equation~\ref{eq:deltac1cupc2}, we get that 

\begin{equation}
    \Delta_a^{C_1 \cup C_2} \leq \max(\Delta_a^{C_1}, \Delta_a^{C_2})
    \label{eq:helpful}
\end{equation}

Note that the empty cluster will have $\Delta_a^{\emptyset} = 0, \forall a \in [l]$, so $\max (\Delta_a^{\emptyset}, \Delta_a^{C_1 \cup C_2}) = \Delta_a^{C_1 \cup C_2}$. This also implies that $\max\limits_{C \in \mathcal{C}} \Delta_a^C \geq \max\limits_{C \in \mathcal{C}'} \Delta_a^C$. Since the attribute $a \in [l]$ was chosen arbitrarily, we also get that 

\begin{equation}
     \sum\limits_{a \in [l]}\max\limits_{C \in \mathcal{C}} \Delta_a^C \geq \sum\limits_{a \in [l]}\max\limits_{C \in \mathcal{C}'} \Delta_a^C,
\end{equation}
and thus the \textsc{Group Utilitarian} objective cannot increase by merging two clusters. Similarly, for the \textsc{Group Utilitarian-Sum}, since $0 = \Delta_a^{\emptyset} \leq \min(\Delta_a^{C_1}, \Delta_a^{C_2})$ and $ \Delta_a^{C_1 \cup C_2} \leq \max(\Delta_a^{C_1}, \Delta_a^{C_2})$, and all other clusters have the same proportional violation in both clusterings, we also get that $\sum\limits_{C \in \mathcal{C}} \Delta_a^C \geq \sum\limits_{C \in \mathcal{C}'} \Delta_a^C$. Since the attribute $a\in [l]$ was chosen arbitrarily, we also get that 

\begin{equation}
    \sum\limits_{a \in [l]} \sum\limits_{C \in \mathcal{C}} \Delta_a^C \geq \sum\limits_{a \in [l]} \sum\limits_{C \in \mathcal{C}'} \Delta_a^C,
\end{equation}
and thus the \textsc{Group Utilitarian-Sum} objective can also not increase by merging two clusters. Furthermore, equation~\ref{eq:helpful} implies that $\max(\Delta_a^{\emptyset}, \Delta_a^{C_1 \cup C_2}) \leq \max(\Delta_a^{C_1}, \Delta_a^{C_2})$ which in turn implies that $\max\limits_{C \in \mathcal{C}}\Delta_a^C \geq \max\limits_{C \in \mathcal{C}'}\Delta_a^C$. Since this holds for any arbitrary $a \in [l]$, it also implies that 

\begin{equation}
    \max\limits_{a \in [l], C \in \mathcal{C}}\Delta_a^C \geq \max\limits_{a \in [l], C \in \mathcal{C}'}\Delta_a^C
\end{equation}
and thus the \textsc{Group Egalitarian} objective cannot increase by merging two clusters. Finally, since $0 = \Delta_a^{\emptyset} \leq \min(\Delta_a^{C_1}, \Delta_a^{C_2})$ and $ \Delta_a^{C_1 \cup C_2} \leq \max(\Delta_a^{C_1}, \Delta_a^{C_2})$, and all other clusters have the same proportional violation in both clusterings, we also get that $\sum\limits_{C \in \mathcal{C}} \Delta_a^C \geq \sum\limits_{C \in \mathcal{C}'} \Delta_a^C$. Since the attribute $a\in [l]$ was chosen arbitrarily, we also get that 

\begin{equation}
    \max\limits_{a \in [l]} \sum\limits_{C \in \mathcal{C}} \Delta_a^C \geq \max\limits_{a \in [l]} \sum\limits_{C \in \mathcal{C}'} \Delta_a^C,
\end{equation}
and thus the \textsc{Group Egalitarian-Sum} objective cannot increase by merging two clusters.

For the induction step, the proof is identical to the proof for the \textsc{Balance} objective: if mergeability holds for merging a set of $w$ clusters, then it holds for merging a set of $w + 1$ clusters as well, by reducing to the base case: without loss of generality, denote the $w + 1$ clusters to be merged by $C_1, \cdots, C_{w + 1}$. By the induction hypothesis, mergeability holds for merging any $w$ clusters, so for $\cup_{j \in [w]} C_j$. Then, the base case applies for the clusters $\cup_{j \in [w]} C_j$ and $C_{w + 1}$ for all objectives.~\looseness=-1

\end{proof}

\subsection{A discussion on pattern-based and mergeability properties}
\label{sec:appendix-examplesofnonpatternmergeableobj}

\paragraph{Example of non-pattern based fairness objectives.} As mentioned in the introduction, informally, a pattern-based fairness objective computes a per-cluster fairness quantity that only depends on the number of data points from each sensitive attribute. In practice, many fairness objectives satisfy this property, as they aim to operationalize different versions of \textit{proportional representation}, with the goal of having balanced clusters among different sensitive attributes. Objectives that define fairness through the proportion of $k$-centers that are closest to them are not pattern-based, since they do not depend on the number of other nodes of different attributes in the same cluster but rather on equalizing the proportions of $k$-center assignments to different attributes~\cite{chen2019proportionally,jung2019center,li2021approximate} or the max average distance between points of different groups to their centers~\cite{ghadiri2021socially}. For example, the definition in~\citet{ghadiri2021socially} states that for two groups $R$ (red) and $B$ (blue), the fairness objective is defined for a clustering $\mathcal{C}$ as 

\begin{equation}
    \Phi(S, \mathcal{C}) = \max\left( \frac{f(S, \mathcal{C} \cap \mathcal{X}^R)}{|R|},\frac{f(S, \mathcal{C} \cap \mathcal{X}^B)}{|B|} \right),
\end{equation}

\begin{figure*}
\centering
\subfloat[]{\includegraphics[width=0.3\textwidth] {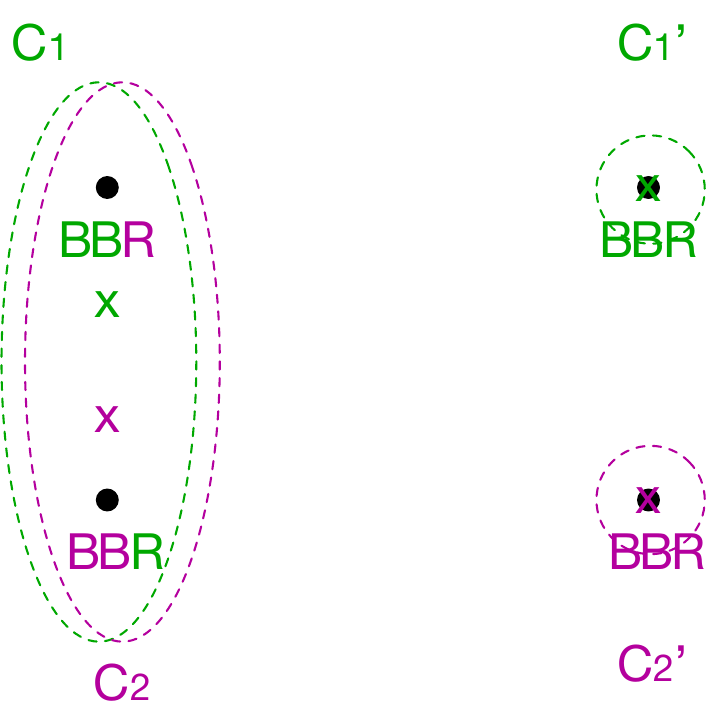}} 
\hspace{2cm}
\subfloat[]{\includegraphics[width=0.3\textwidth] {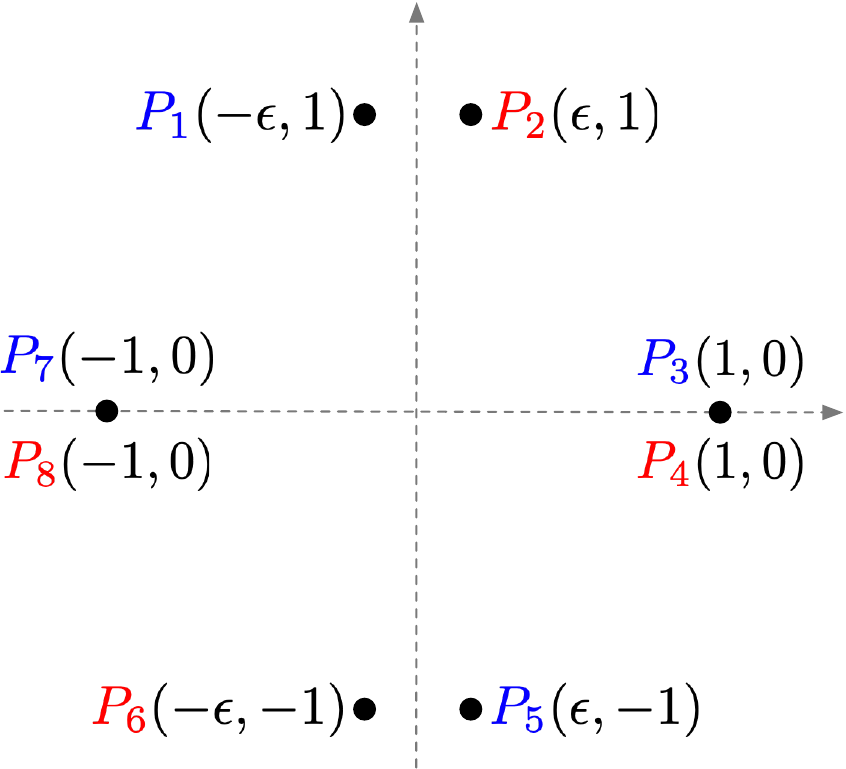}} 
\caption{(a) An illustration of clustering under for non-pattern based fairness objectives. (b) An illustration of the $(P_i)_{i \in [8]}$ sets for non-mergeable fairness objectives.}
\label{fig:pareto_illustrations_nonpattern_nonmergeable}
\end{figure*}

where $\mathcal{X}^A$ is the subset of points from $\mathcal{X}$ with attribute $A$. Take the following set of points: $3$ overlapping points $x_1, x_2, x_3$, with $x_1, x_2 \in B, x_3 \in R$, and $3$ overlapping points $x_4, x_5, x_6$ at different coordinates than the first $3$, with $x_4, x_5 \in B, x_6 \in R$. Then, the clustering $C_1 = \{x_1, x_2, x_6 \}$, $C_2 = \{x_3, x_4, x_5 \}$ has a non-zero fairness objective value with respect to its true centers (by computing the centroids). The clustering $C_1' = \{x_1, x_2, x_3 \}$, $C_2' = \{x_4, x_5, x_6 \}$, however, has the same pattern as $(C_1,C_2)$, but a fairness objective value of $0$ with respect to its true centroids. Even if we evalute $(C_1,C_2)$ with respect to the centroids of $(C_1', C_2')$, it still has a non-zero fairness objective value. See Figure~\ref{fig:pareto_illustrations_nonpattern_nonmergeable} (a) for an illustration.~\looseness=-1

\paragraph{Example of non-mergeable fairness objectives.} As discussed in the introduction, a mergeable objective means that the objective value can only improve or stay the same when merging any number of clusters. While many fairness objectives are mergeable, we give an example below of non-mergeable objectives. We note that objectives that enforce a minimum number of data points in each cluster tend to not be mergeable. For example, the $\tau$-ratio objective defined by~\cite{gupta2023efficient} is non-mergeable, defined as:

\begin{align} 
\sum _{x_i \in X}{\mathbb {I}}{( x_i \in C_j)}{\mathbb {I}}{(\sigma (x_i){=}a  )} {\ge } \tau _{\ell }\!\sum _{x_i \in X} {\mathbb {I}}{( x_i \in C_j )}\ \forall j \in [k] \text { and} \ \forall a \ \in [l] 
\end{align}

As the $\tau$-ratio enforces a minimum number of the total points that must go in each cluster, so empty clusters violate the mergeability condition on the fairness objective (since the number of clusters $k$ is fixed). 

\paragraph{Pareto front approximation gets arbitrarily large without mergeability.}
\label{sec:appendix-bad-example}
We rely on the assignment problem to provide an approximation for the clustering problem when computing the Pareto front, which works for pattern-based and mergeable fairness objectives. However, for fairness objectives that are pattern-based but are \textbf{not} mergeable, if we apply directly Algorithm~\ref{alg:dynprogr}, without using a modified fairness function to filter the dominated patterns and adjust the clusterings, as we did in the proof of Theorem \ref{thm:nonmerge-FC}, then the resulting approximation ratio is no longer necessarily bounded, as we show in the example below. We showcase this for the $\tau$-ratio objective.

We construct a dataset $\mathcal{X}$ such that $|\mathcal{X}| = 8m$ that contains $8$ sets of $m$ points each on the plane. Each point has a sensitive attribute, denote by $b$ (blue) or $r$ (red). We denote these sets by $(P_i)_{i \in [8]}$, constructed in the Euclidean space with the following coordinates (see Figure~\ref{fig:pareto_illustrations_nonpattern_nonmergeable} (b) for an illustration):~\looseness=-1

\begin{itemize}
    \item $P_1$ contains $2m-1$ blue points situated at coordinates $(-\epsilon, 1)$
    \item $P_2$ contains $2m-1$ red points situated at coordinates at $(\epsilon, 1)$
    \item $P_3$ contains $1$ blue point situated at coordinates $(1, 0)$
    \item $P_4$ contains $1$ red point of situated at coordinates $(1, 0)$
    \item $P_5$ contains $2m-1$ blue points situated at coordinates $(\epsilon, -1)$
    \item $P_6$ contains $2m-1$ red points situated at coordinates $(-\epsilon, -1)$
    \item $P_7$ contains $1$ blue point situated at coordinates $(-1, 0)$
    \item $P_8$ contains $1$ red point situated at coordinates $(-1, 0)$
\end{itemize}

Set $\epsilon < \frac{1}{8m}$, and $k=4$. Then, the cost of not assigning the $4$ points at $(1,0)$ and $(-1,0)$ their own two centers outweighs the benefits of assigning $2$ centers to the points clustered near $(0, 1)$ or $(0, -1)$. Therefore, the best $k$-means and $k$-median clustering yields centers $S = \{(0, 1), (1, 0), (0,-1), (-1, 0)\}$.

We set $\tau = \frac{1}{4}$ as our fairness constraint, which constrains every center to have exactly $m$ red points and exactly $m$ blue points. Observe that the optimal quality way $\phi$ to assign such points to $s_2$ is to
assign $m-1$ of the red points in $P_2$ and $m-1$ of the blue points in $P_5$ to $S_2$ (in addition to the two points assigned to $s_2$ by the unfair $k$-means/medians clustering). So the center at $s_2$ clusters together 2 sets of $m-1$ points separated by distance 2 on the plane. Even if we allow moving the location of $s_2$ after the assignment, the clustering cost is still lower bounded by$(2m-2)^{1/p}$, since the best center is unit distance from $2(m-1)$ points. Symmetrically, we do the corresponding assignment to $s_4$ and get a clustering cost lower bound of $2(2m-2)^{1/p}$.

Setting our centers to be $S^{*} = \{(0, 1), (0, 1), (0,-1), (0, -1)\}$, the assignment function $\phi^*$ with optimal quality under the fairness constraint maps $m$ points from each of $P_1$ and $P_2$ to $s^*_1$ and maps the remainder of the points from $P_1$ and $P_2$, as well as the points in $P_3$ and $P_4$ to $s^*_2$. Symmetrically, we do the corresponding assignment to $s^*_3$ and $s^*_4$ with the points from $P_5$, $P_6$, $P_7$, and $P_8$. This clustering achieves a clustering cost at most $ ((8m-4)\epsilon^p + 4(\sqrt{2})^p)^{1/p}$, where the first term comes from the $8m-4$ points in sets $P_1, P_2, P_5, P_6$ and the second term comes from the 4 points in sets $P_3, P_4, P_7, P_8$.~\looseness=-1

Let $\phi$ be the lowest cost assignment to centers $S$, and let $S^m$ be the best possible centers for $\phi$. Then we have that:

\begin{equation}\frac{c(\phi, S^m)}{c(\phi^*, S^*)} \geq \frac{2(2m-2)^{1/p}}{((8m-4)\epsilon^p + 4(\sqrt{2})^p)^{1/p}}
\end{equation}
Considering the $k$-means or the $k$-median clustering objectives, we get $(8m-4)\epsilon^p \leq 1$, since we set $\epsilon < \frac{1}{8m}$. Therefore,
\begin{equation}
    \lim_{m\rightarrow\infty}\frac{c(\phi, S^m)}{c(\phi^*, S^*)} = \infty
\end{equation}
Thus, for any constant $b$, the assignment Pareto set is not a $(b, 1)$ multiplicative approximation for the clustering Pareto set for non-mergeable fairness functions. This justifies the need for a modified algorithm for non-mergeable fairness objectives that can change the set of centers while searching for the best assignment, as described in Appendix \ref{sec:appendix-proofs}.

\section{Datasets details}
\label{sec:appendix-datadetails}
For all datasets, we subsample $1,000$ data points, as the datasets sizes are prohibitively large. For the Adult and Census datasets, we use all numerical features available in the data and embed them in Euclidean space. For the BlueBike dataset, we use the starting and ending latitude and longitude as coordinates, directly embedded in Euclidean space. For all datasets, the gender of users is self-reported. Table~\ref{table:data-details} describes the characteristics of all datasets. 

\begin{table}[ht!]
    \caption{Data and experimental details.}
    \label{table:data-details}
    \begin{center}
    \begin{tabular}{cccl}
    \toprule
      &  \# of features
      &  Sensitive attribute & $\delta$\\
    \midrule
    Adult &  5 &  Gender & 0.05\\
    Census1990 &  66  & Gender & 0.001 \\    
    BlueBike &  4 & Gender & 0.01 
\end{tabular}
\end{center}
\end{table} 

We note that some datasets naturally cluster into fairer clusters than others. For this reason, setting $\delta$ too high renders a single point on the Pareto front, since the vanilla clustering itself will be fair for proportional violation-based objectives. In fact, the proportional violation value will be equal to $0$. For this reason, we experiment with various values of $\delta$, reported in Table~\ref{table:data-details} for the experiments presented in the main text. 

\section{Repeated-FCBC: Experimental Details}
\label{sec:appendix-exp-details}

\begin{wrapfigure}{r}{0.4\textwidth}
\centering
\includegraphics[width=0.3\textwidth] {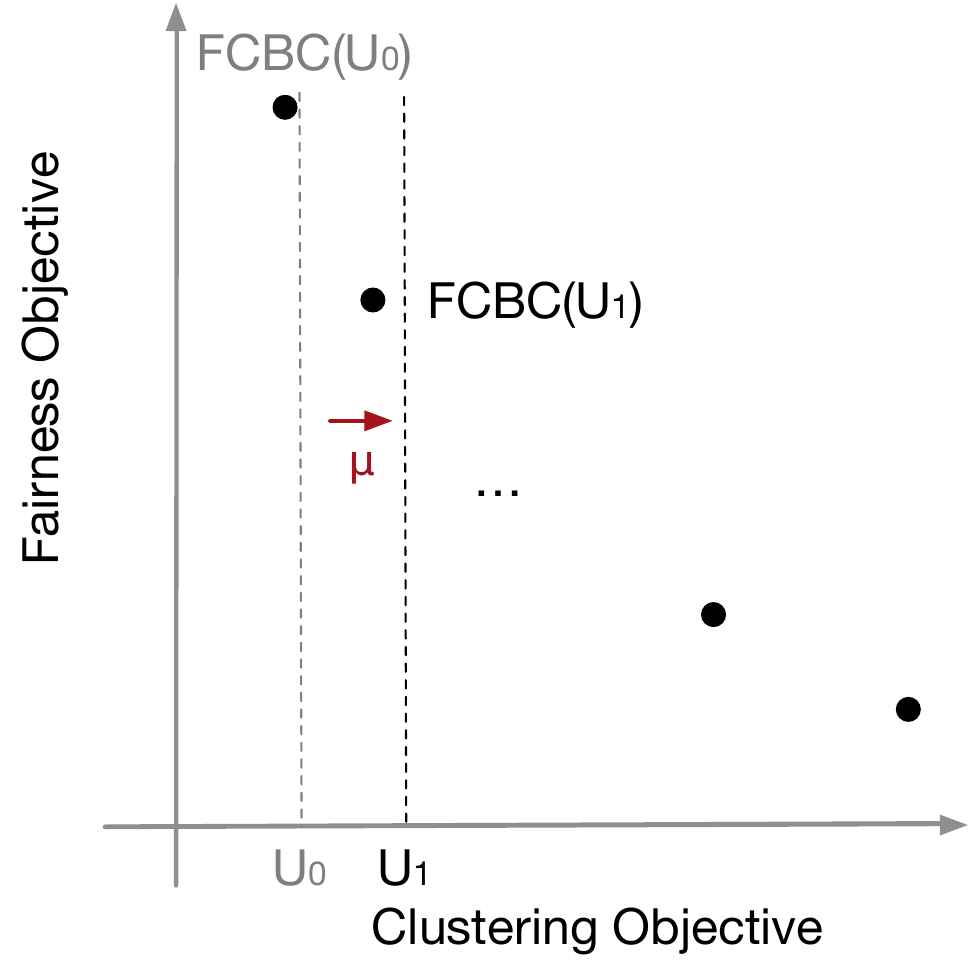}
\caption{An illustration of implementing the repeated FCBC algorithm as the clustering cost upper bound U varies.}
\label{fig:pareto_illustrations_fcbc}
\end{wrapfigure}

As described in the main text, we investigate alternatives for faster recovery of the Pareto front in the case of two objectives: clustering cost and fairness. We employ a recently proposed linear programming method that bounds the clustering objective and optimizes a fairness objective, up to an approximation, denoted the FCBC algorithm (Algorithm $1$ in~\citet{esmaeili2021fair}). We implement the FCBC algorithm repeatedly by discretizing over the clustering cost space. We perform a sweep over the upper bound $U$ values, setting its minimum value equal to the clustering cost of a vanilla clustering algorithm and its max value equal to a constant times the clustering cost of a vanilla clustering algorithm, for some chosen constant. We describe the approach formally in Algorithm~\ref{alg:repeated_fcbc}, with an illustration in Figure~\ref{fig:pareto_illustrations_fcbc}.

We compare the Pareto front recovered from Algorithm~\ref{alg:repeated_fcbc} with the Pareto front recovered from the dynamic programming approach in Algorithm~\ref{alg:dynprogr} on all the real-world datasets in Figure~\ref{fig:adult_300_500_dynprogr_fcbc} in Section~\ref{sec:experiments}, setting $U_{\mathrm{max}} = U_{\mathrm{min}} \cdot C$, where $U_{\mathrm{min}}$ is the clustering cost of a vanilla clustering algorithm (in our case, k-means++), $C = 1.5$, and $N  = 50$. We note that setting $C$ equal to $1$ is equivalent to constraining the clustering cost in the FCBC algorithm to be equal to the vanilla clustering algorithm cost.~\looseness=-1

\begin{algorithm}
\caption{Repeated FCBC for recovering the Pareto Front}
\begin{algorithmic}
\State Input: $U_{\mathrm{min}}, U_{\mathrm{max}}, N, \textsc{Clustering-Obj}, \textsc{Fairness-Obj}$. 
\State Step $1$: Discretize the interval $[U_{\mathrm{min}}, U_{\mathrm{max}}]$: set $\eta =\left( U_{\mathrm{max}} - U_{\mathrm{max}} \right)/ N $ and
\For{$i \gets 0$ to $N$}     
\State Set $U_i = U_{\mathrm{min}} + \eta \cdot i$
\EndFor
\State Step $2$: apply the FCBC algorithm for the discretized clustering cost bound sequence $(U_i)_i$, adding results to the recovered Pareto front approximation $X_P^{\mathrm{FCBC}}$, initialized to  $X_P^{\mathrm{FCBC}} = \emptyset$:
\For{$i \gets 0$ to $N$}   
    \State $X_P^{\textsc{FCBC}} \gets X_P^{\mathrm{FCBC}} \cup \{FCBC(U_i, \textsc{Clustering-Obj}, \textsc{Fairness-Obj}) \}$
\EndFor
\State Output: $X_P^{\mathrm{FCBC}}$
\end{algorithmic}
\label{alg:repeated_fcbc}
\end{algorithm}

\begin{figure}[!ht]
\centering
\subfloat[Group Util]
{\includegraphics[width=0.33\textwidth] {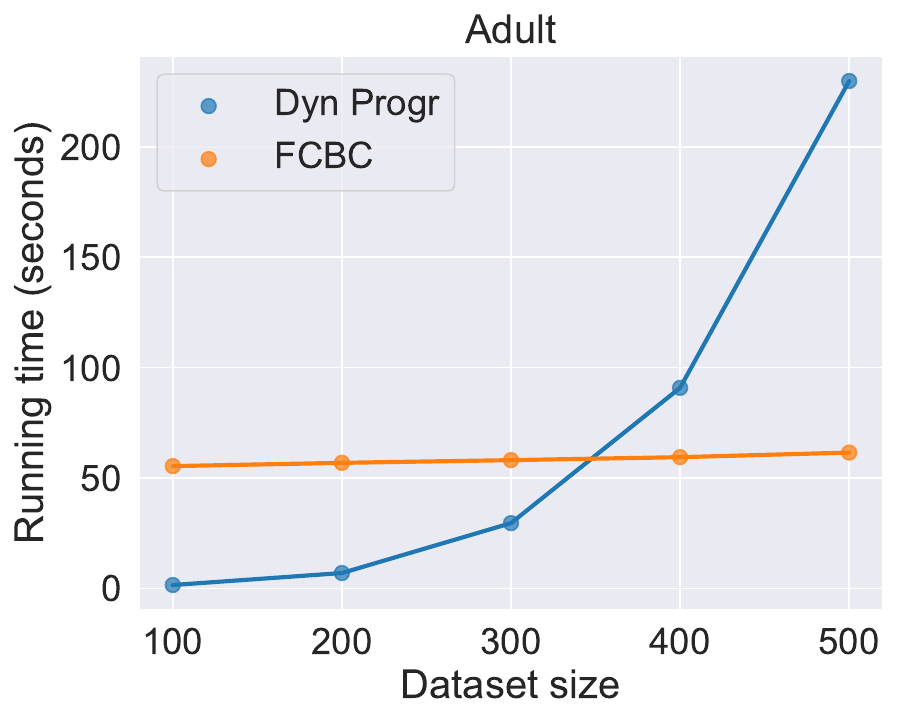}}
\subfloat[Group Util]
{\includegraphics[width=0.33\textwidth] {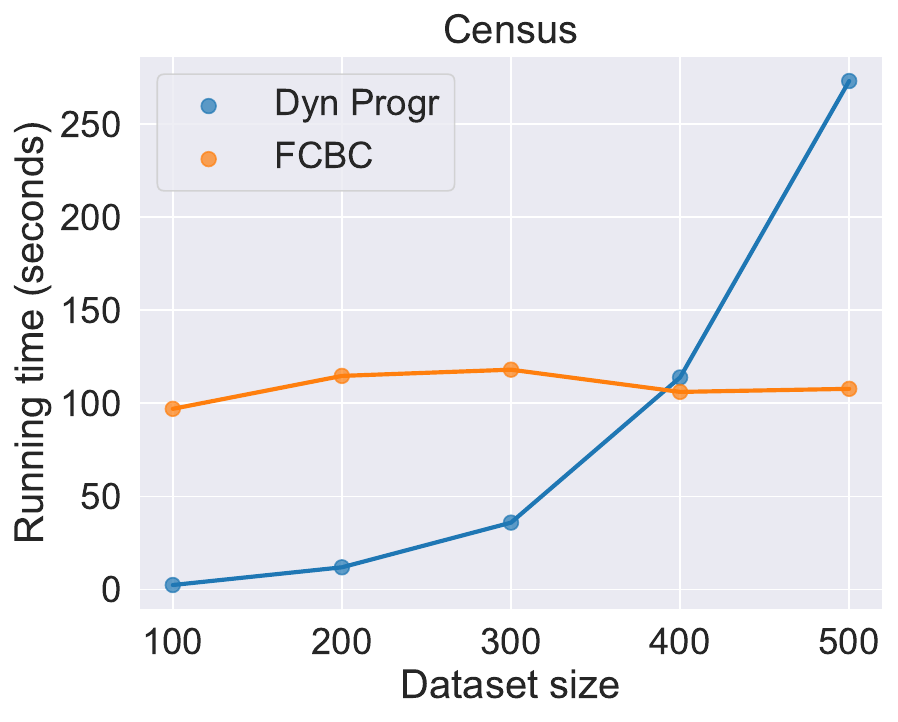}}
\subfloat[Group Util]
{\includegraphics[width=0.33\textwidth] {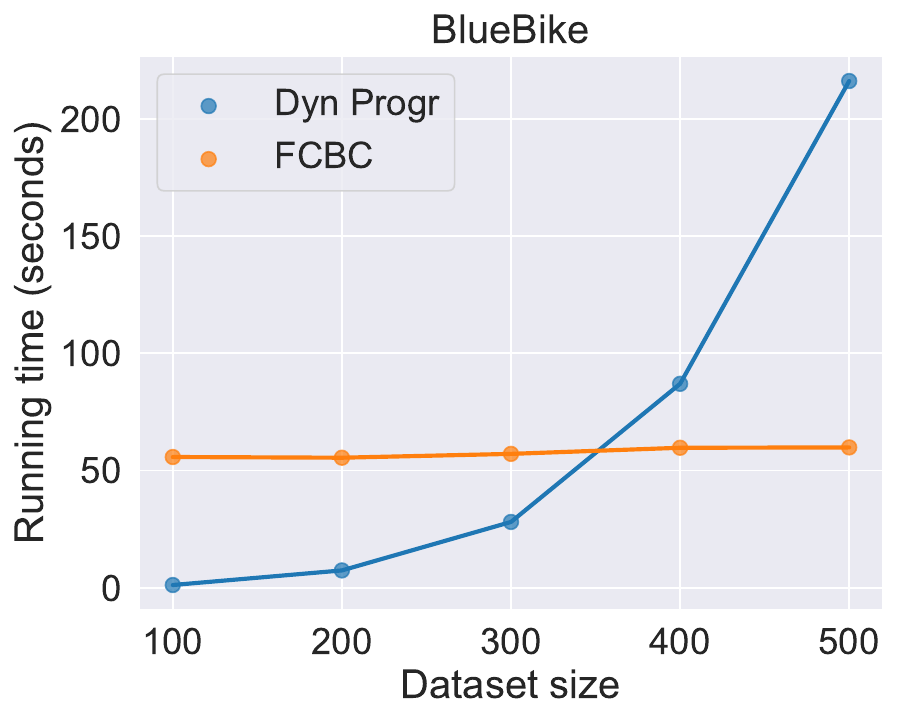}} \\
\subfloat[Group Egalit]
{\includegraphics[width=0.33\textwidth] {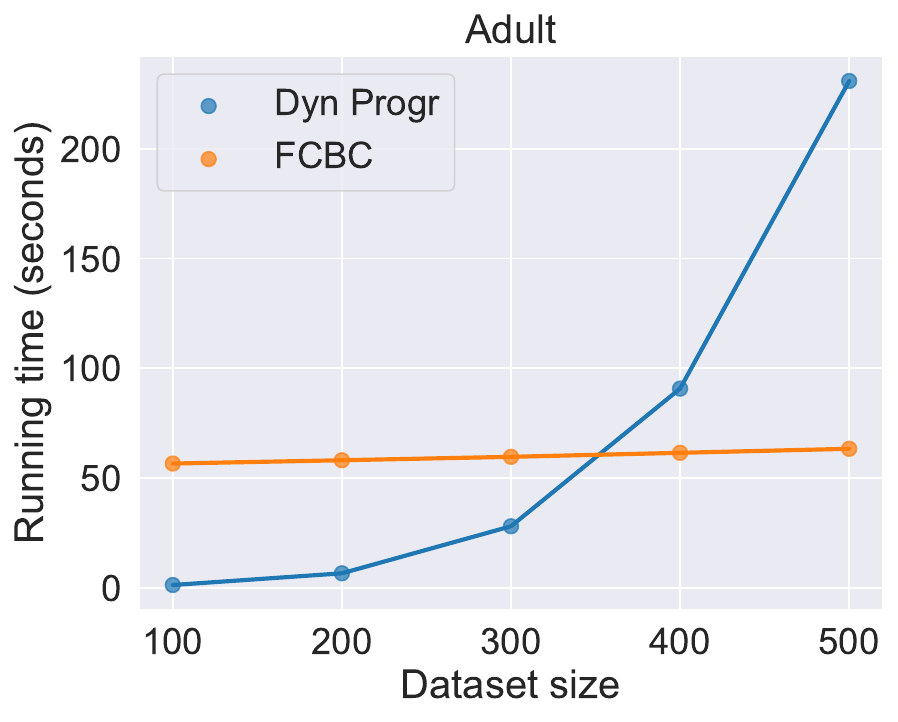}}
\subfloat[Group Egalit]
{\includegraphics[width=0.33\textwidth] {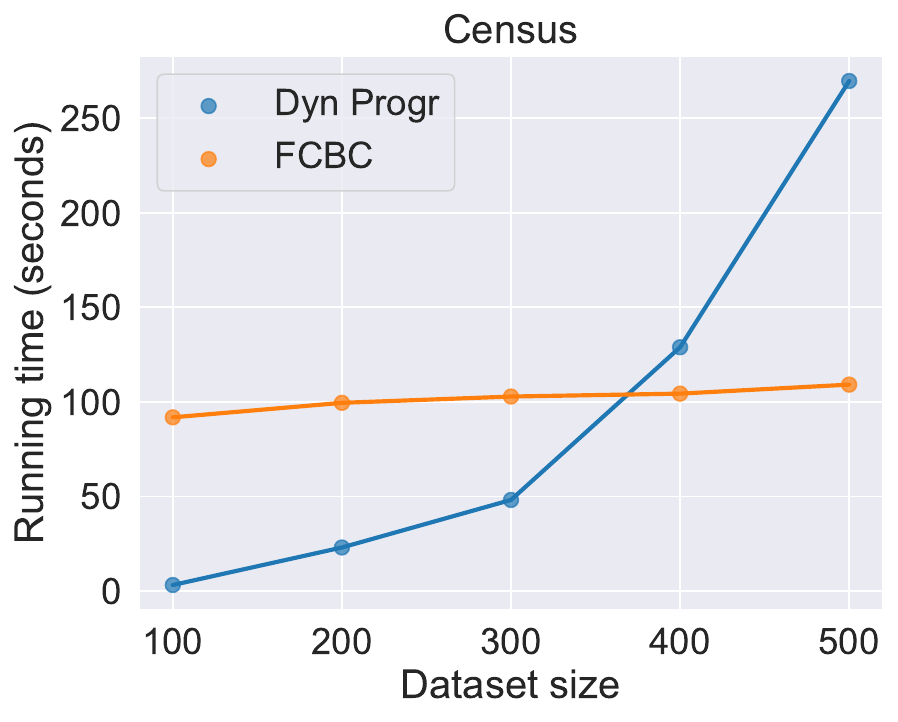}}
\subfloat[Group Egalit]
{\includegraphics[width=0.33\textwidth] {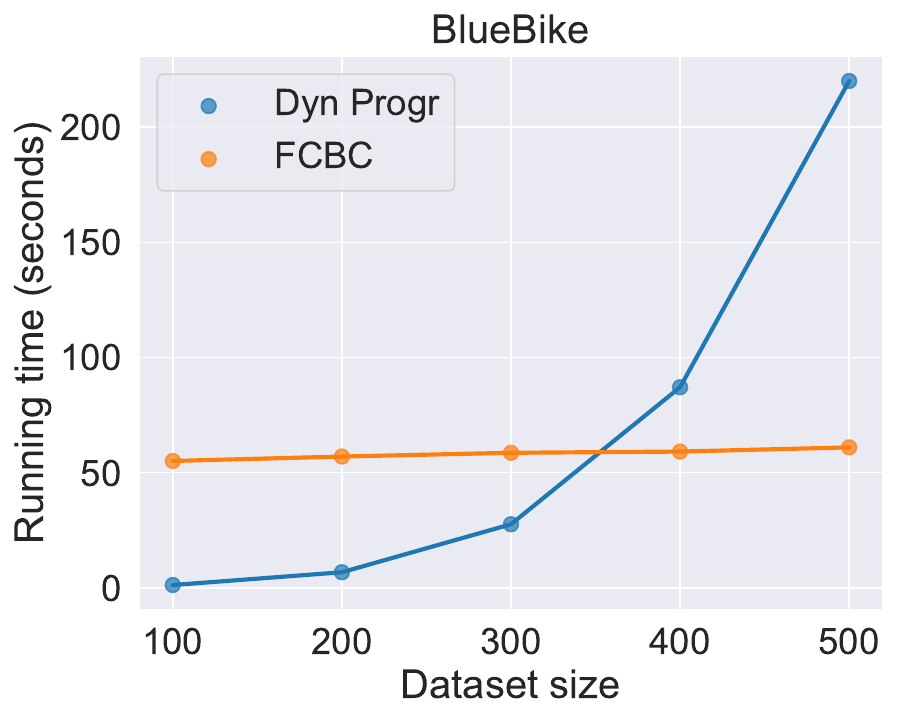}} \\
\caption{Running time comparison with our dynamic programming approach from Algorithm~\ref{alg:dynprogr}, labeled as `Dyn Progr', and the repeated FCBC approach from Algorithm~\ref{alg:repeated_fcbc}, labeled as `FCBC', for each dataset (by column) and for the \textsc{Group Utilitarian} and \textsc{Group Egalitarian} objective (by row).~\looseness=-1}
\label{fig:dynprogr_fcbc_runningtime}
\end{figure}

\paragraph{Running time comparison.} We illustrate in Figure~\ref{fig:dynprogr_fcbc_runningtime} the running time comparison between the dynamic programming approach from Algorithm~\ref{alg:dynprogr}, labeled as `Dyn Progr', and the repeated FCBC approach from Algorithm~\ref{alg:repeated_fcbc}, labeled as `FCBC', for samples of different sizes of all the datasets. We showcase two objectives, \textsc{Group Utilitarian} and \textsc{Group Egalitarian}, since the original FCBC algorithm~\cite{esmaeili2021fair} is designed only for these two, among the five objectives we described in Section~\ref{sec:experiments}. We note that Algorithm~\ref{alg:dynprogr} has a similar running time for all other objectives. The experimental running time follows the analysis presented in Section~\ref{sec:theory}: for $k = 2$ clusters and $l = 2$ sensitive attributes, one would expect a running time of $O(n^2)$. 

\newpage
\section{Additional Experiments}
\label{sec:appendix-moreexp}

\begin{figure*}[!ht]
\centering
\subfloat[Balance]
{\includegraphics[width=0.2\textwidth] {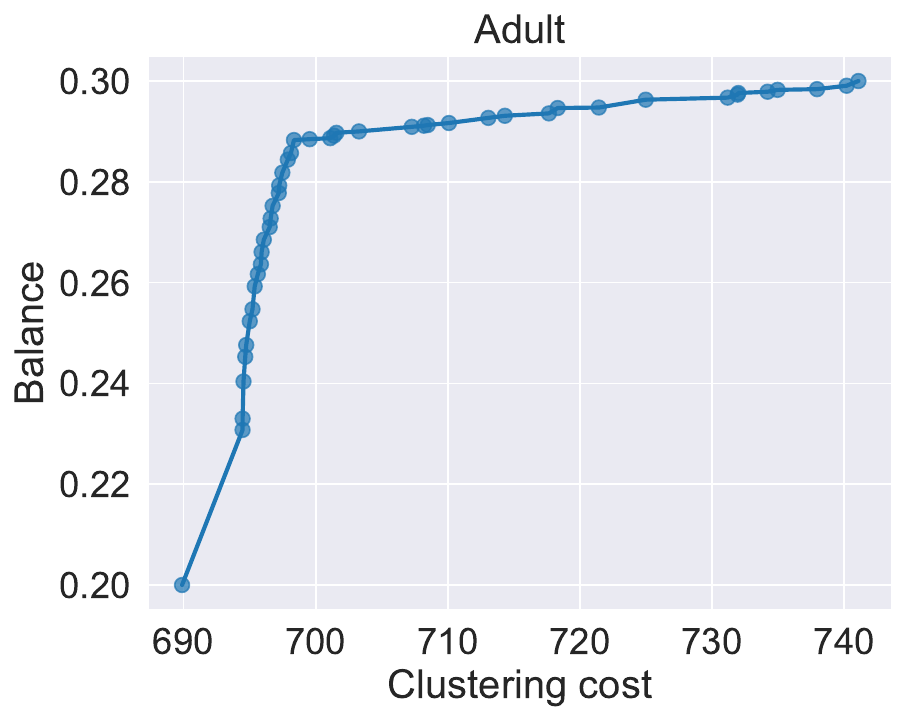}}
\subfloat[Group Util]
{\includegraphics[width=0.2\textwidth] {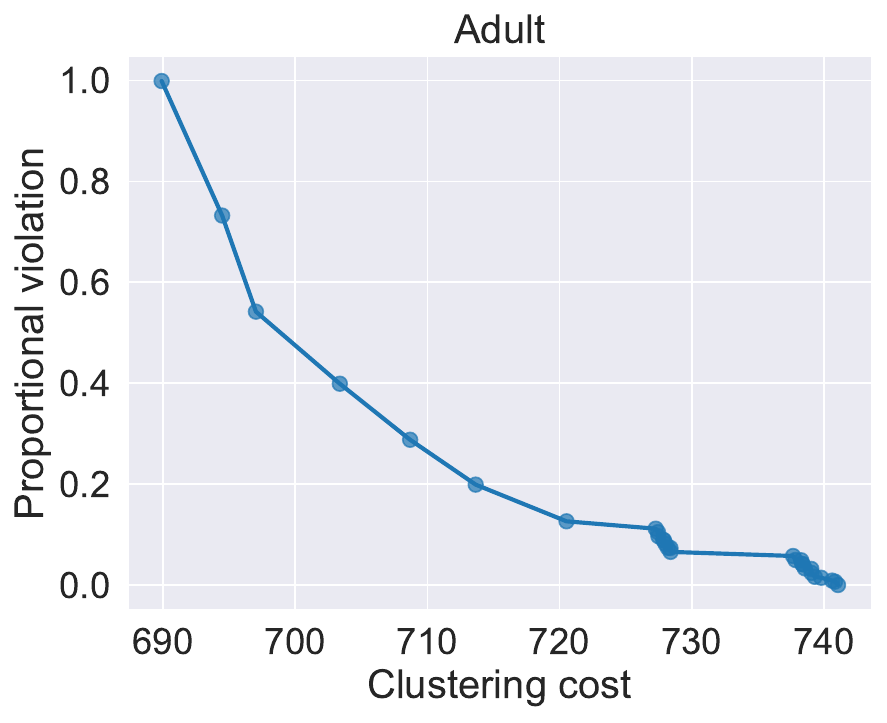}}
\subfloat[Group Util-Sum]
{\includegraphics[width=0.2\textwidth] {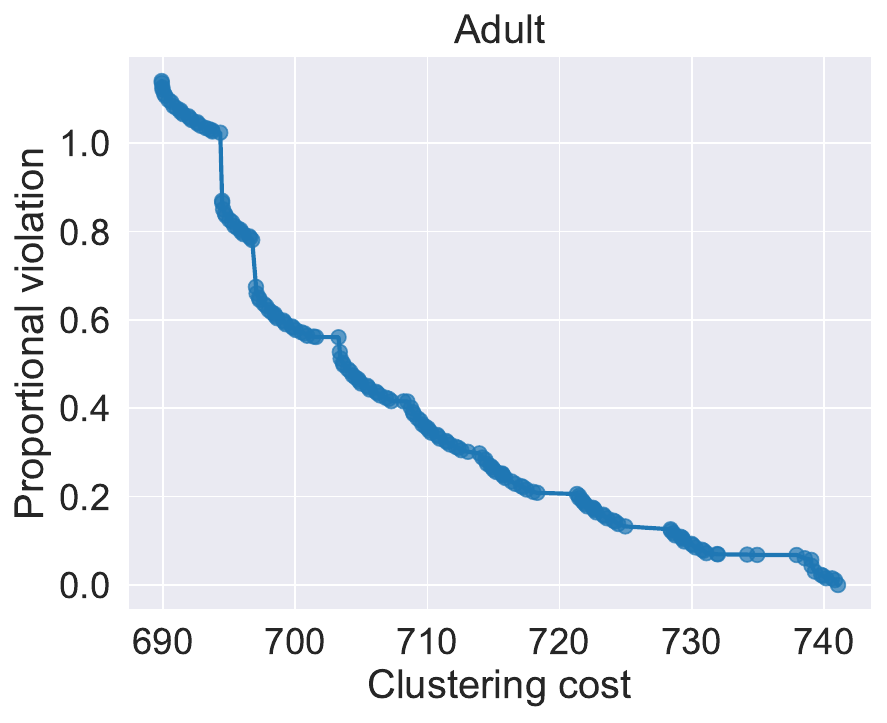}} 
\subfloat[Group Egalit]
{\includegraphics[width=0.2\textwidth] {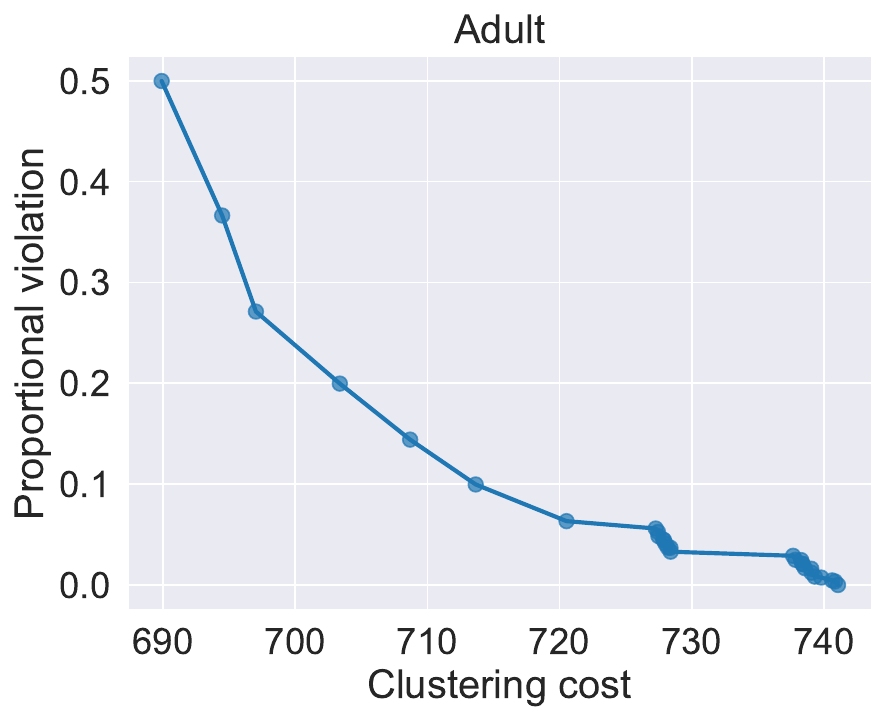}} 
\subfloat[Group Egalit-Sum]
{\includegraphics[width=0.2\textwidth] {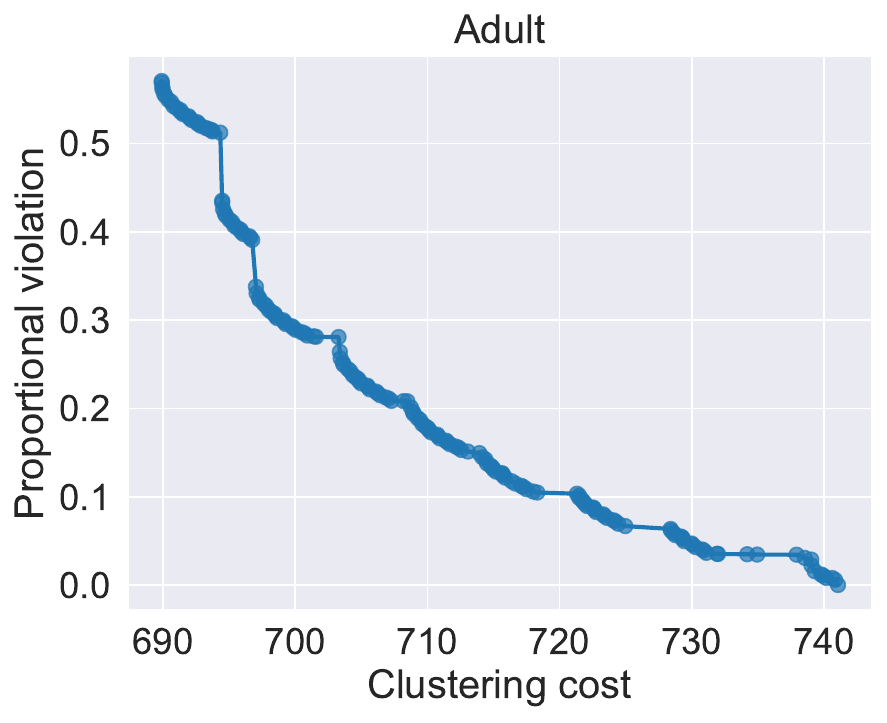}} \\ 
\subfloat[Balance]
{\includegraphics[width=0.2\textwidth] {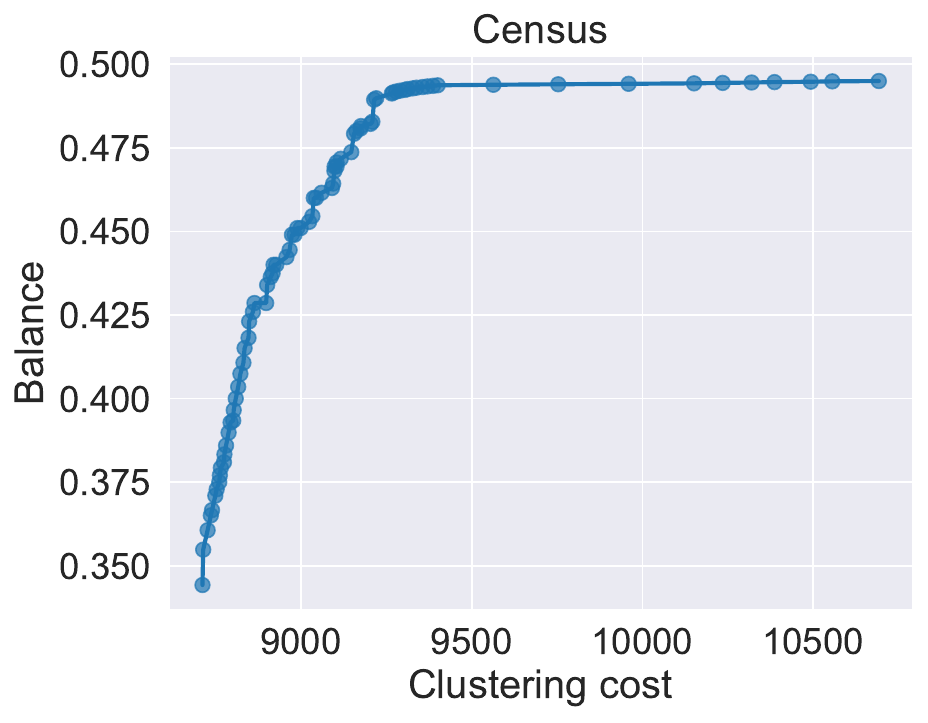}}
\subfloat[Group Util]
{\includegraphics[width=0.2\textwidth] {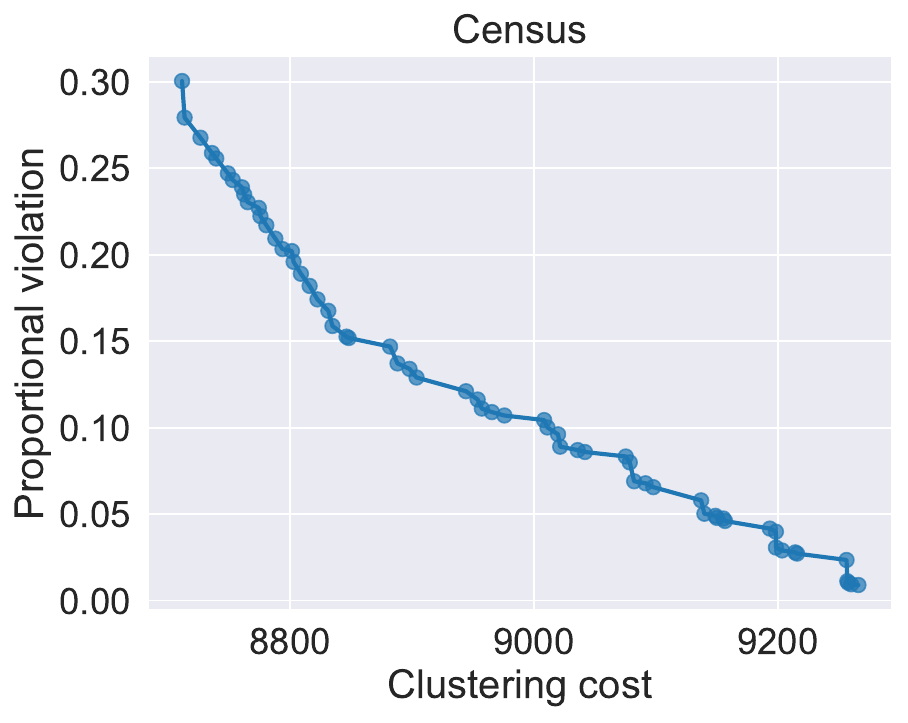}}
\subfloat[Group Util-Sum]
{\includegraphics[width=0.2\textwidth] {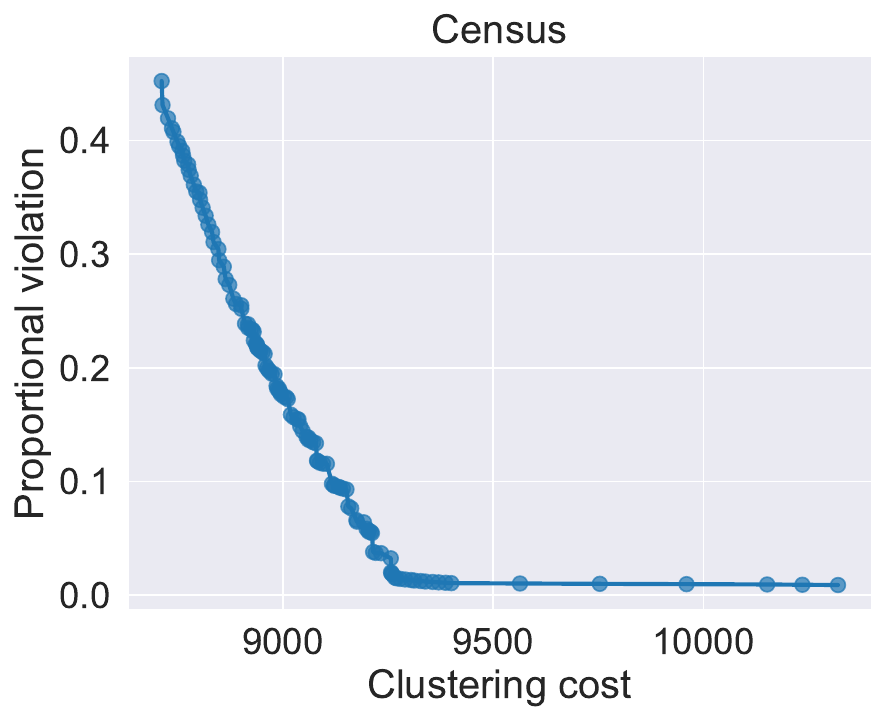}} 
\subfloat[Group Egalit]
{\includegraphics[width=0.2\textwidth] {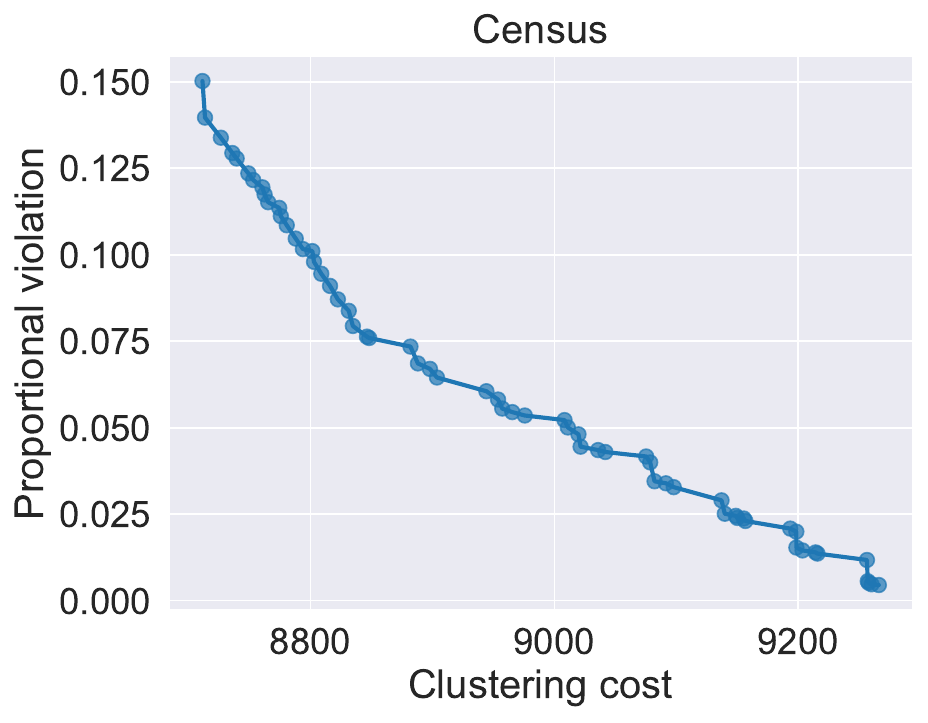}} 
\subfloat[Group Egalit-Sum]
{\includegraphics[width=0.2\textwidth] {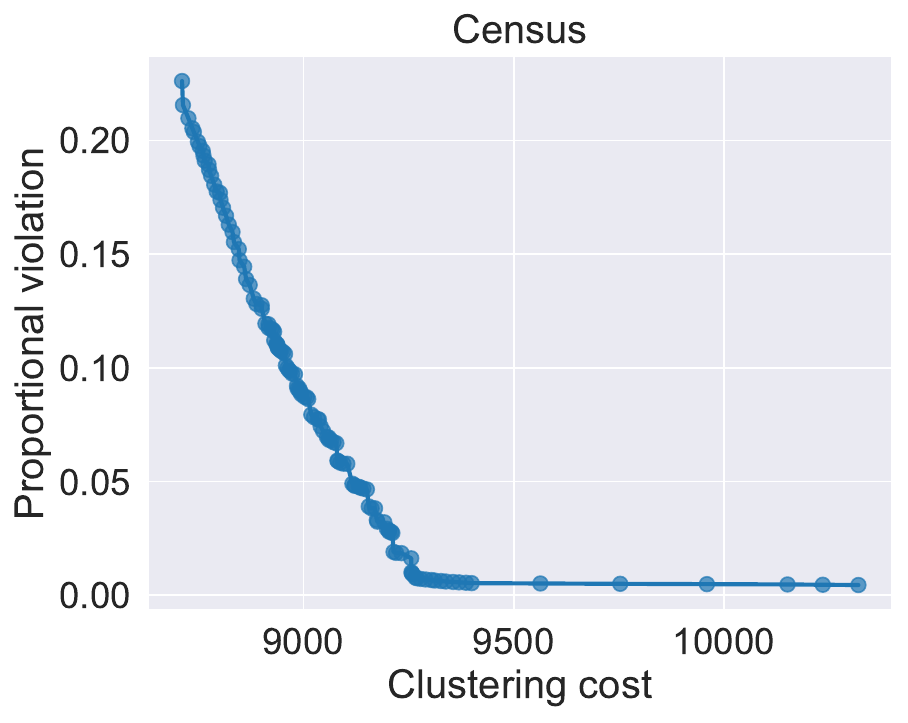}} \\
\subfloat[Balance]
{\includegraphics[width=0.2\textwidth] {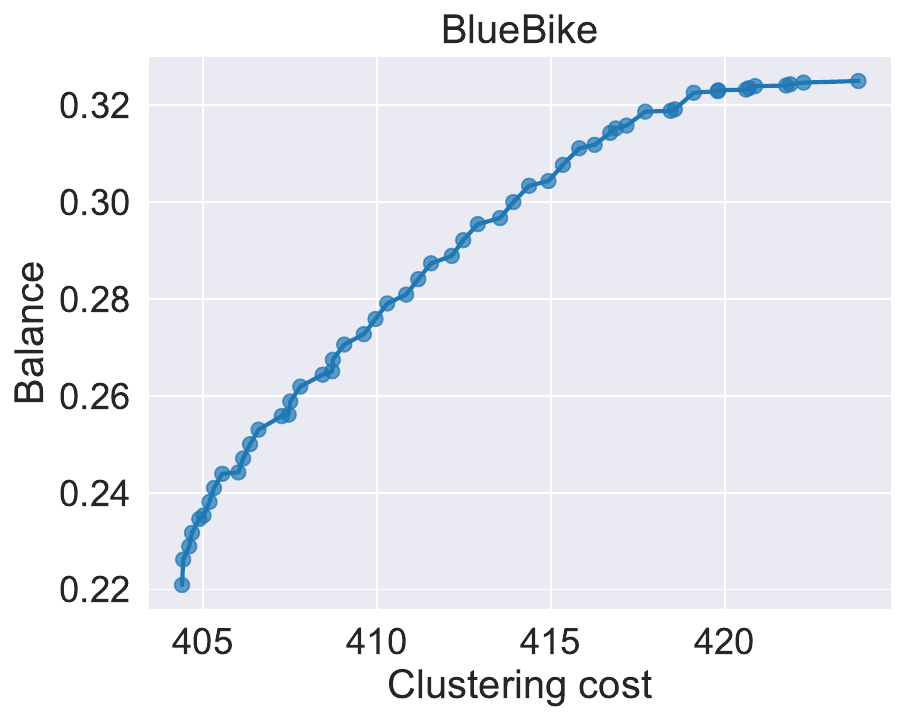}}
\subfloat[Group Util]
{\includegraphics[width=0.2\textwidth] {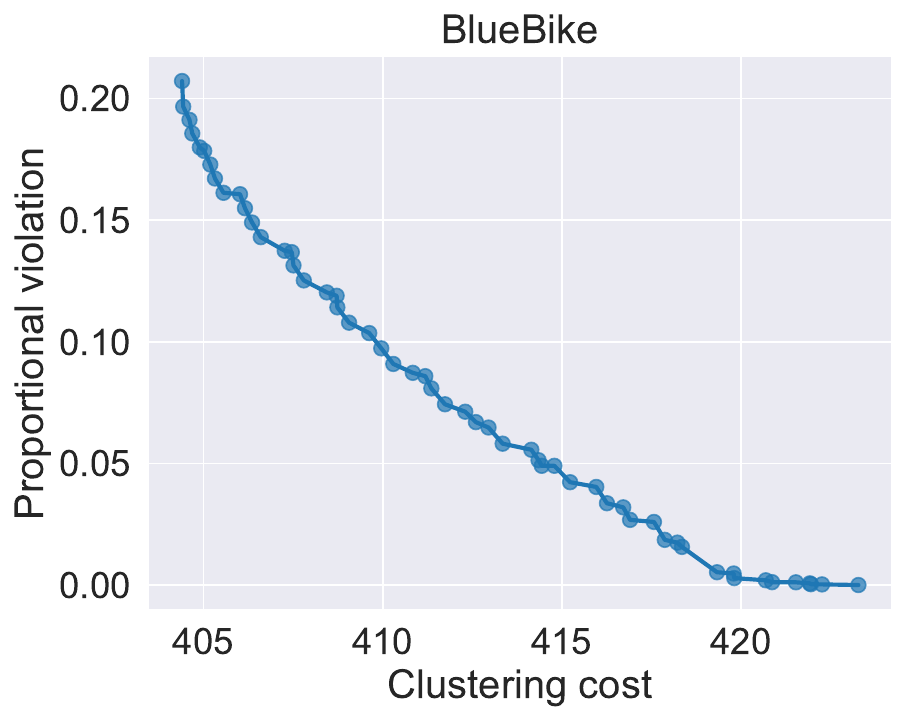}}
\subfloat[Group Util-Sum]
{\includegraphics[width=0.2\textwidth] {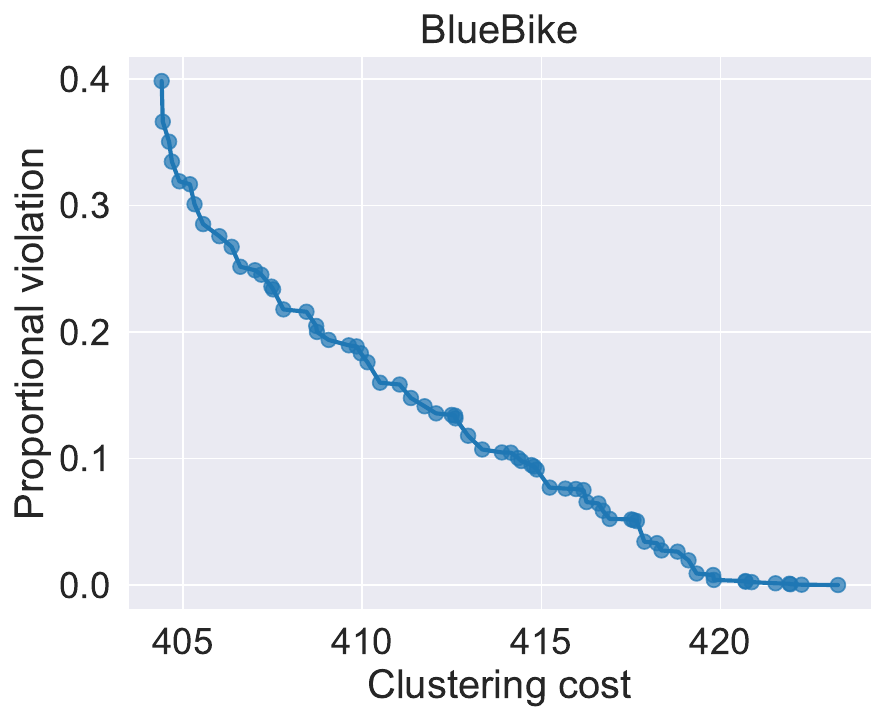}} 
\subfloat[Group Egalit]
{\includegraphics[width=0.2\textwidth] {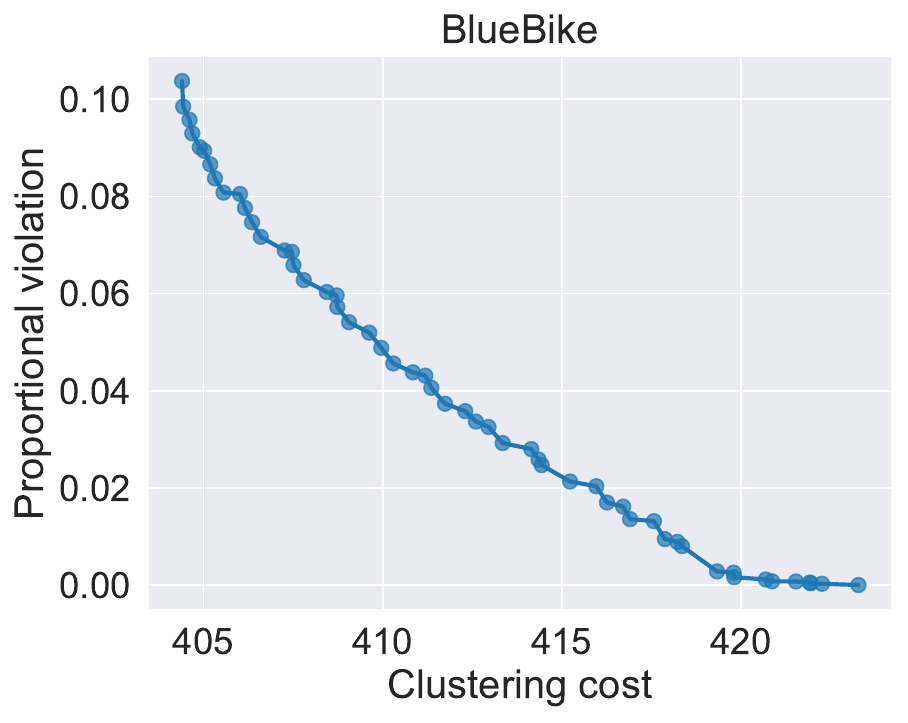}} 
\subfloat[Group Egalit-Sum]
{\includegraphics[width=0.2\textwidth] {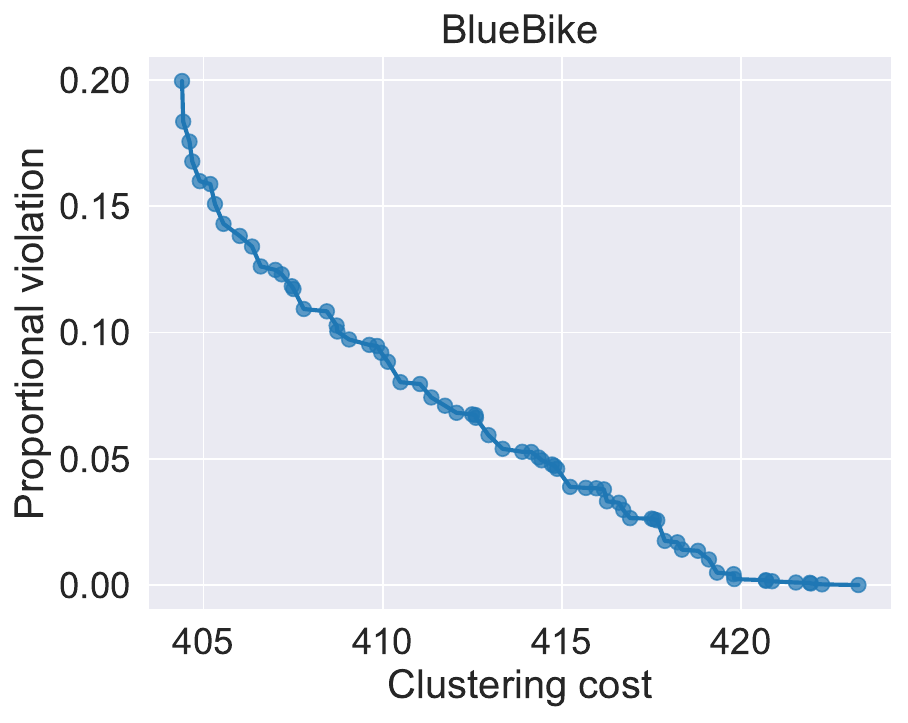}} 
\caption{Pareto front recovered by Algorithm~\ref{alg:dynprogr} for the Adult, Census, and BlueBike datasets (by row), for various fairness objectives (by column), for $k = 3$ clusters.}
\label{fig:alldata_1000_dynprogr-k3}
\end{figure*}

\begin{figure*}[!ht]
\centering
\subfloat[Group Util]
{\includegraphics[width=0.33\textwidth] {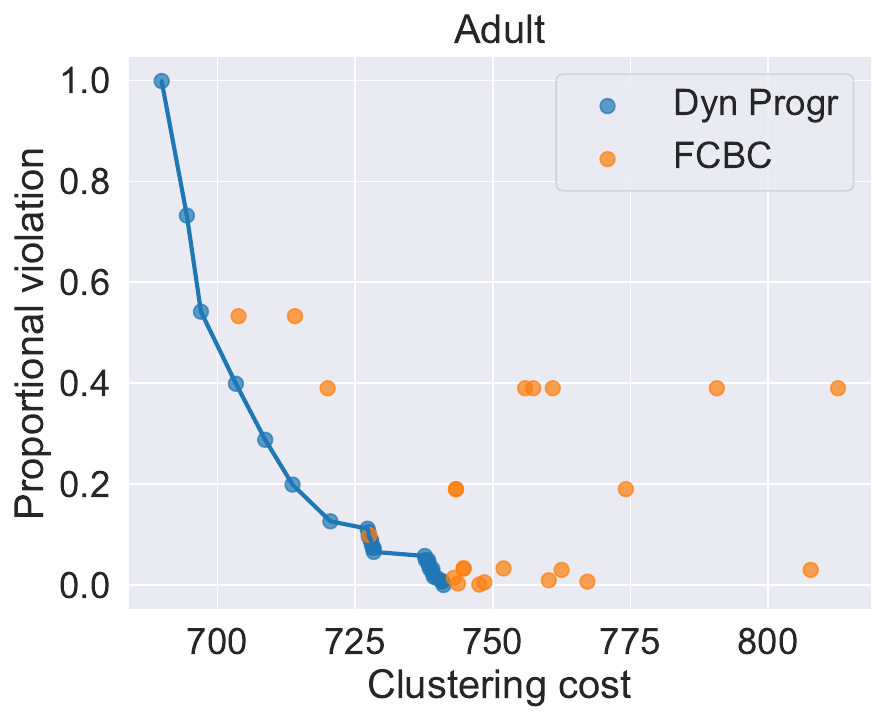}} 
\subfloat[Group Util]
{\includegraphics[width=0.33\textwidth] {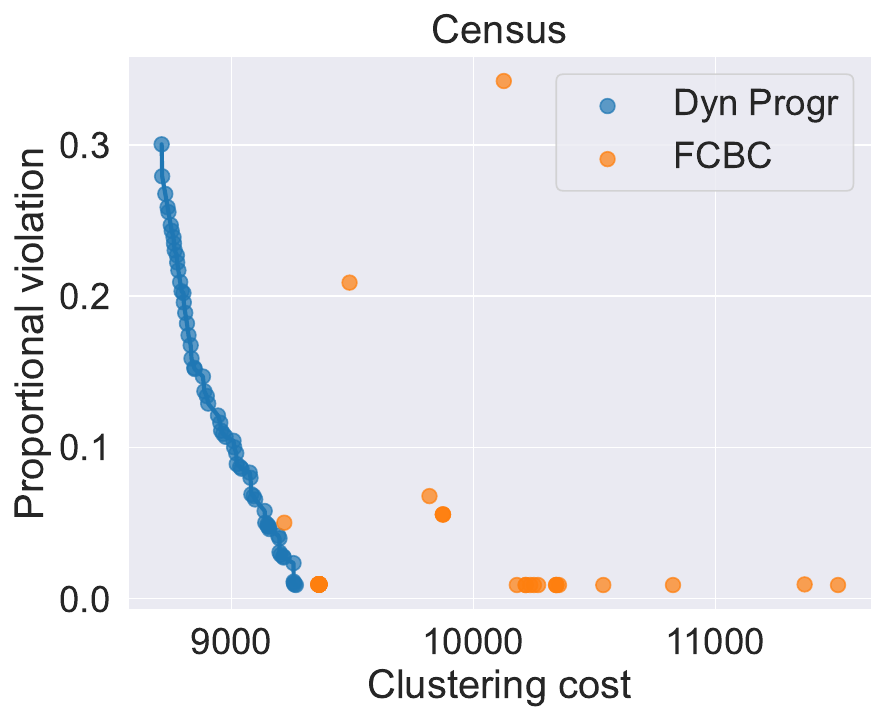}} 
\subfloat[Group Util]
{\includegraphics[width=0.34\textwidth] {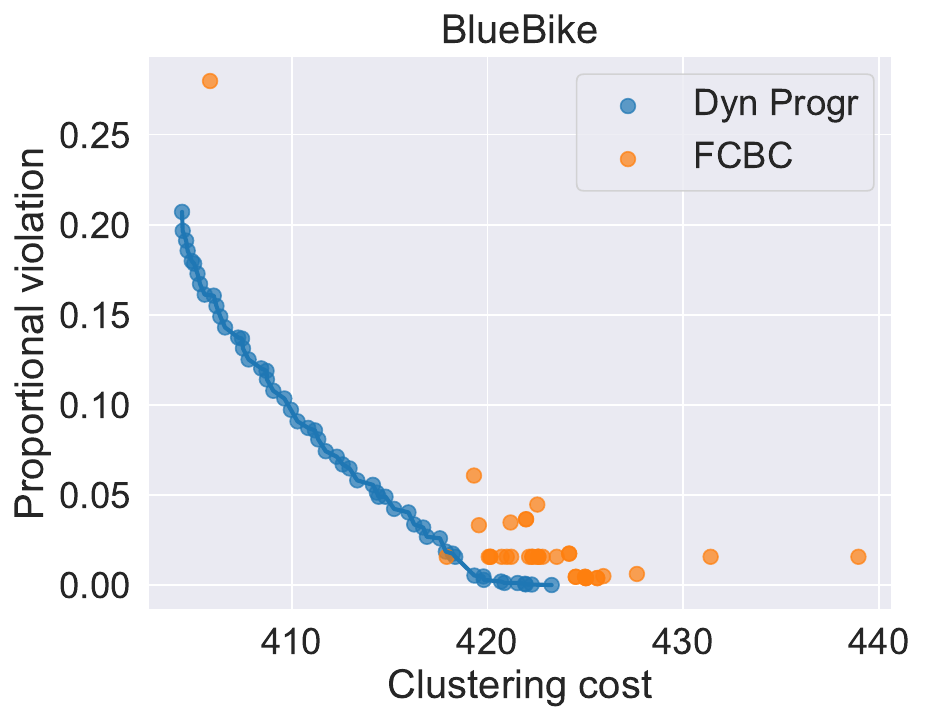}} \\
\subfloat[Group Egalit]
{\includegraphics[width=0.33\textwidth] {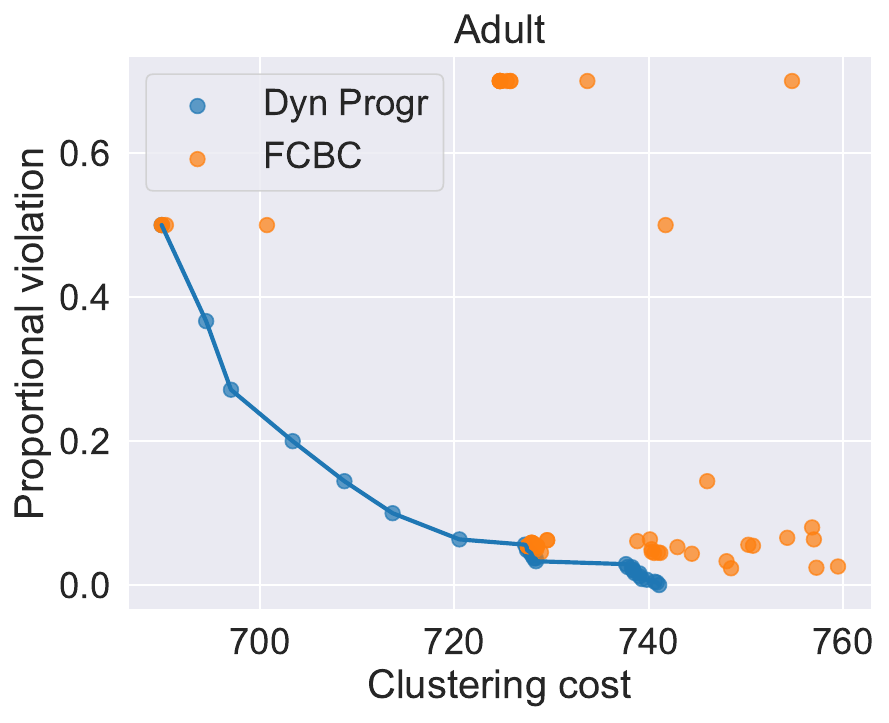}} 
\subfloat[Group Egalit]
{\includegraphics[width=0.33\textwidth] {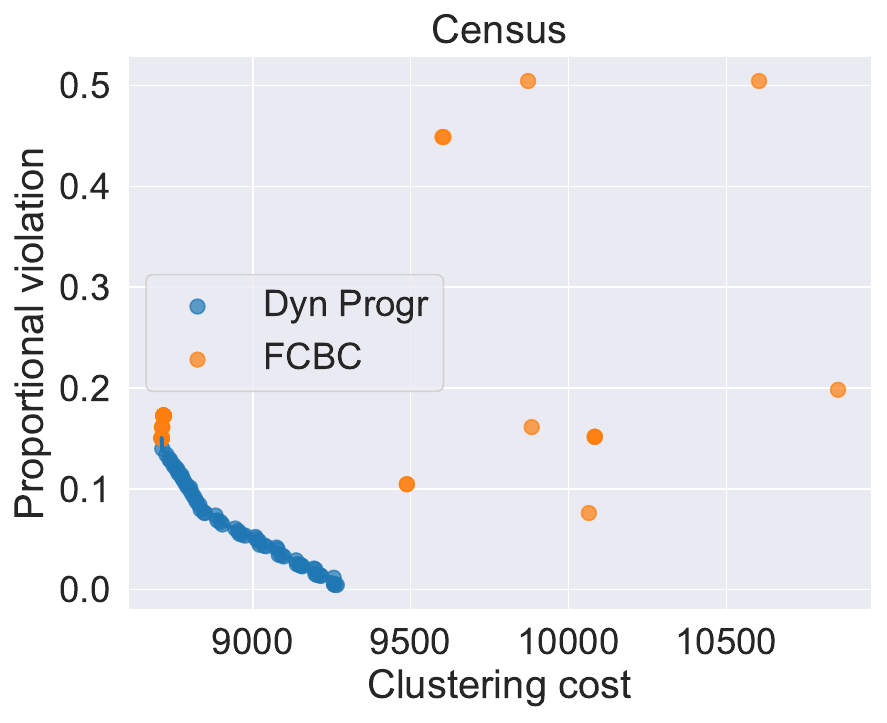}} 
\subfloat[Group Egalit]
{\includegraphics[width=0.34\textwidth] {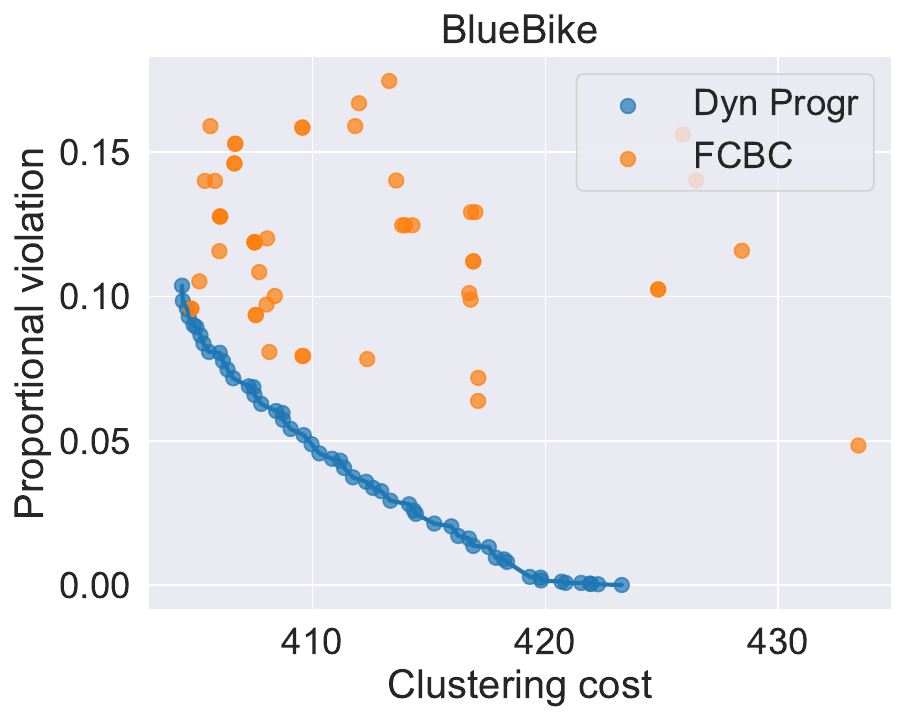}} 
\caption{Pareto front recovered by Algorithm~\ref{alg:dynprogr} (labeled as `Dyn Progr', in blue) and by Algorithm~\ref{alg:repeated_fcbc} (labeled as `FCBC', in orange) for the Adult, Census, and BlueBike datasets (by column) and for the \textsc{Group Utilitarian} and \textsc{Group Egalitarian} objectives (by row), for $k = 3$ clusters.}
\label{fig:adult_300_500_dynprogr_fcbc-k3}
\end{figure*}

We present experimental results on the three datasets and the five fairness objectives defined in Section~\ref{sec:experiments} for $k = 3$ clusters in Figures~\ref{fig:alldata_1000_dynprogr-k3} and~\ref{fig:adult_300_500_dynprogr_fcbc-k3}. We note that results are qualitatively similar as for $k = 2$ clusters. As mentioned in the main text, we note that the Pareto front need not be strictly convex for two minimization objectives, nor strictly concave for a minimization objective and a maximization objective (as in the case of the clustering objective and the balance objective), since it simply consists of the undominated points. ~\looseness=-1

\begin{figure*}[!ht]
\centering
\subfloat[]
{\includegraphics[width=0.33\textwidth] {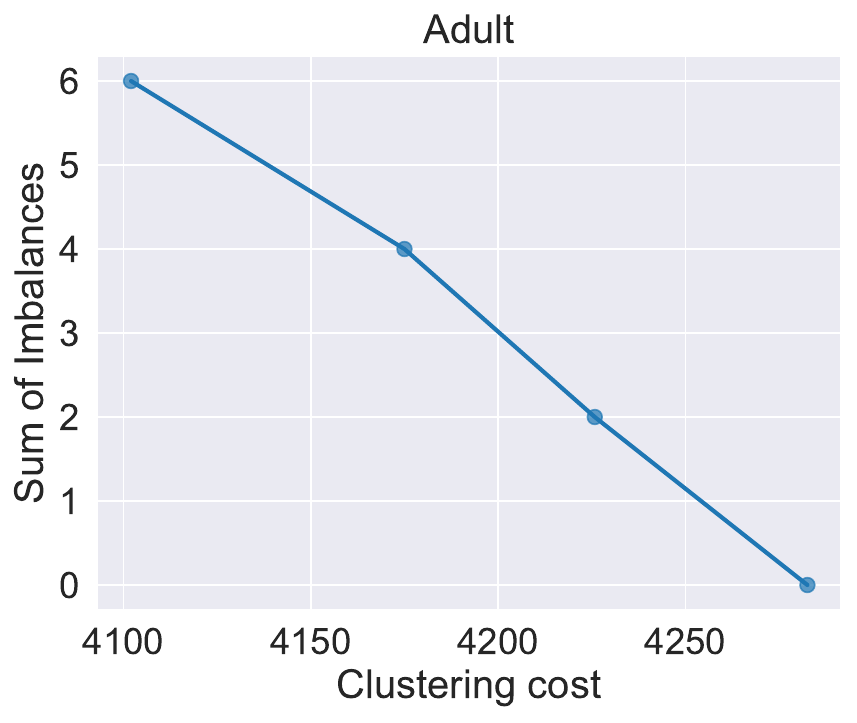}} 
\subfloat[]
{\includegraphics[width=0.34\textwidth] {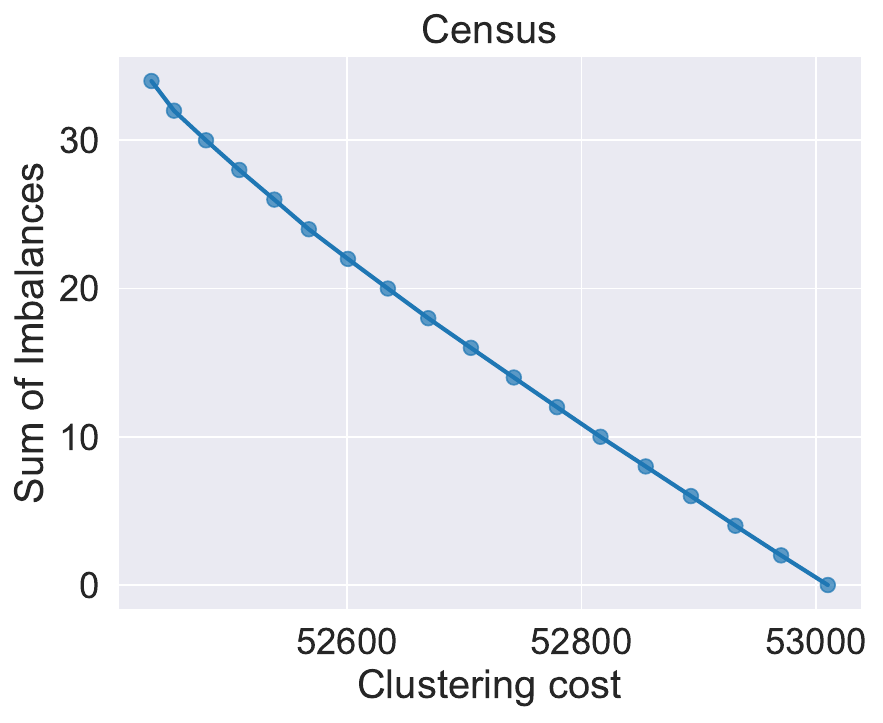}} 
\subfloat[]
{\includegraphics[width=0.34\textwidth] {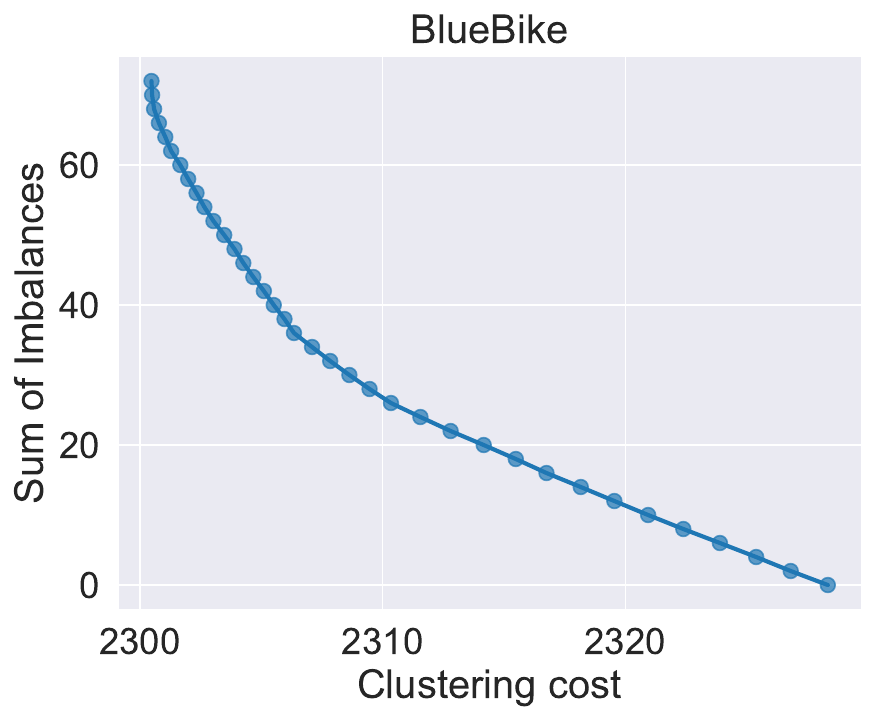}} \\
\subfloat[]
{\includegraphics[width=0.33\textwidth] {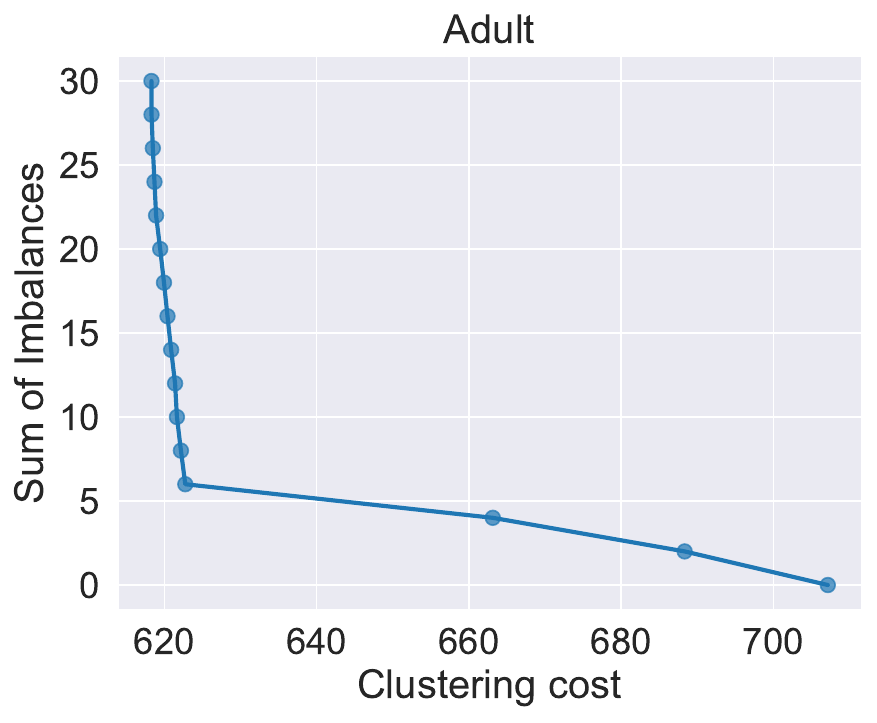}} 
\subfloat[]
{\includegraphics[width=0.34\textwidth] {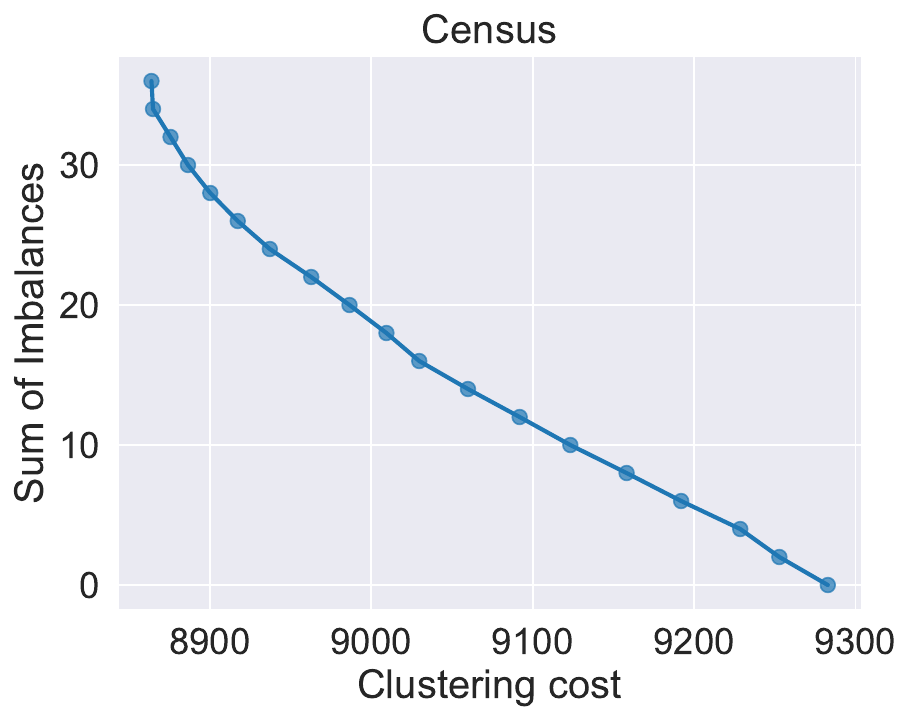}} 
\subfloat[]
{\includegraphics[width=0.33\textwidth] {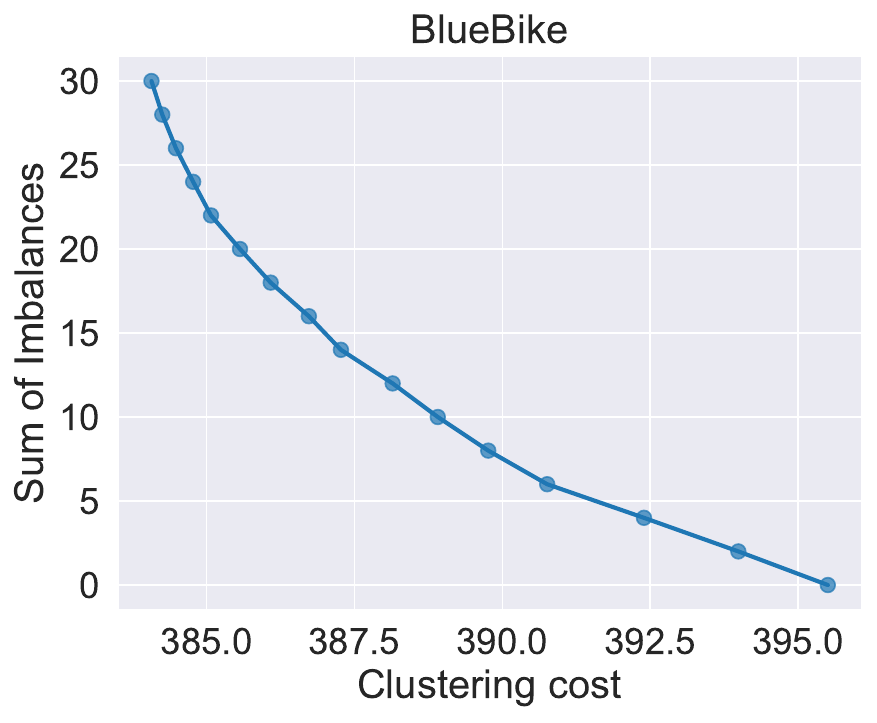}}
\caption{Pareto front recovered for the \textsc{Sum of Imbalances} objective for the Adult, Census, and BlueBike datasets (by column), for $k = 2$ (top row) and $k = 3$ (bottom row) clusters.}
\label{fig:adult_1000_dynprogr_sumimbalances-k2}
\end{figure*}

For the interested reader, we also showcase the Pareto front for the \textsc{Sum of Imbalances} objective, for all three datasets, for $k = 2$ and $k = 3$, in Figure~\ref{fig:adult_1000_dynprogr_sumimbalances-k2}. We note that since this objective is best suited for data with relatively equal proportions between the two groups, we subsample equal proportions of each gender for each of the three datasets, Adult, Census, and BlueBike.

